\documentclass{article}

\usepackage{microtype}
\usepackage{graphicx}
\usepackage{subcaption}
\usepackage{booktabs} % for professional tables
\usepackage{multirow} % for multi-row cells in tables

\usepackage{caption}
\usepackage{amsfonts}
\usepackage{amssymb}
\usepackage{mathrsfs}

\usepackage{hyperref}

\usepackage{enumitem}
\usepackage[accepted]{icml2026}

\usepackage{amsmath}
\usepackage{amssymb}
\usepackage{mathtools}
\usepackage{amsthm}

\usepackage[capitalize,noabbrev]{cleveref}

\theoremstyle{plain}
\newtheorem{theorem}{Theorem}[section]

\newtheorem{lemma}[theorem]{Lemma}

\theoremstyle{definition}

\newtheorem{assumption}[theorem]{Assumption}
\theoremstyle{remark}
\newtheorem{remark}[theorem]{Remark}

\usepackage[disable,textsize=tiny]{todonotes}
\icmltitlerunning{When LLMs Develop Languages}

\begin{document}

\twocolumn[
  \icmltitle{When LLMs Develop Languages: \\Symbolic Communication for Efficient Multi-Agent Reasoning}



  \icmlsetsymbol{equal}{*}

  \begin{icmlauthorlist}
    \icmlauthor{Zhengqi Pei}{ict,ucas}
    \icmlauthor{Qingming Huang}{ict,ucas}
    \icmlauthor{Shuhui Wang}{ict}
  \end{icmlauthorlist}

  \icmlaffiliation{ict}{State Key Lab of AI Safety, Institute of Computing Technology, Chinese Academy of Sciences, Beijing, China.}
  \icmlaffiliation{ucas}{School of Computer Science and Technology, University of Chinese Academy of Sciences, Beijing, China.}

  \icmlcorrespondingauthor{Shuhui Wang}{wangshuhui@ict.ac.cn}

  %\icmlkeywords{Machine Learning, ICML}
  \icmlkeywords{Large Language Models, Symbolic Communication, Reasoning Efficiency, Multi-Agent Systems}

  \vskip 0.3in
]



\printAffiliationsAndNotice{}  % no special notice (required even if empty)

\begin{abstract}
Chain-of-Thought (CoT) improves large language models (LLMs) on difficult reasoning tasks, but it often incurs long natural-language rationales that are poorly aligned with efficient machine reasoning.
We propose \emph{Communicative Language Symbolism Routing} (CLSR), a test-time framework in which multiple LLM agents autonomously invent, evolve, and share compact \emph{Language Symbolism Frameworks} (LSFs), while a latent-free router adaptively selects and composes these languages per query to optimize the accuracy--token trade-off.
Unlike prompt optimization that refines surface instructions, CLSR treats each LSF as a reusable symbolic protocol with compact symbols, usage rules, and a message-passing contract, and improves it through an evolutionary loop driven by correctness and token cost.
At inference time, the router may invoke a single low-cost LSF call, ensemble multiple LSFs, or execute a multi-round LSF composition protocol on harder queries.
Across challenging benchmarks, CLSR reduces latency-oriented generated token completion by $3\sim 6\!\times$ compared to standard CoT while maintaining accuracy.
We further derive an information-theoretic lower bound on token cost under arbitrary symbolism and show that, under an interpreter-realizability premise, multi-round LSF protocols conditionally subsume program-execution pipelines.
Code is publicly available\footnote{
%\url{https://github.com/pzqpzq/LSF_MDia}
\url{https://github.com/Plab-Lead/CLSR_MDia}
}
\end{abstract}

\begin{figure*}[t]
\centering
    \begin{minipage}[t]{0.46\textwidth}
        \centering\vspace{0pt}
        \subfloat[Conceptual comparison.
        \label{subfig: mainFig_concept}]
        {
        \includegraphics[width=\linewidth]{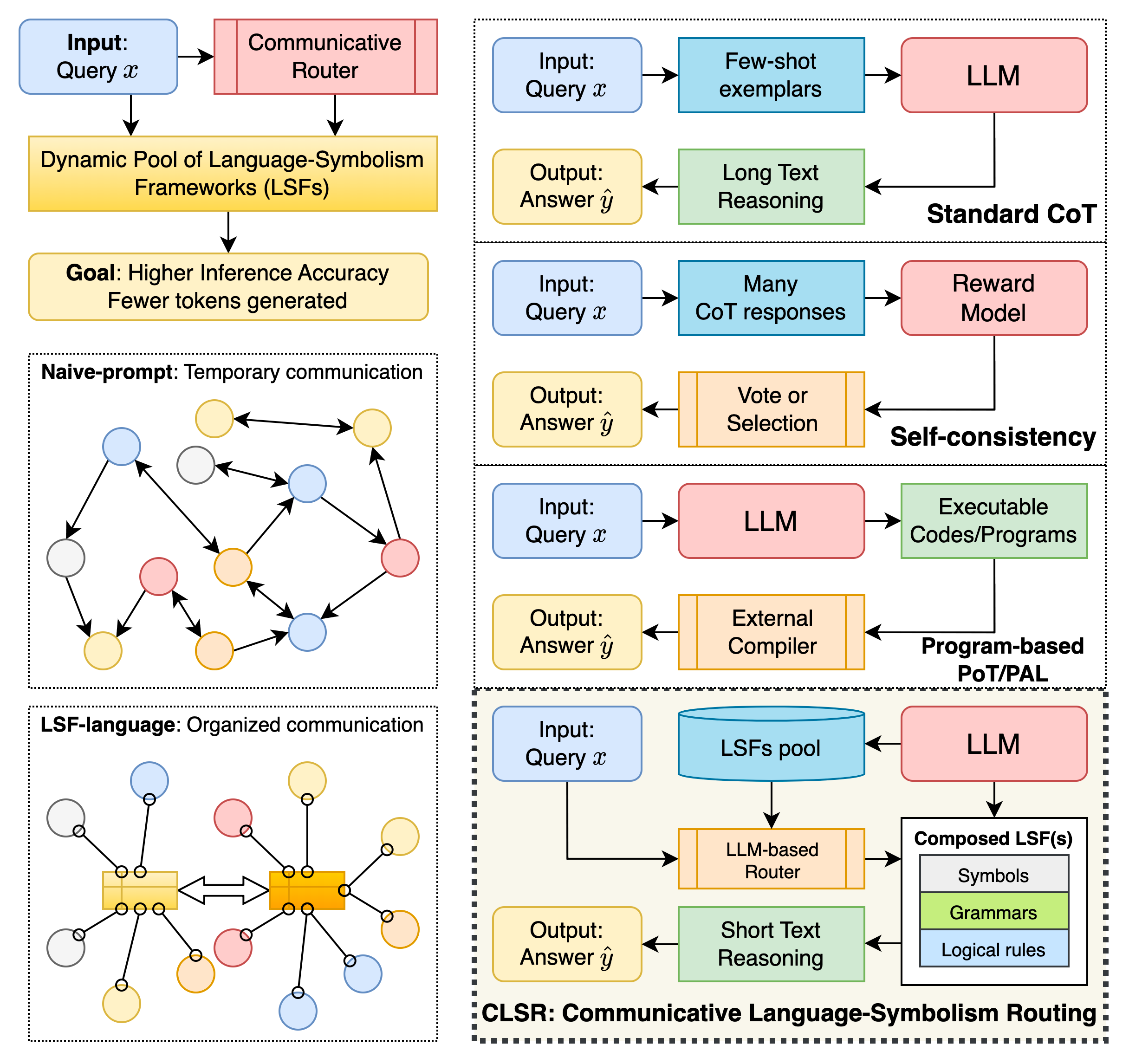}
        }
    \end{minipage}
    %\hfill
    \hspace{0.03\textwidth}
    \begin{minipage}[t]{0.42\textwidth}
        \centering\vspace{0pt}
        \subfloat[
        Multiple agents evolve new LSFs.
        \label{subfig: mainFig_evolvePLL}]
        {
            \includegraphics[width=\linewidth]{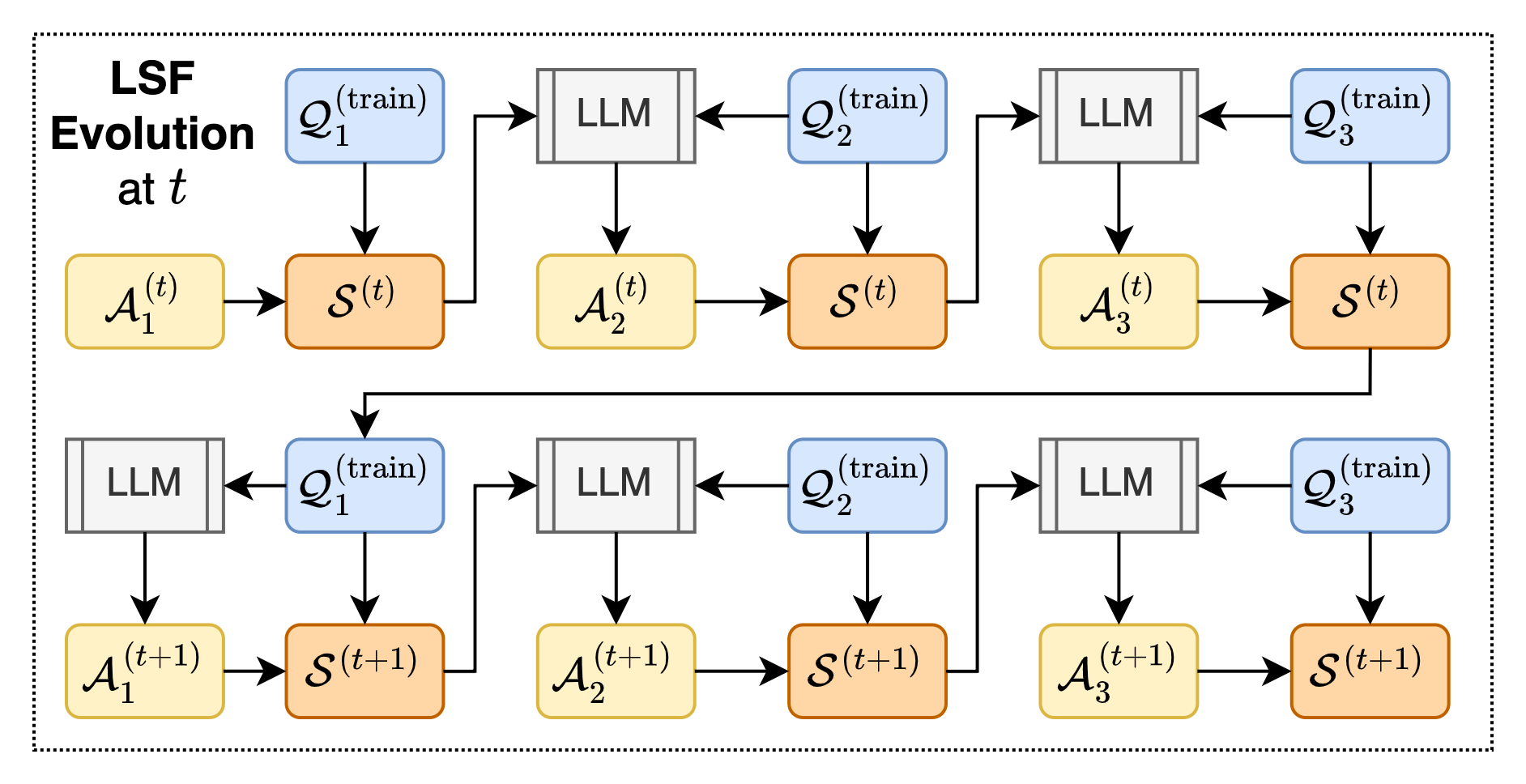}
        }
        \vfil
        \subfloat[Query-adaptive LSF planning.
        \label{subfig: mainFig_LLMrouter}]
        {
            \includegraphics[width=\linewidth]{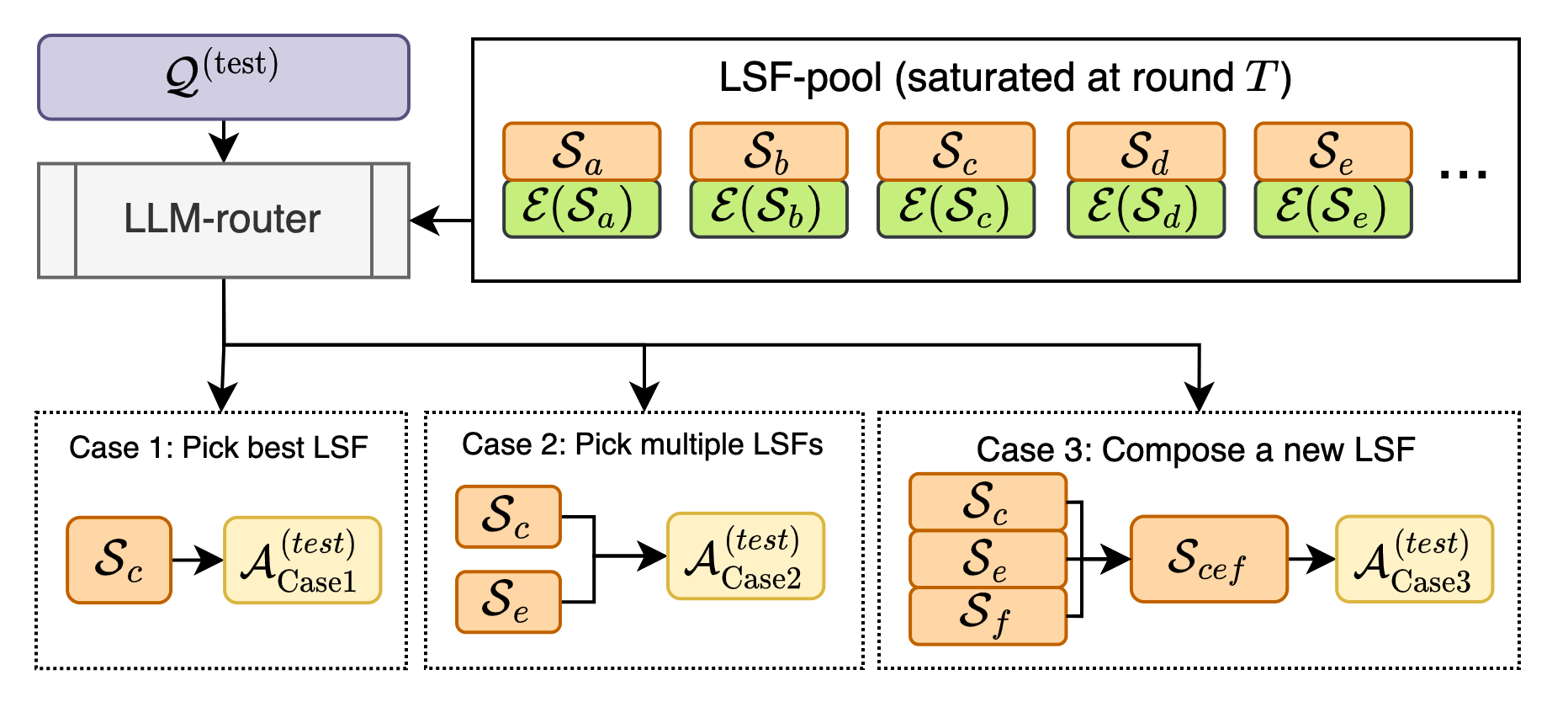}
        }
    \end{minipage}
    \caption{
    \textbf{Communicative Language Symbolism Routing (CLSR).} 
    \textbf{(a)} 
    We let LLMs self-evolve an LLM-oriented compact \emph{Language Symbolism Framework} (LSF), {\it i.e.}, an information-dense reasoning ``dialect'', rather than 
    expanding natural language rationales ({\it e.g.}, CoT) or relying on an external executor ({\it e.g.}, PoT).
    %, CLSR routes each query to a compact \emph{Language Symbolism Framework} (LSF) that specifies an information-dense reasoning ``dialect''.
    An \emph{LLM router} then selects, ensembles, or composes LSFs, enabling a controllable accuracy--token trade-off.
    \textbf{(b)} Multiple agents iteratively propose, critique, and mutate candidate LSFs based on high-leverage selection.
    \textbf{(c)} 
    The saturated LSF pool is then queried at inference time for query-adaptive protocol planning.
    }
    \label{fig: CLSR-pipeline}
\vskip -0.2in
\end{figure*}

\section{Introduction}

Large Language Models (LLMs) have demonstrated remarkable reasoning abilities, especially when employing chain-of-thought and self-consistency strategies that explicitly generate intermediate reasoning steps~\citep{wei2022chain,yao2023tree,wang2023self,besta2024graph}. 
Prompting an LLM to produce a step-by-step explanation often improves its accuracy on complex tasks~\citep{zhang2025prompt,sprague2025cot}. 
However, these reasoning chains are usually expressed in natural language designed for human understanding.
Linguistic reasoning chains can be verbose and are not specifically designed for compact logical inference~\citep{turpin2023language,tanmay2025orion}, thus they may not be the most efficient internal representation of the model~\citep{yu2024natural,chen2025towards},
especially for budget-constrained deployments~\citep{nayab2024concise,arora2025training}. 
This raises an intriguing question that we aim to address in this study: Can LLMs {\it invent} new symbolic languages to reason more efficiently?

Recent work leaves large gaps in our question.
% Specifically, 
% prompt optimizers such as 
% OPRO~\citep{yang2023large}, 
% PromptBreeder~\citep{fernando2024promptbreeder}, 
% Self-Discover~\citep{zhou2024self},
% and SoT~\citep{aytes2025sketch} 
Prompt optimization methods~\citep{yang2023large, fernando2024promptbreeder, zhou2024self, aytes2025sketch}
primarily refine natural-language instructions rather than inducing a persistent compositional symbol system that can be reused and transferred between tasks. 
% Program-based methods such as 
% PAL~\citep{gao2023pal},
% PoT~\citep{chen2023program},
% and LILO~\citep{grand2024lilo} 
Program-based methods~\citep{gao2023pal,chen2023program,grand2024lilo}
heavily depend on human-designed intermediate languages or programs, therefore, do not study whether LLMs can self-discover discrete symbolic languages optimized for their own tokenization and inductive biases. 
% RL-based approaches such as 
% ORION~\citep{tanmay2025orion} 
% and RLEF~\citep{gehring2025rlef} 
RL-based approaches~\citep{tanmay2025orion,gehring2025rlef} 
produce compact symbolic traces using verifier-based RL, yet these RL pipelines typically incur substantial training cost and optimization instability, making them heavy-weight and less modular for rapid multi-agent language evolution.
% Compression methods such as 
% CCoT~\citep{cheng2024compressed} 
Compression methods~\citep{cheng2024compressed} 
reduce the length of explicit reasoning by moving computation into dense tokens, but they do not yield a socially shareable discrete ``language'' that can evolve via inter-agent adoption and merger. 
Neuro-symbolic approaches~\citep{beiser2025intermediate} and logical reasoning~\citep{xu2025rlg} show that an intermediate language can motivate a systematic search over representations; 
yet, their pipelines typically select among fixed formal languages rather than evolve new ones in a socially grounded way.

In parallel, emergent communication methods~\citep{lazaridou2020emergent, tucker2022trading,pei2024modeling,pei2025neuronal} formalize how discrete protocols can arise under computational pressures, offering an information-theoretic lens on why structured conventions emerge.
Motivated by this perspective, we ask whether LLM agents can autonomously design and evolve languages, then subject them to natural selection.
We operationalize this question as an evolutionary search over reusable LSFs: protocols that are simultaneously accurate, compact, and reusable are preferentially retained and made available for downstream routing.
%We assume that a language that is more accurate and more token-efficient can be much more transferable and mergeable across agents.
This setting yields emergent large-scale symbolic systems shaped by collective use rather than isolated prompt tuning\footnote{
See the Appendix~\ref{app-sec: related_work} for more discussion of related work.
}.

In this work, we introduce CLSR (Communicative Language Symbolism Routing), a paradigm in which multiple LLM agents autonomously develop their own symbolic communication systems for reasoning. 
Each agent is a black-box LLM instance that proposes, evaluates, or refines a Language Symbolism Framework (LSF), where an LSF is a customized set of symbols, syntax, and usage rules that serves as a reusable reasoning protocol.
%Each LLM treated as an independent agent is tasked with creating a Language Symbol Framework (LSF),
%a customized set of tokens, syntax, and logical rules,
%to serve as its private \emph{reasoning language}. 
% These LSFs are not pre-defined by humans; 
% instead, the LLMs themselves generate their languages from scratch, specifying the meanings of invented symbols and inference rules in a guideline prompt. 
We emphasize that LSFs are not manually pre-defined by humans as a fixed formal language.
They are induced by the LLMs from exemplars and refined through selection.
Thus, our claim is not that LSFs are free of human priors.
Rather, the claim is that the \emph{operational protocol} used at inference time can be automatically generated, selected, shared, and composed by LLM agents, rather than being crafted manually as a fixed prompt or program.

This view naturally connects CLSR to a socio-linguistic perspective inspired by the cultural language evolution process~\citep{kouwenhoven2025searching}.
Across evolutionary rounds, more effective LSFs are preferentially retained and reused, while less accurate or overly verbose LSFs are discarded.
The resulting process resembles cultural language evolution under an explicit communication bottleneck: 
correctness plays the role of communicative success, token length plays the role of production cost, and routing plays the role of pragmatic code-switching between compact and redundant LSFs.
Over generations, more effective LSFs spread across agents, while less useful ones are discarded, leading to converging robust symbolic languages.

We design experiments on a diverse set of challenging tasks, including knowledge-intensive QA~\citep{wang2024mmlu,rein2023gpqa,lu2022sciqa}, mathematical problem solving~\citep{cobbe2021gsm8k,hendrycks2021math500,aime_1983_2024}, and multi-hop reasoning~\citep{yang2018hotpotqa}, to evaluate the impact of LLM-created languages. 
%In particular, we use seven benchmarks (MMLU-Pro~\citep{wang2024mmlu}, GPQA~\citep{rein2023gpqa}, GSM8K~\citep{cobbe2021gsm8k}, MATH500~\citep{hendrycks2021math500}, AIME (21-24)~\citep{aime_1983_2024}, ScienceQA~\citep{lu2022sciqa}, and HotpotQA~\citep{yang2018hotpotqa}),
%covering domains from academic trivia to math word problems. 
For generality, we conduct experiments with various open-source LLMs, {\it e.g.}, Qwen3-8B/32B~\citep{yang2025qwen3}, LLaMA3-8B~\citep{llama3modelcard}, 
and DeepSeek-R1~\citep{deepseekai2025deepseekr1},
%Mistral-7B~\citep{jiang2023mistral7b}, 
treating each model as an agent in different runs. 
Across multiple open-source backbones, CLSR improves the Pareto frontier of accuracy versus generated tokens, outperforming baselines covering token-reduction, program-execution, and prompt optimization. 
% for instance, on LLaMA3-8B, CLSR reduces GPQA tokens from 1209 to 362 while slightly improving accuracy by 0.4\%, and reduces MMLU-Pro tokens from 960 to 320 while improving accuracy by 0.3\%. 
% In Qwen3-8B, CLSR cuts GPQA tokens from 1085 to 228 (4.8$\!\times$) and reduces MATH500 tokens from 878 to 287 (3.1$\!\times$) while preserving accuracy. 

Ablations further show that 
(i) deeper evolutionary generations systematically shift languages toward better accuracy-per-token, 
(ii) scaling the agent population improves the chance of discovering robust, reusable LSFs, provided the selection objective balances correctness and compression.
% We also analyze how these languages evolve and amalgamate through inter-agent communication, potentially forming larger unified languages across various domains.
These empirical results lead to an information-theoretic interpretation regarding the curve between reasoning accuracy and token efficiency, given any language symbolism.
% By measuring both the reasoning accuracy and the token efficiency, we investigate whether LLM-generated symbolic languages can produce better-than-baseline performance with fewer tokens. 
We also conduct larger-backbone transfer, population-size ablations, cache-aware token accounting, and category-free domain transfer to validate methodological robustness.
In summary, this work presents an approach to improving LLM reasoning through socially emergent symbolic communication. 
%In summary, this work introduces socially emerging symbolic communication as a practical mechanism for \emph{compressing} reasoning in test time in frozen LLMs. 
Our main contributions include:
% Our contributions include: 
\begin{itemize}[noitemsep, nolistsep]
    \item \textbf{CLSR: routing over emergent languages.} 
    We propose a new test-time paradigm that treats LLM-created symbolic languages (LSFs) as modular ``experts'' and routes, ensembles, or composes them per query to explicitly optimize the accuracy--token budget trade-off.
    \item \textbf{LLM-driven language synthesis and evolution.} 
    We introduce an evolutionary bootstrapping procedure that generates LSFs from scratch; we further enable cross-agent adoption and cross-benchmark pooling to promote the transferring and merging of LSFs.
    \item \textbf{Strong empirical Pareto gains.} 
    On seven diverse reasoning benchmarks and multiple open-source backbones, CLSR consistently improves the accuracy--efficiency frontier, outperforming representative token reduction baselines.
    \item \textbf{A theory of accuracy--token optimality under symbolism.} 
    We formalize CLSR as a constrained stochastic control problem, derive an information-theoretic lower bound on minimal token cost, and characterize multi-round multi-LSF protocols as a conditional generalization of program-execution inference.
\end{itemize}

%\clearpage
\section{Methodology}
\label{sec: methodology}
We introduce the CLSR framework for test-time reasoning compression through LLM-generated symbolic languages.
Unlike prompt optimization that searches for instructions or alters surface phrasing, 
CLSR optimizes \emph{a reusable communication protocol}:
(i) the intermediate messages are constrained by grammar/contract and validated;
(ii) the same LSF is reused across many queries (not per-instance prompt tuning);
(iii) the router explicitly composes multiple LSF calls into a multi-round protocol under a cost budget.
During inference, the LLM participates in a multi-round communication where different LSFs act as specialized ``experts'', and a router adaptively selects which LSFs should speak, in which order, and when to stop, explicitly optimizing the accuracy--token trade-off.
These ``experts'' are emergent symbolic languages rather than separate neural modules, and the LLM's weights remain inaccessible.
% Empirically, we compare against prompt-optimization baselines under matched budgets to isolate gains due to symbolic protocol structure rather than improved instruction phrasing.

%Unlike prompt optimization that searches for instruction strings or alters surface phrasing, 
% CLSR treats each LLM-generated symbolism LSF as a modular reasoning protocol with 
% (i) a compact alphabet or symbolism, (ii) a constrained grammar, and (iii) a message-passing contract.
% In inference, the LLM does not merely ``think in a quirky style''; 
% instead, it participates in a multi-round communication protocol where different LSFs act as specialized ``experts'', and a router (a lightweight network or LLM) adaptively selects which LSFs should speak, in which order, and when to stop, explicitly optimizing the accuracy--token trade-off.
% This design is conceptually aligned with conditional computation ideas, {\it e.g.}, routing among experts, but differs in that our ``experts'' are emergent symbolic languages rather than separate neural modules, and the LLM's weights remain frozen and inaccessible.

\subsection{Problem Setup and Metrics}
\label{subsec: Problem setup and metrics}

Let $x$ be a query and $y$ be the ground-truth answer drawn from a benchmark.
Suppose that we have an evolved LSF pool $\mathcal{S}=\{\mathcal{S}_{k}\}_{k=1}^{K}$.
In inference, CLSR runs for $T$ rounds, where $T$ can be fixed or determined by the router based on the test query.
At round $t\in\{1,\ldots,T\}$, the router selects an index set $\mathcal{I}_{t}\subseteq [K]$, where $[K]=\{1,\ldots,K\}$, and invokes the corresponding LSF subset $\{\mathcal{S}_{k}:k\in\mathcal{I}_{t}\}$.
The LLM then produces one response per selected LSF.
Empirically, a harder problem often yields a larger $T$ or $\lvert \mathcal{I}_{t}\rvert$.
% Let $x$ be a query, and $y$ be the ground-truth answer drawn from a benchmark.
% Suppose that we currently have $K$ generated or hand-made LSFs.
% In inference, CLSR runs for $T$ rounds, where $T$ can be fixed or determined by the router based on the test query.
% In round $t\in\{1,...,T\}$, a router selects a subset of $K$ discrete LSFs $\mathcal{S}_t$ with $\lvert \mathcal{S}_t \rvert \leq K$.
% The LLM then produces one response per selected LSF.
% Empirically, a harder problem often yields a higher $T$ or $\lvert \mathcal{S}_t \rvert$.

\textbf{Compute cost (token accounting).}
Let $r_{t,k}$ denote the LLM response produced in round $t$ under LSF $\mathcal{S}_{k}$, and let $r_{t}^{\text{router}}$ denote the LLM-router's generated planning output in round $t$.
We define the generation cost as
\begin{equation}\label{eq:main-realized-cost}
    \begin{gathered}
    C \;=\; \sum_{t=1}^{T} \Big(\lvert r_{t}^{\text{router}} \rvert + \sum_{k\in \mathcal{I}_{t}} \lvert r_{t,k}\rvert\Big),
    \end{gathered}
\end{equation}
where $\lvert\cdot\rvert$ counts generated completion tokens.
Thus, all online tokens emitted by the LLM are included, including router outputs, intermediate LSF responses, and aggregation-related responses.
This metric is motivated by the standard latency decomposition of LLM inference into a prompt \emph{prefill} phase (time-to-first-token, TTFT) and an auto-regressive \emph{decode} phase (time-per-output-token, TPOT).
For reasoning workloads with long completions, the decode term is often the dominant latency component, so completion tokens provide a useful latency-oriented proxy.
At the same time, LSF profiles do introduce input-token overhead.
In deployment, these cards can be arranged as a fixed canonical prompt prefix and reused across many queries, so the relevant overhead depends on prefix caching and serving infrastructure.
For this reason, we additionally report a cache-aware token-equivalent metric in Appendix~\ref{app:latency_token_accounting}, while keeping $C$ as the main generated-token metric throughout the paper.

% \textbf{Compute cost (token accounting).}
% Let $r_{t,k}$ denote the LLM response produced in round $t$ under LSF $k$, and let $r_{t}^\text{router}$ denote the LLM-based router's response in round $t$.
% %$|r_{t,k}|$ count the \emph{generated} (completion) tokens of that call. 
% We define the total generation cost as
% \begin{equation}
%     \begin{gathered}
%     C \;=\; \sum_{t=1}^{T} \Big(\lvert r_t^\text{router} \rvert + \sum_{k\in S_t} \lvert r_{t,k}\rvert\Big).
%     \end{gathered}
% \end{equation}
% This choice is motivated by the standard latency decomposition of LLM inference into a prompt \emph{prefill} phase (time-to-first-token, TTFT) and an auto-regressive \emph{decode} phase (time-per-output-token, TPOT). 
% For a single request, the latency is well-approximated by $\text{TTFT}+\text{TPOT}\cdot C\approx t_\text{token}\cdot C$, so for reasoning workloads with long completions, the decode term dominates. 
% In our accounting, \emph{all tokens emitted by LLM are included}, including any router-emitted tokens and any intermediate outputs used for aggregation. 
% For completeness, we provide a latency-oriented interpretation in the Appendix~\ref{app:latency_token_accounting} regarding TTFT/TPOT.

\textbf{Objective.}
Given a budget $B$, our objective is to maximize expected correctness under the expected realized cost:
\[
\max_{\pi}\; \mathbb{E}\!\left[\mathbb{I}\{\hat{y}=y\}\right]
\quad
\text{s.t.}
\quad
\mathbb{E}[C]\le B,
\]
where policy $\pi$ specifies the router, stopping, and aggregation rules.
The expectation is over the benchmark distribution and any policy-induced randomness.

\subsection{Language Symbolism Framework (LSF)}

\textbf{Conceptual definition.}
In idealized form, an LSF can be described as a symbolic communication protocol comprising three components:
(i) symbol naming, namely a compact lexicon;
(ii) syntax, namely a compositional grammar; and
(iii) constraints, namely well-formedness and usage rules.
However, in practice, fully hand-designing these components would collapse the method into prompt engineering.
We therefore delegate LSF construction to the LLM itself and use human intervention only to define the high-level optimization goal: to preserve reasoning accuracy while reducing token usage.

\paragraph{Operational LSF card.}
In implementation, each LSF is stored as a compact card
$S_k = \big(\mathcal{V}_k,\mathcal{G}_k,\mathcal{R}_k,\psi_k,\rho_k\big)$, where $\mathcal{V}_k$ is the symbol inventory, $\mathcal{G}_k$ is a lightweight grammar or output schema, $\mathcal{R}_k$ contains usage rules and validity constraints, $\psi_k$ maps symbols or templates to their intended reasoning operations, and $\rho_k$ is an empirical profile summarizing cost, accuracy, domains, and failure modes.
The card is represented textually because the backbone is a language model, but its role is not that of a one-off instruction string.
An LSF is reused across many queries, produces messages constrained by a grammar or contract, can be selected or composed by the router, and is evaluated as a persistent protocol whose utility is measured over a population of tasks.
This operational distinction is important: prompt optimization searches for better surface instructions, whereas CLSR searches over reusable symbolic communication systems and their routing policies.

% \textbf{Conceptual definition.} 
% In idealized form, an LSF can be described as a symbolic communication code comprising three components: 
% (i) symbol naming (a compact lexicon), 
% (ii) syntax (a compositional grammar), and 
% (iii) constraints (well-formedness and usage rules). 
% However, in  practice, hand-designing these components risks injecting human bias and collapsing the method into prompt engineering. 
% Therefore, we make a deliberate design choice: 
% the entire LSF construction process is delegated to the LLM itself.
%, with minimal scaffolding.

\textbf{Seed exemplars and LSF synthesis prompt.}
Given a benchmark training set, we randomly sample a small set of QA exemplars (typically dozens) and place them in context.
We then prompt the LLM to freely invent a symbolic language that aims to reduce token usage while preserving reasoning capability, {\it e.g.},
design an LSF based on the exemplars in the chat history to minimize the number of tokens while maintaining reasoning capacity.
Manual editing, symbol pruning, or grammar correction is not performed in the default pipeline.
The exact prompt templates for LSF operations are provided in the Appendix~\ref{app-sec: prompt_templates}.
%The exact prompt templates for LSF synthesis, mutation, profile summarization, and routing are provided in the Appendix~\ref{app-sec: prompt_templates}.

% \textbf{Seed exemplars and LSF synthesis prompt.} 
% Given a benchmark training set, we randomly sample a small set of QA exemplars (typically dozens) and place them in-context. 
% We then prompt the LLM to freely invent a symbolic language that aims to reduce token usage while preserving reasoning capability, 
% {\it e.g.}, 
% design an LSF based on the exemplars in the chat history to minimize the number of tokens while maintaining the reasoning capacity.
% Manual editing, symbol pruning, or grammar correction is not performed.

\textbf{Diversity and positioning.}
We observe that a higher sampling temperature naturally yields a diverse LSFs population, spanning a spectrum from strict LSFs (machine-like, compressed, rigid formats) to soft LSFs (closer to natural language but still symbolically structured).
This diversity is crucial for downstream selection and routing, as different symbolic protocols may dominate on different sub-distributions of queries.
While the generation loop superficially resembles iterative prompt discovery, the optimized object is not an instruction string but a reusable symbolic language system.
LSFs are repeatedly invoked across many queries, come with a compact grammar/contract, and can be selected or composed by the router as a modular protocol.
This distinguishes CLSR from prompt evolution / optimization methods~\citep{yang2023large,fernando2024promptbreeder}.

\subsection{LSF Evolution: Iterative Bootstrapping}

We introduce an evolutionary bootstrapping procedure that progressively refines the LSF pool using only training data and black-box LLM calls.

\textbf{Agent population.}
In our implementation, an ``agent'' is an independently sampled black-box LLM instance specified by a backbone, random seed, exemplar subset, and LSF-generation context.
Agents are not separate trainable neural modules; they are proposal, critique, and mutation workers used to diversify the LSF population.
Different agents may discover different surface LSFs, but the selection objective favors LSFs that repeatedly achieve high correctness with low generated-token cost.
This definition also clarifies the agent-count ablation in Appendix~\ref{app-abl: Effect of number of agents and length-coefficient}: increasing the number of agents expands the search over possible symbolic conventions, while the final inference model remains frozen.

\textbf{Dataset partition and generational schedule.}
We divide the training set into $M$ disjoint groups $\{\mathcal{G}_1,...,\mathcal{G}_M\}$, each containing exemplars $(x,y)$, then sequentially generate:
\begin{enumerate}[noitemsep, nolistsep]
    \item \textbf{Generation}. Use exemplars from $\mathcal{G}_1$ to induce an initial LSF pool $\mathcal{S}^{(1)}=\{\mathcal{S}_k^{(1)}\}_{k=1}^{K}$.
    \item \textbf{Evaluation}. For each $\mathcal{S}_k^{(1)}$, query the LLM on the queries of $\mathcal{G}_2$, produce answers $\hat{y}$ and record the token cost of each answer.
    \item \textbf{Selection and mutation}. From all the answers generated on $\mathcal{G}_2$, select a subset of high-leverage (correct and token-efficient) answers and feed them to the LLM to generate the next-generation LSF pool $\mathcal{S}^{(2)}$.
    \item \textbf{Repeat}. use $\mathcal{S}^{(t)}$ on $\mathcal{G}_{t+1}$, select high-leverage answers, and synthesize $\mathcal{S}^{(t+1)}$ until the performance in the validation set is saturated.
\end{enumerate}
This procedure is reminiscent of iterative optimization in prompt-evolution systems,
%(generate candidates $\mapsto$ evaluate $\mapsto$ retain the best traces $\mapsto$ regenerate), 
but differs in that the feedback signal is used to evolve symbol inventories and constraints rather than only instructions in natural-language.

\textbf{High-leverage selection and mutation.}
Let an LSF-conditioned answer be a tuple $(\hat{y},c)$, where $c$ is the generated token count.
We prioritize the reasoning traces that are:
(i) \emph{correct}, meaning $\hat{y}=y$ under the benchmark's evaluator, and
(ii) \emph{token-efficient}, meaning that $c$ is low relative to other correct candidates for the same query.
In practice, we implement selection via a Pareto criterion, {\it i.e.}, accuracy first and then minimal tokens.
The selected high-leverage traces, together with the parent LSFs that produced them, are then used as context for the mutation step.
The LLM is asked to refine or recombine the parent symbolic protocol so that it preserves the successful inference pattern while removing redundant notation, ambiguous rules, or failure-prone conventions.
Thus, mutation operates on the LSF definition rather than merely rewriting an output answer.
%The exact mutation prompt is given in Appendix~\ref{app-sec: prompt_templates}.

% \textbf{High-leverage selection.}
% Let an LSF-conditioned answer be a tuple $(\hat{y},c)$, where $c$ is the generated token count. 
% We define a selection rule that prioritizes solutions that are:
% (i) \emph{Correct} ($\hat{y}=y$ under the benchmark's evaluator);
% (ii) \emph{Token-efficiency} ($c$ is low relative to other correct candidates for the same query).
% % \begin{itemize}[noitemsep, nolistsep]
% %     \item \textbf{Correct}: $\hat{y}=y$ under the benchmark's evaluator.
% %     \item \textbf{Token-efficiency}: $c$ is low relative to other correct candidates for the same query.
% % \end{itemize}
% In practice, we implement selection via a Pareto criterion (accuracy first, then minimal tokens).
% %, optionally with diversity constraints to avoid collapsing to a single dialect.

\textbf{Elitist inheritance across generations.}
If certain LSFs consistently generate high-leverage answers, we allow them to survive unchanged into the next generation.
In practice, the best performing LSFs $\mathcal{S}_{k}^{(t)}$ are copied into $\mathcal{S}^{(t+1)}$ without modification, while the remaining population is filled with mutated or recombined variants.
This prevents the population from losing strong LSFs due to stochastic mutation, while still allowing exploration.

% \textbf{Elitist inheritance across generations.}
% If certain LSFs consistently generate high-leverage answers, we allow them to survive unchanged into the next generation (elitism). 
% In practice, the best performing LSFs $\mathcal{S}_{k}^{(t)}$ are copied into $\mathcal{S}^{(t+1)}$ without modification.
% %and their capacity might be updated with newly synthesized variants derived from selected answer traces.

\textbf{LSF Profile: Cost–Accuracy Metadata for Routing.}
%To make the LSF pool usable beyond static prompting, w
We maintain an LSF's profile summarizing empirical behavior over training rollouts. 
For each LSF $\mathcal{S}_k$, we track:
(i) Accuracy estimate in evaluated groups,
(ii) Token statistics (mean/median, tail behavior),
(iii) Reliability indications (variance between query types, failure modes), and
(iv) Domain tags inferred from the queries where $\mathcal{S}_k$ excel.
% \begin{itemize}[noitemsep, nolistsep]
%     \item \textbf{Accuracy estimate} in evaluated groups,
%     \item \textbf{Token statistics} (mean/median, tail behavior),
%     \item \textbf{Reliability indications} (variance between query types, failure modes), 
%     \item \textbf{Domain tags} inferred from the queries where $\mathcal{S}_k$ excel.
% \end{itemize}
These profiles are later consumed by a router to decide which LSF(s) to apply in inference.

\subsection{Test-time LSF Routing}

CLSR moves beyond ``choose one prompt'' and instead treats LSFs as a pool of symbolic protocols that can be selected, ensembled, and composed per query. 
%We consider two routing implementations.

% \paragraph{Option 1: A lightweight learned router (encoder+MLP)}
% We train a smaller router network that predicts which LSF(s) are most suitable for a given query.
% %\textbf{Latent representations.}
% We use an external pretrained embedding model $E(\cdot)$ to embed:
% (i) the query $x$, and 
% (ii) each LSF together with its profile summary. 
% This yields $e_x{=}E(x)$ and $e_k{=}E\big(\text{spec}(\mathcal{S}_k),\text{profile}(\mathcal{S}_k)\big)$.
% %\textbf{Routing model.}
% A lightweight MLP computes scores $s_k{=}f(e_x,e_k)$ and outputs either a single best LSF or a top-$r$ subset for ensembling.
% We use historical rollouts to train $f$: 
% for each query, the router is trained to predict which LSF achieves the best cost–quality trade-off under a chosen objective.
% This choice is conceptually related to cost–quality routing in multi-LLM serving~\citep{ding2024hybrid}, where a router selects among models, but our selection space is symbolic languages rather than different backbone LLMs.

%\paragraph{Option 2: Latent-free LLM router}
\textbf{Latent-free LLM-routing policy}
To eliminate additional trainable routing networks and access to the latent states of an LLM or network,
we can use an LLM-router by delegating routing decisions to the same LLM. 
This is achieved by (i) compressing each LSF's ``battle'' profile into a short description 
and (ii) prompting the LLM to synthesize a query-specific protocol for leveraging the LSF pool.

\textbf{LSF profile summarization.}
Offline, we ask the LLM to produce a concise descriptor for each LSF conditioned on its empirical profile: 
what types of queries it tends to solve well, typical token footprint, and common failure cases.

\textbf{Protocol planning.}
At test time, given a query $x$ and the set of LSF descriptors, the LLM first decides among three inference modes:
\begin{enumerate}[noitemsep, nolistsep]
    \item \textbf{Single LSF direct answer}: select one LSF and generate a final answer.
    \item \textbf{multi-LSF aggregation}: select multiple LSFs to independently propose answers, then aggregate (e.g., majority vote), similar to self-consistency. 
    \item \textbf{Implicit LSF composition}: for hard queries, the router specifies a multi-step protocol where responses from many LSFs become a context for subsequent rounds.
    %, {\it e.g.}, ``use LSF-A to compress the problem; use LSF-B to derive; use LSF-C to check''.
    %, aligning with broader test-time deliberation frameworks that explore or refine intermediate ``thought units''. 
\end{enumerate}
Concretely, the router produces a plan:
(i) the number of rounds,
(ii) which LSF(s) to invoke per round,
(iii) how to post-process intermediate responses,
and (iv) when to stop.
See Appendix~\ref{app-sec: router_plan} for more details.
%This design is compatible with iterative self-improvement, {\it e.g.}, refinement via reflection memory, but instantiated as symbolic-protocol routing rather than free-form natural language deliberation.

% \paragraph{Choice of router in our pipeline.}
% Empirically, Option 1 and Option 2 provide comparable overall performance. 
% Since Option 2 avoids introducing any additional trainable neural components and preserves a fully ``LLM-driven'' pipeline, we adopt Option 2 (LLM-router) as the default implementation.

% \paragraph{Why LSF is not mere prompt optimization.}
% An LSF is a \emph{protocol} with an explicit \emph{interface contract}:
% (i) a structured message schema (fields, allowable symbols, and invariants),
% (ii) a deterministic \emph{execution semantics} that maps protocol fields to LLM calls and state updates,
% and (iii) an enforcement layer (validation and refine) that ensures well-formed protocol states.
% In contrast, prompt optimization changes an instruction string but typically lacks (ii) and (iii).
% %: the system does not validate intermediate states nor enforce schema invariants across rounds.
% In our implementation, each LSF comes with a validater and lightweight refinement, so that the LLM solver is always conditioned on a valid protocol state.

\subsection{End-to-End Inference Procedure}
Given a test query $x$ and an evolved LSF pool $\mathcal{S}$, we:
\begin{enumerate}[noitemsep, nolistsep]
    \item \textbf{Retrieve LSF descriptors} and optionally a small subset of profile statistics.
    \item \textbf{Router planning}: the LLM outputs a protocol specifying which LSF(s) to use, whether to parallelize, and whether to run multiple rounds.
    \item \textbf{Execute protocol}: generate intermediate responses as specified. All generated tokens across all rounds and all invoked LSFs are counted toward the total cost.
    \item \textbf{Aggregate to the final answer} according to the router-generated or self-consistency protocol.
\end{enumerate}
This yields an adaptive reasoning pattern: 
easy queries often require a single low-cost LSF call ($T{=}1$), whereas difficult queries may trigger multi-round LSF communication ($T{>}1$), trading extra tokens for higher reasoning reliability and inference accuracy.

%\clearpage
\section{Theoretical Analysis}
\label{sec: theory}

This section provides a compact, theory-oriented view of CLSR.
The formal proofs are deferred to the Appendix~\ref{app-sec: more theoretical validation}.
We also include a sociolinguistic interpretation of the evolution of LSF, which explains why compact machine-oriented LSFs can emerge from repeated selection under an accuracy--efficiency pressure.

% First, we formalize test-time LSF routing as a constrained stochastic control problem and characterize the accuracy--token Pareto frontier.
% Second, we derive an information-theoretic lower bound showing how the required number of generated tokens depends on the target accuracy, task difficulty, and the per-token information rate of the symbolic protocol.
% Third, we compare CLSR with program-execution pipelines and show that multi-round multi-LSF protocols can be viewed as an in-model generalization of ``generate a program then execute'' inference under an interpreter-realizability premise.
% The formal proofs are deferred to the Appendix~\ref{app-sec: more theoretical validation}.
% We also include a sociolinguistic interpretation of the evolution of LSF, which explains why compact machine-oriented dialects can emerge from repeated selection under an accuracy--efficiency pressure.

% In this section, we formalize two theoretical results:
% (i) the information lower-bound on the number of tokens generated by an LLM for target reasoning accuracy, and
% (ii) the equivalence between multi-round multi-LSF protocols and arbitrary LLM-generated programs.
% %, and
% (ii) a brief socio-linguistic lens on why LLM-generated languages emerge.
% The formal statements and full proof are presented in the Appendix~\ref{app-sec: more theoretical validation}.

\subsection{Token-Accuracy Optimality}
We formalize CLSR test-time inference as a constrained stochastic control problem.
For a fixed test query $x$, an interactive inference policy $\pi$ (routing, stopping, aggregation) induces a multi-round transcript $\mathscr{T}$ (all tokens generated across invoked LSFs/rounds) and a final prediction $\hat{Y}$.
We evaluate policies according to expected token cost $J_C(\pi)=\mathbb{E}\big[\lvert \mathscr{T} \rvert\big]$ and accuracy $J_A(\pi)=\mathbb{P}\big[\hat{Y}= Y\big]$, and define the optimal accuracy within the budget $B$ as 
\begin{equation}
    \begin{gathered}
        A^*(B)=\sup_{\pi:J_C(\pi)\leq B}J_A(\pi).
    \end{gathered}
\end{equation}

\begin{theorem}\label{thm: existence}
    \textbf{(Existence of an optimal interactive inference policy).}
    For a query $x$.
    Assume:
    (i) a finite LSF pool of size $K$;
    (ii) a finite (or effectively bounded) horizon $T_{max}$ via a stopping action;
    (iii) the per-step token cost is nonnegative and integrable;
    (iv) the LLM decoding under each LSF defines a well-posed stochastic kernel over the next tokens conditioned on the current history.
    % \begin{enumerate}[nolistsep]
    %     \item a finite LSF pool of size $K$;
    %     \item a finite (or effectively bounded) horizon $T_{max}$ via a stopping action;
    %     \item the per-step token cost is nonnegative and integrable;
    %     \item the LLM decoding under each LSF defines a well-posed stochastic kernel over the next tokens conditioned on the current history.
    % \end{enumerate}
    Then for every budget $B\geq 0$, the supremum $A^*(B)$ is attained: 
    there exists a possibly randomized policy $\pi_{B}^*$ such that $J_C(\pi_{B}^*)\leq B$ and $J_A(\pi_{B}^*)=A^*(B)$.  
\end{theorem}

Moreover, for any Lagrange multiplier $\lambda\geq 0$, there exists an optimal deterministic finite-horizon policy $\pi_{\lambda}^*$ that maximizes the objective $\mathbb{E}\big[{\bf 1}\{\hat{Y}=Y\}-\lambda\lvert \mathscr{T}\rvert\big]$;
the boundary points of the Pareto frontier $(J_C,J_A)$ can be implemented by some $\pi_{\lambda}^*$ or a mixture of two adjacent multipliers.  

% \begin{proof}
%     (scratch).
%     Model the interaction as a finite-horizon constrained MDP (CMDP) whose state is the (measurable) transcript/history and whose actions are ``choose an LSF subset + stop/continue + aggregation rule'', consistent with our definition of policy decisions. 
%     For finite action sets and finite horizons, standard dynamic programming yields the existence of an optimal deterministic policy for the Lagrangian-relaxed problem, {\it e.g.}, finite horizon optimality~\citep{puterman2014markov}. 
%     For the constrained problem, use the convex analysis / occupation measure formulation of CMDPs: 
%     the feasible set of occupation measures is compact and the objective and constraints are linear/continuous, implying the maximum is reached~\citep{dufour2012expected,altman2021constrained}.
% \end{proof}

\begin{theorem}\label{thm: lower-bound}
    \textbf{(Lower bound on expected generated tokens).}
    Fix a query $x$ and let $\mathcal{Y}_x$ denote the finite set of effective answers induced by the evaluator for $x$.
    Consider any interactive CLSR policy $\pi$ whose transcript $\mathscr{T}$ is generated token-by-token until a stopping action, and whose final answer is $\widehat{Y}=g(\mathscr{T},x)$.
    Suppose $\pi$ achieves target accuracy $\alpha\in(0,1)$:
    $\mathbb{P}_{\pi}\!\left[\widehat{Y}=Y\mid X=x\right]\geq \alpha$.
    % \begin{equation}
    %     \begin{gathered}
    %         \mathbb{P}_{\pi}\!\left[\widehat{Y}=Y\mid X=x\right]\geq \alpha .
    %     \end{gathered}
    % \end{equation}
    Let $\delta=1-\alpha$ and define the required information
    \begin{equation}
        \begin{gathered}
            I_{\mathrm{req}}(x,\delta)
            =
            H(Y\mid X=x)
            -
            h_2(\delta)
            -
            \delta\log_2\!\left(|\mathcal{Y}_x|-1\right),
        \end{gathered}
    \end{equation}
    where $h_2(\cdot)$ is the binary entropy.
    Let $\kappa_{\theta}(x)$ be an upper bound on the conditional information revealed by one \emph{active} generated token under the allowed LSF protocols:
    \begin{equation}
        \begin{gathered}
            \kappa_{\theta}(x)
            =
            \sup_{\pi,t,h_{<t}}
            I\!\left(Y;Z_t \mid X=x, Z_{<t}=h_{<t},\,|\mathscr{T}|\geq t\right),
        \end{gathered}
    \end{equation}
    where the supremum ranges over policies, time steps, and reachable active histories.
    Then
    \begin{equation}
        \begin{gathered}
            \mathbb{E}_{\pi}\!\left[|\mathscr{T}|\mid X=x\right]
            \geq
            \frac{\max\{I_{\mathrm{req}}(x,\delta),0\}}{\kappa_{\theta}(x)} .
        \end{gathered}
    \end{equation}
\end{theorem}

The bound separates three factors.
First, a higher target accuracy reduces $\delta$ and increases the Fano term, raising the information that the transcript must convey.
Second, harder queries have larger evaluator-induced uncertainty $H(Y\mid X=x)$ or larger effective answer classes $\mathcal{Y}_x$, increasing $I_{\mathrm{req}}$.
Third, stronger symbolic protocols increase $\kappa_{\theta}(x)$ by packing more task-relevant information into each generated token.
In this sense, CLSR improves efficiency not by violating an information limit but by changing the representation so that each emitted token carries more useful reasoning state.
The active-history definition of $\kappa_{\theta}(x)$ is important for adaptive stopping: a policy may continue only on difficult histories, so the per-token information rate must be conditionally bounded on the process still being active.

\begin{table*}[t]
  \caption{
    \textbf{Comparison with Token-reduction approaches on the accuracy--token trade-off).} 
    Inference accuracy (\%) and average completion tokens per problem for Raw CoT~\citep{wei2022chain}, 
    CCoT~\citep{nayab2024CCoT}, 
    CoD~\citep{xu2025CoD}, 
    SoT~\citep{aytes2025SoT}, and 
    CLSR (ours) across seven benchmarks and
    four backbones (LLaMA3-8B, DeepSeek-R1-Qwen3-8B, Qwen3-8B, and Qwen3-32B). 
    CLSR consistently improves the accuracy--token frontier, typically matching Raw CoT accuracy while reducing token usage by roughly $3/4$, and outperforming length-controlled prompting baselines at comparable budgets.
  }
  \label{table: full-eval}
  \begin{center}
    \begin{scriptsize}
      \begin{sc}
        \begin{tabular}{@{}cccccccccccccccc@{}}
        \toprule
         \multirow{2}{*}{LLMs} & \multirow{2}{*}{Methods} & \multicolumn{2}{c}{MMLU-pro} & \multicolumn{2}{c}{GPQA-main} & \multicolumn{2}{c}{GSM8K} & \multicolumn{2}{c}{MATH-500} & \multicolumn{2}{c}{AIME21-24} & \multicolumn{2}{c}{Sci-QA} & \multicolumn{2}{c}{Hotpot-QA} \\ \cmidrule(l){3-16} 
         &  & Acc & Tkn & Acc & Tkn & Acc & Tkn & Acc & Tkn & Acc & Tkn & Acc & Tkn & Acc & Tkn \\ \midrule
         \multirow{5}{*}{LLaMA3-8B} & Raw CoT & 38.6 & 960 & 30.4 & 1209 & 84.9 & 124 & 51.6 & 704 & 6.1 & 1705 & 61.3 & 51 & 45.1 & 32 \\ \cmidrule(r){2-16}
         & CoD & 35.8 & 352 & 26.4 & 386 & 80.5 & \textbf{51} & 42.2 & 280 & 4.9 & 1041 & 59.8 & 19 & 40.2 & 24 \\
         & CCoT & 37.2 & 370 & 29.3 & 450 & 83.6 & 65 & 45.6 & 315 & 5.8 & 1205 & 60.2 & 22 & 44.3 & 28 \\
         & SoT & 36.4 & 340 & 28.2 & 398 & 82.3 & 55 & 45.2 & 289 & 5.3 & 1051 & 61.2 & 19 & \textbf{45.8} & 21 \\
         & CLSR (ours) & \textbf{38.9} & \textbf{320} & \textbf{30.8} & \textbf{362} & \textbf{84.3} & 52 & \textbf{48.2} & \textbf{257} & \textbf{6.2} & \textbf{1035} & \textbf{61.4} & \textbf{18} & 45.5 & \textbf{14} \\ \midrule
        \multirow{5}{*}{\begin{tabular}[c]{@{}c@{}}DeepSeek-R1\\ -0528\\ -Qwen3-8B\end{tabular}} & Raw CoT & 71.1 & 1789 & 73.2 & 5380 & 92.2 & 433 & 91.2 & 3544 & 86.2 & 8190 & 71.3 & 125 & 55.8 & 213 \\ \cmidrule(r){2-16}
         & CoD & 68.3 & 582 & \textbf{73.2} & 1892 & 86.5 & 204 & 80.5 & 1842 & 77.4 & 3545 & 68.4 & 37 & 51.2 & 56 \\
         & CCoT & 64.5 & 516 & 70.4 & 1605 & 90.3 & 220 & 83.4 & 2032 & 79.6 & 3946 & 70.2 & 43 & 53.0 & 65 \\
         & SoT & 69.2 & 458 & 71.4 & 1536 & 91.8 & 205 & 84.1 & 1982 & 81.3 & 3266 & 71.0 & 41 & \textbf{54.8} & 59 \\
         & CLSR (ours) & \textbf{70.3} & \textbf{356} & 73.1 & \textbf{1326} & \textbf{92.4} & \textbf{198} & \textbf{88.1} & \textbf{1753} & \textbf{82.5} & \textbf{2979} & \textbf{71.5} & \textbf{30} & 54.5 & \textbf{43} \\ \midrule
         % \multirow{5}{*}{Qwen3-4B} & Raw CoT & 54.7 & 363 & 45.9 & 1247 & 89.0 & 178 & 84.8 & 808 & 35.9 & 3854 & 76.4 & 94 & 53.8 & 65 \\ \cmidrule(r){2-16}
         % & CoD & 52.1 & 130 & 41.2 & 410 & 83.2 & \textbf{80} & 70.8 & 356 & 33.2 & 2650 & 74.2 & 35 & 50.2 & 33 \\
         % & CCoT & 45.6 & 135 & 40.3 & 420 & 85.4 & 95 & 73.2 & 380 & 34.7 & 2762 & 74.5 & 45 & 52.2 & 37 \\
         % & SoT & 52.2 & 127 & 42.6 & 408 & 86.4 & 89 & 74.4 & 351 & 34.9 & 2637 & 76.0 & \textbf{32} & 53.2 & 34 \\
         % & CLSR (ours) & \textbf{55.2} & \textbf{113} & \textbf{45.1} & \textbf{329} & \textbf{90.3} & 93 & \textbf{83.2} & \textbf{273} & \textbf{35.7} & \textbf{1970} & \textbf{77.0} & 39 & \textbf{54.5} & \textbf{24} \\ \midrule
        \multirow{5}{*}{Qwen3-8B} & Raw CoT & 60.2 & 276 & 49.1 & 1085 & 90.9 & 243 & 87.2 & 878 & 46.1 & 4234 & 77.8 & 75 & 66.4 & 134 \\ \cmidrule(r){2-16}
         & CoD & 55.6 & 125 & 42.7 & 282 & 85.4 & 100 & 74.5 & 389 & 30.2 & 2610 & 76.2 & 46 & 63.2 & 58 \\
         & CCoT & 50.2 & 135 & 41.3 & 245 & 89.2 & 125 & 78.4 & 423 & 36.7 & 2805 & 78.2 & 49 & 65.2 & 70 \\
         & SoT & 58.5 & 118 & 45.2 & 228 & 90.3 & 107 & 81.3 & 294 & 38.0 & 2637 & \textbf{79.3} & 45 & 65.8 & 63 \\
         & CLSR (ours) & \textbf{60.4} & \textbf{96} & \textbf{47.7} & \textbf{228} & \textbf{91.2} & \textbf{89} & \textbf{86.8} & \textbf{257} & \textbf{43.7} & \textbf{2361} & 77.8 & \textbf{35} & \textbf{66.4} & \textbf{48} \\ \midrule
         \multirow{5}{*}{Qwen3-32B} & Raw CoT & 67.5 & 405 & 54.2 & 983 & 91.3 & 216 & 89.3 & 845 & 46.8 & 4572 & 79.2 & 81 & 68.4 & 165 \\ \cmidrule(r){2-16}
         & CoD & 64.2 & 318 & 49.2 & 276 & 82.1 & 105 & 81.5 & 365 & 32.5 & 2912 & 77.0 & 53 & 64.6 & 78\\
         & CCoT & 65.1 & 192 & 50.8 & 312 & 88.2 & 94 & 82.0 & 382 & 37.3 & 3025 & 79.1 & 56 & 65.8 & 81\\
         & SoT & 64.3 & 134 & 51.4 & 247 & 90.5 & 112 & 83.8 & 278 & 38.5 & 2453 & 78.9 & 48& 66.2 & 72\\
         & CLSR (ours) & \textbf{68.1} & \textbf{118} & \textbf{54.0} & \textbf{209} & \textbf{91.6} & \textbf{91} & \textbf{89.4} & \textbf{206} & \textbf{45.2} & \textbf{2038} & \textbf{80.2} & \textbf{39} & \textbf{68.8} & \textbf{55}\\ \bottomrule
        \end{tabular}
      \end{sc}
    \end{scriptsize}
  \end{center}
  \vskip -0.2in
\end{table*}

\subsection{A Sociolinguistic Lens on LSF Evolution}

The empirical behavior of LSFs has a useful sociolinguistic interpretation.
CLSR does not create language from an empty prior: the base LLM has already internalized human linguistic conventions, mathematical notation, domain-specific abbreviations, and tokenizer-specific regularities.
Evolved LSFs are therefore best understood as \emph{machine-oriented languages}: they are constrained by pretrained linguistic priors, but selected under a new objective that rewards correctness per generated token.

This lens explains three recurring patterns.
First, successful LSFs follow a principle of least effort: they remove low-information narrative text while preserving symbols that carry reusable reasoning states, such as variable bindings, subgoal markers, verification tags, and short operators.
Second, evolution induces conventionalization.
A symbol survives not because it is intrinsically meaningful to humans, but because it repeatedly supports reliable inference for the model population.
Third, routing implements pragmatic code-switching.
For easy queries, a strict and terse LSF often suffices; 
for ambiguous or difficult queries, the router may select a more redundant LSF, aggregate multiple LSFs, or invoke additional rounds for verification.

Accordingly, CLSR should be viewed as a test-time mechanism for studying how LLM agents invent, select, and reuse compact social conventions under an explicit communication bottleneck, rather than as a claim of human-independent language emergence.
%This CLSR framework should be considered a test-time mechanism for studying how black-box LLM agents invent and use compact social conventions under a communication bottleneck.
Appendix~\ref{app-subsec: zipf_iterated} expands this connection to least-effort and iterated-learning principles.

% \subsection{A Sociolinguistic Lens on LSF Evolution}

% The empirical behavior of LSFs has a useful sociolinguistic interpretation.
% CLSR does not create language from an empty prior: 
% the base LLM has already internalized human linguistic conventions, mathematical notation, and tokenizer-specific regularities.
% Therefore, evolved LSFs are best understood as \emph{machine-oriented dialects} that are constrained by pretrained priors but selected by a new pressure: 
% solve the task with fewer generated tokens.
% This explains why many successful LSFs become partially human-interpretable rather than arbitrary strings.
% They often retain compact mathematical symbols, subgoal markers, verification tags, and short control tokens because these conventions are already easy for the model to manipulate reliably.

% This view also clarifies why evolution helps.
% Across generations, LSFs undergo a process analogous to conventionalization: 
% symbols that repeatedly support correct and concise inference survive, while brittle or overly terse conventions are discarded.
% The router then performs a form of pragmatic code-switching.
% For easy queries, it selects a strict low-cost dialect; for ambiguous or difficult queries, it composes several dialects or invokes additional rounds for verification.
% In this sense, CLSR is not merely a token-compression trick but a test-time mechanism for forming and selecting compact social conventions among LLM agents.
% Appendix~\ref{app-subsec: zipf_iterated} expands this connection to the principles of least-effort and iterated-learning.

\subsection{Universality and Finite Programs}
\label{subsec: Universality and Finite Programs}

The CLSR protocol class can be viewed as a programmable computation system: 
the router implements control flow (which ``expert language'' speaks next and when to stop), 
each LSF implements an instructive symbolic coding regime, 
the transcript $\mathscr{T}$ is writable memory, and aggregation is an arbitrary computable readout. 
Next, we demonstrate that the multi-round multi-LSF interaction is a natural generalization of the ``generate-a-program then execute'' paradigms.
Consider two families of inference procedures:
\begin{enumerate}[noitemsep, nolistsep]
    \item \textbf{Program-Execution Pipelines (PE).}
    A well-formed PE procedure makes finite calls to the base LLM and obtains an LLM-generated transcript $\mathscr{T}\in\Sigma$ ({\it e.g.}, code/program plus any intermediate strings).
    Then it applies a deterministic logic-only executor $\text{Exec}: \Sigma\mapsto\mathcal{A}$ that does not access external knowledge.
    %, {\it i.e.}, $\text{Exec}$ is a fixed computable function of the transcript, and outputs $\hat{y}=\text{Exec}(T)\in\mathcal{A}$, where $\mathcal{A}$ is the set of admissible strings.
    \item \textbf{LSF-only CLSR Protocols.}
    A policy $\pi$ is a multi-round multi-LSF protocol that can adaptively choose LSFs and stopping rules but does not invoke any external executor.
    It outputs $\hat{y}$ solely from LLM generations.
\end{enumerate}

\textbf{Premise (Interpreter Realizability).}
For every deterministic computable executor $\mathrm{Exec}:\Sigma^*\mapsto\mathcal{A}$ in the allowed executor class, there exists an LSF $S_{\mathrm{Exec}}$ such that, for any input string $s\in\Sigma^*$, one LLM call conditioned on $S_{\mathrm{Exec}}$ and $s$ produces $\mathrm{Exec}(s)$ with failure probability at most $\varepsilon_{\mathrm{int}}<1/2$ and with output length at most $|\mathrm{Exec}(s)|+c_0$ for a constant $c_0$.
This is an idealized internal-simulation premise.
It is supported in principle by universality results for Transformer-like architectures~\citep{perez2019turing,perez2021attention}, but it is not automatically guaranteed for any finite pretrained checkpoint.

\begin{theorem}\label{thm: program-exec}
    \textbf{(Conditional subsumption of program execution).}
    Under interpreter realizability, consider any program-execution pipeline PE that achieves accuracy of at least $\alpha$ with an expected LLM-generated token cost of $B$.
    Let $\bar{\ell}_{\mathrm{Exec}}(x)$ denote the expected length of the executor output plus the constant formatting overhead under the PE transcript distribution.
    Then for any $\delta\in(0,1)$, there exists an LSF-only CLSR protocol $\pi$ such that
    \begin{itemize}[nolistsep,noitemsep]
        \item \textbf{Accuracy}: $\mathbb{P}_{\pi}[\widehat{Y}=Y\mid X=x]\geq \alpha-\delta$;
        \item \textbf{Token cost}: $J_C(\pi)\leq B+O\!\left(\bar{\ell}_{\mathrm{Exec}}(x)\log(1/\delta)\right)$.
    \end{itemize}
    In short-answer reasoning benchmarks where $\bar{\ell}_{\mathrm{Exec}}(x)=O(1)$, this reduces to $J_C(\pi)\leq B+O(\log(1/\delta))$.
    In the exact-interpreter and bounded-output regime, the achievable token--accuracy region of CLSR contains that of PE up to an arbitrarily small consensus overhead.
\end{theorem}

% \textbf{Premise (Interpreter Realizability).}
% For every deterministic computable $\text{Exec}:\Sigma\mapsto\mathcal{A}$ in the allowed executor class, there exists an LSF $\mathcal{S}_{\text{Exec}}$ such that, for any input string $s\in\Sigma$, one call to the base LLM conditioned on $\mathcal{S}_{\text{Exec}}$ and $s$ produces $\text{Exec}(s)$ with failure probability at most $\varepsilon_{int}$ and with output length $\lvert\text{Exec}(s)+O(1)\rvert$.
% This is an idealized ``internal simulation'' premise;
% it is supported in principle by universality results for Transformer-like architectures~\citep{perez2019turing,perez2021attention}, but is not automatically guaranteed for any finite pretrained checkpoint.

% \begin{theorem}\label{thm: program-exec}
%     Under interpreter realizability, for any program-execution pipeline PE that achieving precision $\alpha$ at the expected token cost $B$, there exists an LSF-only CLSR protocol $\pi$ such that
%     \begin{itemize}[nolistsep,noitemsep]
%         \item \textbf{Accuracy}: $\text{Acc}(\pi)\geq \alpha-\varepsilon_{int}$ and can be increased to more than $\alpha -\delta$ for any $\delta>0$ by standard repetition or consensus at an $O(\log (1/\delta))$ token overhead;
%         \item \textbf{Token cost}: $J_C(\pi)\leq B+O(\log (1/\delta))$.
%     \end{itemize}
%     In particular, in the exact-interpreter limit $\varepsilon_{int}=0$, the achievable token-accuracy region of CLSR contains that of PE;
%     hence, the CLSR's optimality is no worse than that of external program execution.
% \end{theorem}

\textbf{When the assumption holds or fails.}
Interpreter realizability is plausible when: 
(i) $\mathrm{Exec}$ is within the base model's algorithmic competence (e.g., simple arithmetic, bounded symbolic manipulation), 
(ii) the required execution trace fits within the model's effective context/attention constraints, and 
(iii) decoding is sufficiently reliable (or can be stabilized by self-consistency). 
It may fail for executors that require an exact long-horizon computation or strict formal guaranties.

% \begin{figure*}[t]
% \centering
%     \subfloat[Inference Accuracy.]
%     {
%         \includegraphics[width=2.0in]{images/EVvsAcc_gpqa_v0.png}%
%         \label{sub_fig: EVvsAcc_gpqa}
%     }
%     \hfil\hfil
%     \subfloat[Number of generated tokens.]
%     {
%         \includegraphics[width=2.0in]{images/EVvsNT_gpqa_v0.png}%
%         \label{sub_fig: EVvsNT_gpqa}
%     }
%     \hfil\hfil
%     \subfloat[Ratio of Accuracy and \#tokens.]
%     {
%         \includegraphics[width=2.0in]{images/EVvsRatio_gpqa_v0.png}%
%         \label{sub_fig: EVvsRatio_gpqa}
%     }
%     \caption{
%     \textbf{Effect of evolution depth on the accuracy--token frontier.}
%     We vary the number of language-evolution iterations used to refine the LSF pool, while holding the backbone fixed.
%     $N$ denotes the number of training exemplars used for LSF synthesis.
%     We report (a) test accuracy, (b) average generated completion tokens per problem, and (c) an aggregated efficiency score.
%     Increasing the evolution depth improves both accuracy and token efficiency and then saturates, consistent with the view that better-evolved LSF dialects increase the effective per-token information rate predicted by Theorem~\ref{thm: lower-bound}.
%     }
%     \label{fig: varying evolution}
% \vskip -0.1in
% \end{figure*}

\begin{figure*}[t]
\centering
    \subfloat[Scientific benchmarks.]
    {
        \includegraphics[width=1.9in]{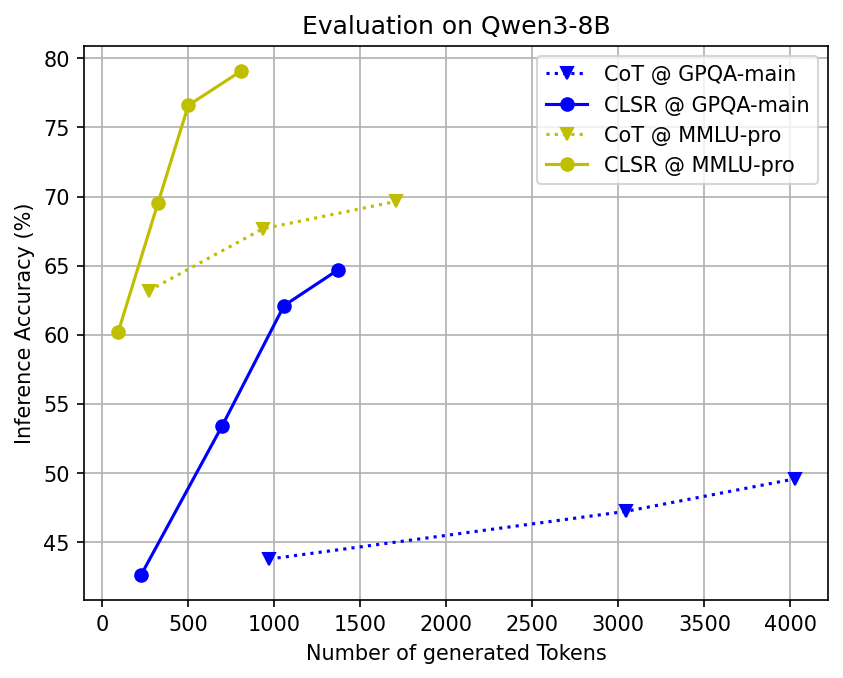}%
        \label{sub_fig: ScaleNT_gpqa_mmlu}
    }
    \hfil\hfil
    \subfloat[Mathematical benchmarks.]
    {
        \includegraphics[width=1.9in]{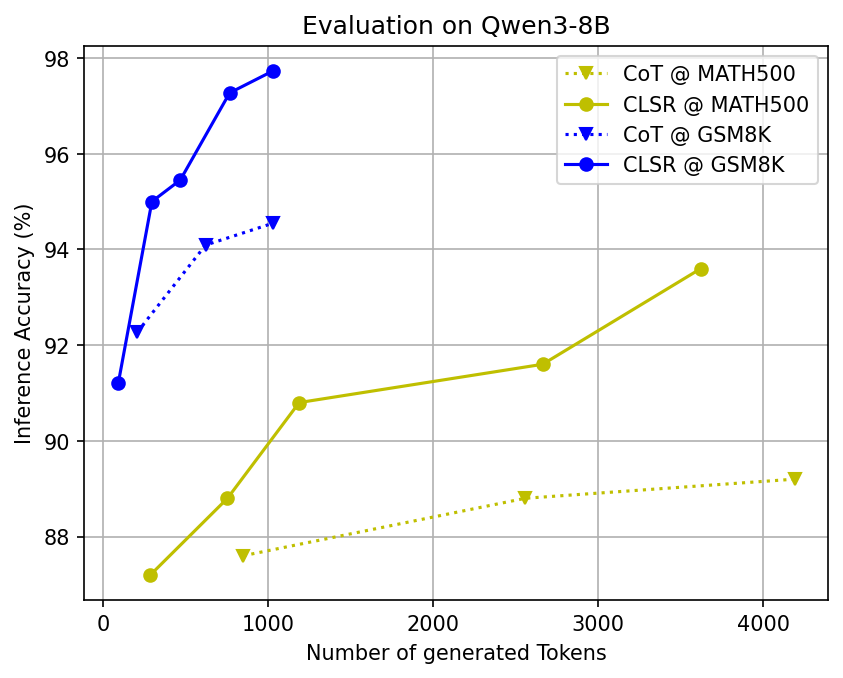}%
        \label{sub_fig: ScaleNT_math_gsm8k}
    }
    \hfil\hfil
    \subfloat[Hard math (AIME21-24).]
    {
        \includegraphics[width=1.9in]{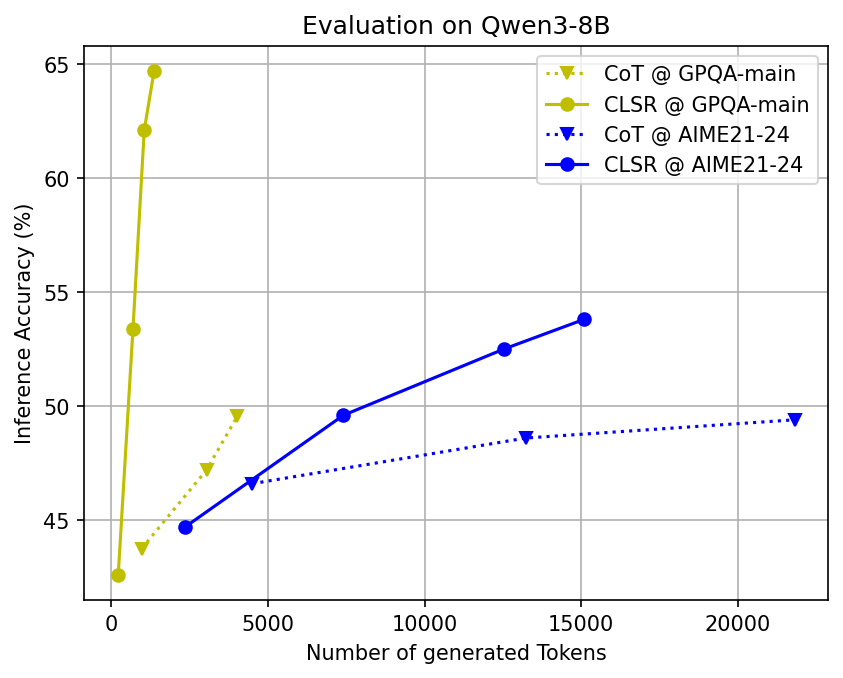}%
        \label{sub_fig: ScaleNT_gpqa_aime}
    }
    \caption{
    \textbf{Scaling generated tokens: CLSR dominates CoT under test-time scaling.}
    We increase the number of generated tokens via test-time scaling (multiple samples with majority vote) and compare standard CoT against CLSR (LSF).
    The reported curves trace the empirical accuracy--token trade-off.
    CLSR achieves higher accuracy at a given token budget (or the same accuracy at fewer tokens) across (a) scientific QA benchmarks,
    (b) math-reasoning benchmarks, and (c) hard math (AIME21--24), supporting our theoretical framing that structured symbolic traces improve the information carried per generated token.
    }
    \label{fig: scaling generated tokens}
\vskip -0.1in
\end{figure*}

\begin{figure*}[t]
\centering
    \subfloat[\#Agents @ MMLU-pro.]
    {
        \includegraphics[width=1.6in]{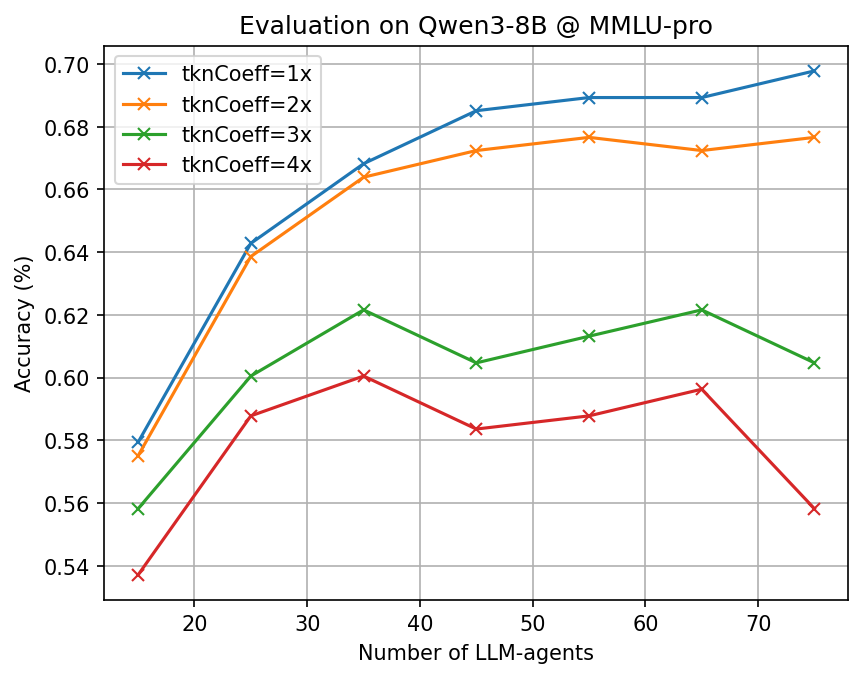}%
        \label{sub_fig: NAvsAcc_mmlu}
    }
    \hfil\hfil
    \subfloat[\#Tokens @ MMLU-pro.]
    {
        \includegraphics[width=1.6in]{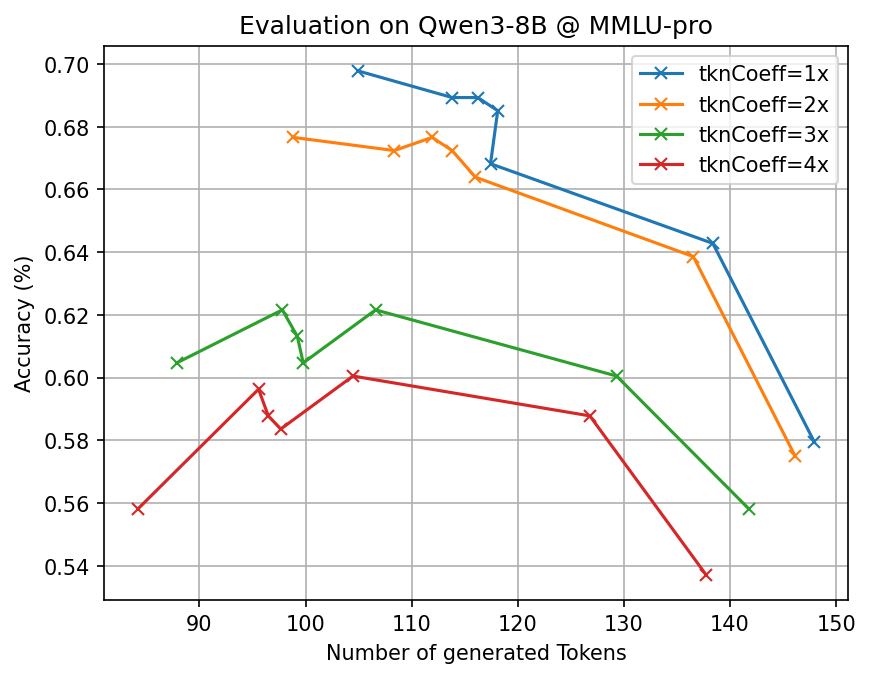}%
        \label{sub_fig: NTvsAcc_mmlu}
    }
    \hfil\hfil
    \subfloat[\#Agents @ GPQA.]
    {
        \includegraphics[width=1.6in]{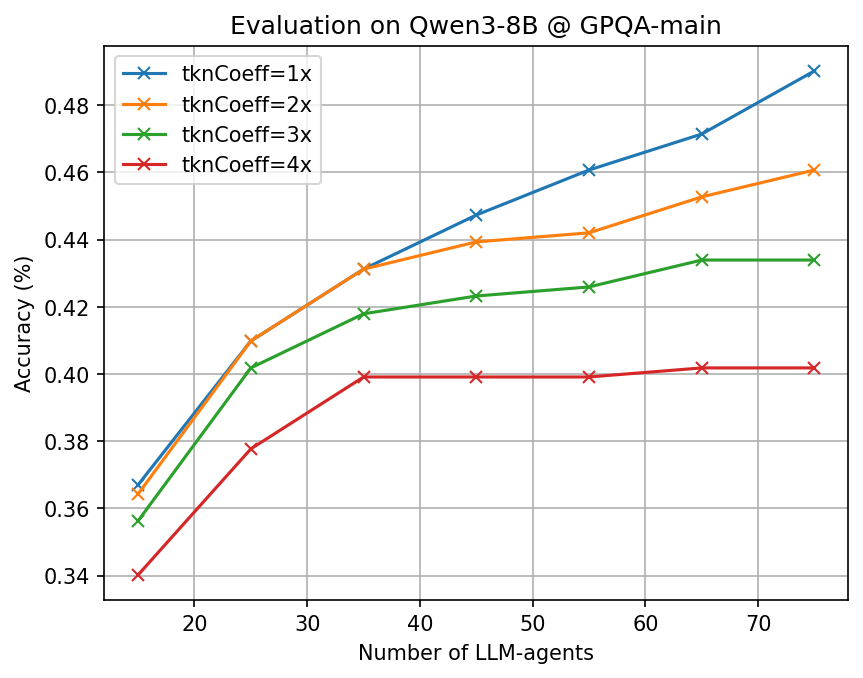}%
        \label{sub_fig: NAvsAcc_gpqa}
    }
    \hfil\hfil
    \subfloat[\#Tokens @ GPQA.]
    {
        \includegraphics[width=1.6in]{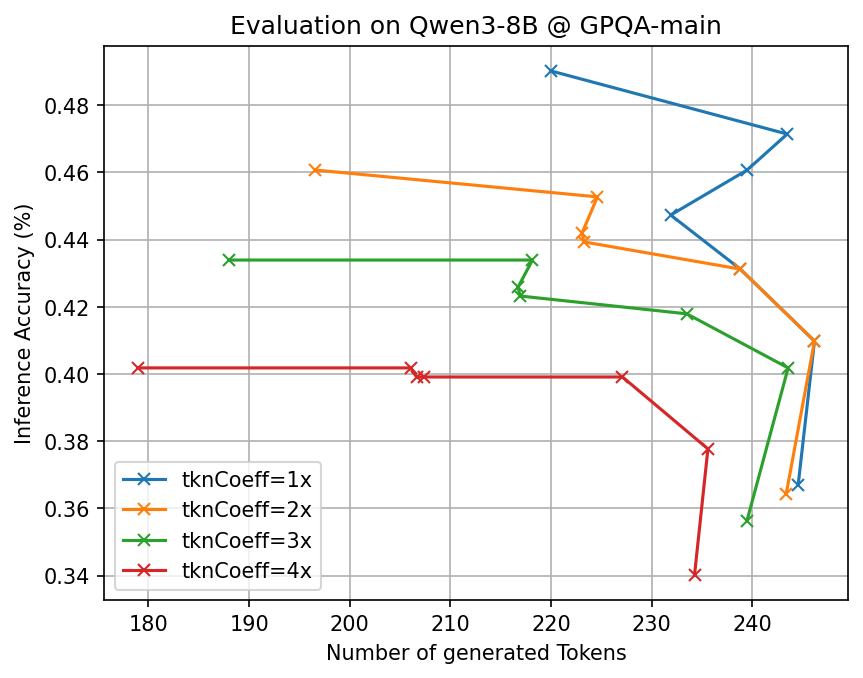}%
        \label{sub_fig: NTvsAcc_gpqa}
    }
    \caption{
    \textbf{Scaling evolution agents improves LSF quality.}
    We ablate the number of LLM agents participating in the offline LSF-evolution process (parallel proposal, critique, and refinement),
    and evaluate the resulting LSF pools on representative benchmarks.
    Across tasks, using more agents typically improves downstream accuracy, while also increasing (or modestly changing) generation cost.
    }
    \label{fig: adding LLM-agents}
\vskip -0.1in
\end{figure*}

\section{Experiments}

\subsection{Experimental Setup}

\textbf{Benchmarks.}
We evaluate symbolic-language reasoning on seven widely used reasoning benchmarks that span knowledge-intensive QA, multi-hop retrieval-style QA, and mathematical problem solving: 
(i) broad-domain factual and professional reasoning (MMLU-Pro~\citep{wang2024mmlu}), 
(ii) expert-level science QA with strong adversarial difficulty (GPQA~\citep{rein2024gpqa}), 
(iii) grade-school multi-step arithmetic (GSM8K), 
(iv) competition-style proof and derivation (MATH500~\citep{hendrycks2021math500} and AIME (21-24)~\citep{aime_1983_2024}), 
(v) science QA with short final answers (ScienceQA~\citep{lu2022sciqa}), 
and (vi) multi-hop question answering (HotpotQA~\citep{yang2018hotpotqa}).

\textbf{Backbone LLM-agents.}
To test generality across model families and scales, we run our pipeline with multiple open LLM backbones, treating each checkpoint as an independent agent in language creation and/or inference. 
We report the results for 
Qwen3-8B/32B~\citep{yang2025qwen3}, LLaMA3-8B~\citep{llama3modelcard}, 
and a distilled variant of DeepSeek-R1 (0528-Qwen3-8B)~\citep{deepseekai2025deepseekr1}.
Throughout, the backbone weights remain \emph{frozen}; improvements come purely from test-time symbolic protocols and routing.

\textbf{Baselines.}
We compare test-time reasoning and token-reduction prompting strategies:
(i) Raw CoT (standard chain-of-thought prompting)~\citep{wei2022chain},
(ii) CoD (Chain-of-Draft)~\citep{xu2025CoD},
(iii) CCoT (Constrained CoT)~\citep{nayab2024CCoT}, and
(iv) SoT (Sketch-of-Thought)~\citep{aytes2025SoT}.
We also compare program-based strategies:
(v) PoT (Program-of-Thoughts prompting)~\citep{chen2023program} and
(vi) PAL (Program-aided Language)~\citep{gao2023pal}.
Additionally, we compare prompt optimization methods:
(vii) Plan-to-Solve~\citep{wang2023plansolve} and
(viii) PromptBreeder~\citep{fernando2024promptbreeder}.
All methods are evaluated on the same benchmark splits.
%and the same backbone LLM for a fair comparison.

\textbf{Metrics (accuracy and token cost).}
For each benchmark, we report:
(i) inference accuracy (\%) and
(ii) the number of generated tokens emitted by the inference LLM and the LLM-router.
%(ii) the number of tokens generated emitted by the LLM inference and the LLM-router.
For CLSR, this includes all tokens generated online in selected LSFs, router plans, intermediate rounds, and aggregation responses, which correspond to the cost definition in Section~\ref{subsec: Problem setup and metrics}.
%When the router triggers on-the-fly LSF recombination during inference, its generated tokens are also counted.
We use generated tokens as the main latency-oriented metric because the decode phase dominates many reasoning workloads, whereas prompt-prefix costs are highly serving-dependent.
To explicitly address the input-overhead issue, Appendix~\ref{app:latency_token_accounting} reports a cache-aware token-equivalent metric that includes reusable LSF-card prefixes and separates uncached input, cached input, and output-token coefficients.

% \textbf{Metrics (accuracy and token cost).}
% For each benchmark, we report:
% (i) Inference accuracy (\%) and
% (ii) the number of tokens generated by the inference LLM and the LLM-router.
% %produced by the model during inference.
% %For our method, 
% The token cost is measured as the total number of generated tokens in all invoked symbolic protocols and in all communication rounds.
% %, consistent with the cost definition in Eq.~\ref{}.
% When the router triggers on-the-fly synthesis of a new LSF during inference (a rare event; see Appendix~\ref{app-abl: fixed T or not}).
% Lower token usage indicates greater test-time efficiency since the prefill phase is negligible compared with the decode phase in our cases (see Appendix~\ref{app:latency_token_accounting}).
% %, we also include the tokens spent to generate that LSF for a fair comparison.
% %, so the reported cost faithfully reflects end-to-end generation.

\subsection{Implementation Details}

\textbf{LSF synthesis from exemplars.}
For each benchmark, we sample $N\in[200,2000]$ training exemplars that include ground-truth reasoning content.
Given these exemplars in context, we prompt the LLM backbone to invent an initial population of hundreds of Language-Symbolism Frameworks (LSFs), each specifying
(i) a compact lexicon,
(ii) a constrained grammar,
and (iii) usage constraints that aim to reduce token footprint while preserving reasoning capability.
%The choice of $100$ is an empirical exploration/compute trade-off rather than a theoretically privileged value; 
Appendix~\ref{app-subsec: population_size} reports a population-size ablation.
We use temperature $0.9$ to generate LSFs and temperature $0.3$ to evaluate the inference.

\textbf{Evolutionary bootstrapping (generations).}
We iteratively refine the LSF population via an evolutionary loop.
At each evolution step, we evaluate candidate LSFs on a held-out validation set, then select high-leverage inference traces that are simultaneously correct and token-efficient.
These selected traces are fed back to the LLM to synthesize the next-generation LSF population.
We use the fifth evolution LSF pool for the final test evaluation by default.

%\paragraph{Cross-benchmark language pool and routing.}
At inference time, we maintain a global pool by mixing LSFs evolved from different benchmarks.
Given a test query, the router may 
(i) select a single existing LSF, which is typical for easier queries,
(ii) select and aggregate multiple LSFs for moderately difficult queries,
or
(iii) compose multiple existing LSFs across rounds for hard cases.
% (i) select a single existing LSF (typical for easy queries), 
% (ii) select and aggregate multiple LSFs (medium difficulty), 
% or (iii) rarely synthesize a new LSF by recombining existing ones (hard corner cases).
By default, the selection objective weights token length and accuracy equally;
the number of tokens of an LSF card ranges from 500--2000 depending on task difficulty and protocol depth.
We use the LLM-router to determine the number of rounds $T$ required for the test problem.
We can also fix $T{=}\{1,3\}$ rather than letting it be determined by the router;
see Table~\ref{table: varying no.round T} in Appendix~\ref{app-sec: more empirical results} for more ablation studies.

\textbf{System and compute.}
All experiments are conducted on 8$\times$ NVIDIA RTX 4090 GPUs.
Generating and evolving LSF populations takes roughly several days in total.
This is a one-time offline protocol-discovery cost, analogous to the pre-deployment search cost in automatic prompt/protocol optimization methods.
CLSR is intended for settings where the evolved LSF pool is reused across many downstream queries, so the offline cost can be amortized over serving-time inference.
The online routing/LSF-selection stage takes $<1$ seconds per query and is small relative to LLM decoding.
See Appendix~\ref{app-subsec: offline_amortization} for details.
%provides a more explicit discussion of amortization.

% \textbf{System and compute.}
% All experiments are conducted on 8$\times$ NVIDIA RTX 4090 GPUs.
% Generating and evolving LSF populations
% %for all seven benchmarks 
% takes roughly 2--3 weeks in total.
% The routing/LSF-selection stage takes $<1$ seconds per query and is negligible relative to LLM decoding for inference.

\subsection{Main Results}
\label{subsec: main results}

Tables~\ref{table: full-eval}-\ref{table: varying no.round T} show the central empirical pattern: CLSR consistently improves the accuracy--token frontier across backbones and benchmarks.
Compared with Raw CoT, CLSR uses substantially fewer completion tokens while preserving, and in many cases slightly improving, accuracy.
Compared with direct token-reduction prompting baselines, our CLSR is more stable on the Pareto frontier: aggressive compression baselines often save tokens by sacrificing correctness, whereas CLSR retains a reusable symbolic protocol and routes among LSFs according to query difficulty.

The gains are not uniform token shortening; they reflect adaptive allocation of reasoning effort.
On knowledge-intensive QA benchmarks such as MMLU-Pro and GPQA, much of the natural-language rationale can be compressed into compact symbolic state without a large accuracy loss.
On deliberation-heavy mathematical benchmarks such as MATH500 and AIME, the router often benefits from selecting stricter LSFs or invoking additional rounds, because the task requires multi-step transformation and verification.
For shorter-answer datasets, the absolute completion length is already small, so the token savings are naturally more limited, but CLSR remains competitive in accuracy.

The figures further explain the mechanism behind the aggregate results.
Figure~\ref{fig: scaling generated tokens} shows that CLSR scales more favorably with additional test-time tokens than standard CoT, indicating that structured symbolic traces use extra budget more efficiently than repeated natural-language sampling.
Figure~\ref{fig: adding LLM-agents} shows that increasing the number of LSF-generating agents improves the final language pool, consistent with broader LSF exploration and stronger selection pressure.
Figure~\ref{fig: varying evolution} shows that deeper evolution improves the accuracy--token ratio before saturation, suggesting that the evolutionary loop is not merely shortening outputs but selecting conventions that preserve more task-relevant information per generated token.

Finally, Table~\ref{table: compare-program-based} compares CLSR with program-execution and prompt-optimization baselines under matched settings.
CLSR remains competitive without relying on an external executor, and it also improves upon prompt-optimization baselines.
These results support the Section~\ref{sec: theory}: 
CLSR is not a fixed brevity heuristic, but an adaptive policy over the accuracy--token frontier whose multi-round LSF protocols can approximate program-like computation when the required symbolic operations remain within the model's internal competence.

\begin{table}[t]
  \caption{
  \textbf{Comparison with program-execution and prompt optimization methods on Qwen3-8B.}
  %We report inference accuracy ($\Uparrow$) and the number of generated tokens ($\Downarrow$) on GSM8K and MATH500.
  We compare CLSR with two program-execution pipelines: PoT (Program-of-Thoughts) and PAL (Program-aided Language), 
  and two prompt optimization methods: P2S (Plan-to-Solve) and PBrd (PromptBreeder).
  For CLSR, $T$ denotes the number of rounds.
  %, {\it i.e.}, independent responses aggregated by majority vote.
  For PoT and PAL, the reported tokens count only tokens used to generate the program.
  }
  \label{table: compare-program-based}
  \begin{center}
    \begin{scriptsize}
      \begin{sc}
        \begin{tabular}{@{}ccccccc@{}}
        \toprule
          & \multicolumn{2}{c}{GSM8K} 
          & \multicolumn{2}{c}{MATH500}
          & \multicolumn{2}{c}{GPQA} \\ \cmidrule(r){2-7}
        Methods & Acc & Tkn & Acc & Tkn & Acc & Tkn \\ \midrule
        CoT & 91.6 & 248 & 87.8 & 905 & 43.8 & 973 \\
        PoT & 92.1 & 113 & 88.1 & 375 & / & /\\
        PAL & 92.5 & 148 & 88.6 & 422 & / & /\\ \midrule
        P2S & 92.2 & 316 & 88.9 & 1205 & 45.2 & 1085 \\
        PBrd & 93.6 & 235 & 88.4 & 843 & 46.1 & 922 \\ \midrule
        CLSR & 92.8 & 90 & 86.8 & 257 & 42.6 & 228 \\
        CLSR ($T{=}1$) & 92.1 & 83 & 86.2 & 134 & 40.9 & 182\\
        CLSR ($T{=}3$) & \textbf{94.8} & 214 & \textbf{89.7} & 417 & \textbf{49.2} & 565 \\ \bottomrule
        \end{tabular}
      \end{sc}
    \end{scriptsize}
  \end{center}
  \vskip -0.2in
\end{table}

% \paragraph{Task-dependent behavior.}
% We observe a consistent pattern across model families:
% (i) on knowledge-intensive QA (MMLU-Pro, GPQA), symbolic protocols provide large efficiency gains with minimal accuracy degradation, suggesting that a substantial fraction of natural-language CoT verbosity is not strictly necessary for correctness;
% (ii) in mathematical benchmarks (MATH500 and AIME), aggressive compression can sometimes reduce accuracy for weaker symbolic dialects, but the method still yields strong Pareto improvements because token savings are large, especially for stronger backbones whose Raw CoT traces are extremely long;
% (iii) in benchmarks where the final answer is short (e.g., HotpotQA / ScienceQA in our configuration), absolute token counts are already small, so improvements in token usage saturate, while CLSR still maintains competitive accuracy.

\subsection{Ablations and Analysis}

We ablate the following design choices:
(i) evolution depth (Sec.~\ref{app-abl: Effect of evolution depth}),
(ii) exemplar count (Sec.~\ref{app-abl: Effect of exemplar count}),
(iii) agent count and length-coefficient (Sec.~\ref{app-abl: Effect of number of agents and length-coefficient}),
(iv) number of multi-rounds $T$ (Sec.~\ref{app-abl: fixed T or not}),
(v) cross-LLM transfer via swapping the LSF generator (Sec.~\ref{app-abl: Effect of LSF-generator}),
and
(vi) qualitative examples (Sec.~\ref{app-subsec: Qualitative examples}).
We also report robustness studies in Appendix~\ref{app-sec: robustness_checks}, including seed stability, population-size sensitivity, larger-model inference, long-context pilot evaluation, category/full-pool routing, cross-domain transfer, and cache-aware token accounting.

%Together, these ablation studies support the claim that CLSR improves token efficiency by discovering and routing among reusable symbolic codes, rather than by relying on a single lucky prompt, a rigid hand-defined category taxonomy, or an uncounted router overhead.

% \subsection{Ablations and Analysis}

% We ablate the following design choices:
% (i) Effect of evolution depth (Sec.~\ref{app-abl: Effect of evolution depth}),
% (ii) Effect of exemplar count (Sec.~\ref{app-abl: Effect of exemplar count}),
% (iii) Effect of number of agents and length-coefficient (Sec.~\ref{app-abl: Effect of number of agents and length-coefficient}), 
% (iv) Number of multi-rounds $T$ (Sec.~\ref{app-abl: fixed T or not}), 
% (v) Cross-LLM transfer via swapping LSF-generator (Sec.~\ref{app-abl: Effect of LSF-generator}), and
% (vi) Qualitative examples (Sec.~\ref{app-subsec: Qualitative examples}).
% The ablation studies validate that 
% CLSR improves token efficiency by discovering and routing among symbolic codes and the resulting behavior is consistent with the Pareto-optimal characterization in Section~\ref{sec: theory}.
%that increase the per-token information rate, 
%and the resulting behavior is consistent with the Pareto-optimal policy characterization in Section~\ref{sec: theory}.

\section{Conclusion and Limitations}

We introduced CLSR, a test-time framework that lets LLMs invent, evolve, and route among compact symbolic communication protocols.
Empirically, CLSR improves the accuracy--efficiency frontier across diverse reasoning benchmarks, with robustness checks showing stable gains.
Conceptually, CLSR suggests that LLMs can develop machine-oriented languages that are not manually crafted by humans.
%This provides a practical bridge between token-efficient reasoning, emerging communication, and sociolinguistic conventionalization.
However, the offline language-evolution can be computationally expensive,
and the synthesis pipeline requires training exemplars with reasoning content. 
Future work should study how symbolic languages transfer across modalities, tools, and multi-agent environments.

% \section{Conclusion and limitations}

% % We studied whether LLMs can move beyond verbose natural-language chain-of-thought and instead develop \emph{machine-efficient} symbolic communication systems for reasoning. 
% We introduce CLSR (Communicative Language Symbolism Routing), which adaptively selects and composes languages at test time to trade tokens for accuracy. 
% Empirically, CLSR consistently improves the accuracy--efficiency frontier, improving accuracy while using substantially fewer generated tokens. 
% However, the offline language-evolution can be computationally expensive; 
% and the synthesis pipeline requires training exemplars with reasoning content. 
% Future work should measure monetary cost under real serving stacks and robustness under adversarial or long-context settings.
% %, as well as study whether symbolic conventions transfer to tool-using agents and multi-modal reasoning.

\clearpage
\section*{Acknowledgements}
This work was supported in part by the National Key R\&D Program of China under Grant 2023YFC2508704, 
in part by the National Natural Science Foundation of China under grant number 62236008,
in part by the Natural Science Foundation of Beijing under grant number L251082,
and in part by Shandong Provincial Natural Science Foundation under project ZR2025ZD01.

% \textbf{Do not} include acknowledgements in the initial version of the paper

% If a paper is accepted, the final camera-ready version can (and usually should)
% include acknowledgements.  Such acknowledgements should be placed at the end of
% the section, in an unnumbered section that does not count towards the paper
% page limit. Typically, this will include thanks to reviewers who gave useful
% comments, to colleagues who contributed to the ideas, and to funding agencies
% and corporate sponsors that provided financial support.

\section*{Impact Statement}
This work aims to improve the efficiency of LLM reasoning by reducing unnecessary generated tokens while preserving task performance.
The main potential risk is that compact symbolic traces may be less directly interpretable than natural-language rationales, which could make human auditing more difficult in real-world deployments.
A practical mitigation is to use CLSR with larger LLMs for internal reasoning while requiring the final system to output a human-readable explanation using smaller LLMs.
%, or to log both the selected LSF protocol and a translated natural-language trace when interpretability is required.
We do not introduce new datasets containing personally identifiable information or new model-training procedures beyond black-box LLM prompting and evaluation.
%There are NO potential societal consequences of our work which we feel must be specifically highlighted here.

% This paper presents work whose goal is to advance the field of Machine Learning. 
% There are NO potential societal consequences of our work which we feel must be specifically highlighted here.


% \clearpage
\bibliography{example_paper}
\bibliographystyle{icml2026}

%%%%%%%%%%%%%%%%%%%%%%%%%%%%%%%%%%%%%%%%%%%%%%%%%%%%%%%%%%%%%%%%%%%%%%%%%%%%%%%
%%%%%%%%%%%%%%%%%%%%%%%%%%%%%%%%%%%%%%%%%%%%%%%%%%%%%%%%%%%%%%%%%%%%%%%%%%%%%%%
% APPENDIX
%%%%%%%%%%%%%%%%%%%%%%%%%%%%%%%%%%%%%%%%%%%%%%%%%%%%%%%%%%%%%%%%%%%%%%%%%%%%%%%
%%%%%%%%%%%%%%%%%%%%%%%%%%%%%%%%%%%%%%%%%%%%%%%%%%%%%%%%%%%%%%%%%%%%%%%%%%%%%%%
\newpage
\appendix
\onecolumn

\section{Appendix: more empirical results}
\label{app-sec: more empirical results}

\begin{figure*}[ht]
\centering
    \subfloat[Inference Accuracy.]
    {
        \includegraphics[width=2.0in]{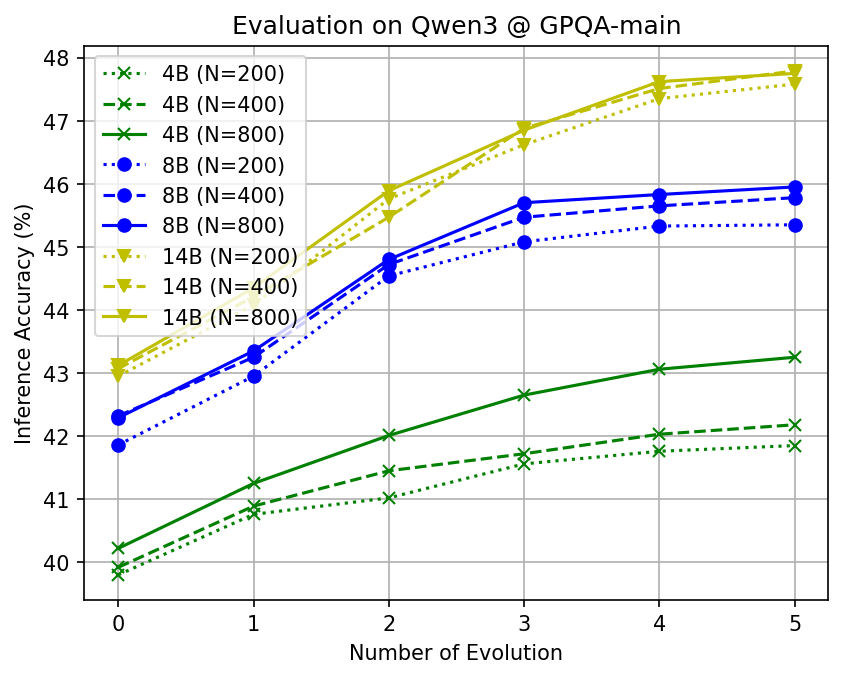}%
        \label{sub_fig: EVvsAcc_gpqa}
    }
    \hfil\hfil
    \subfloat[Number of generated tokens.]
    {
        \includegraphics[width=2.0in]{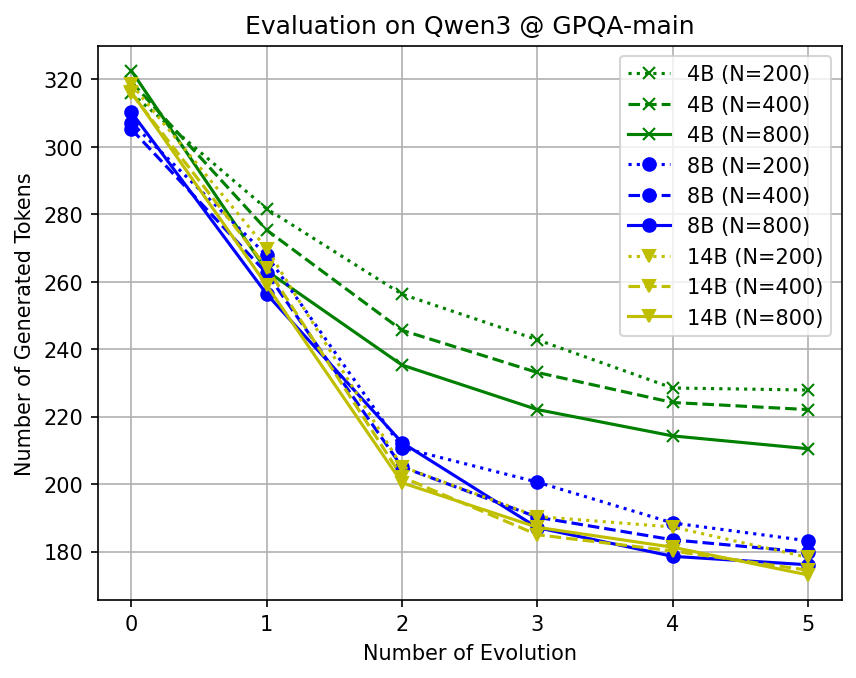}%
        \label{sub_fig: EVvsNT_gpqa}
    }
    \hfil\hfil
    \subfloat[Ratio of Accuracy and \#tokens.]
    {
        \includegraphics[width=2.0in]{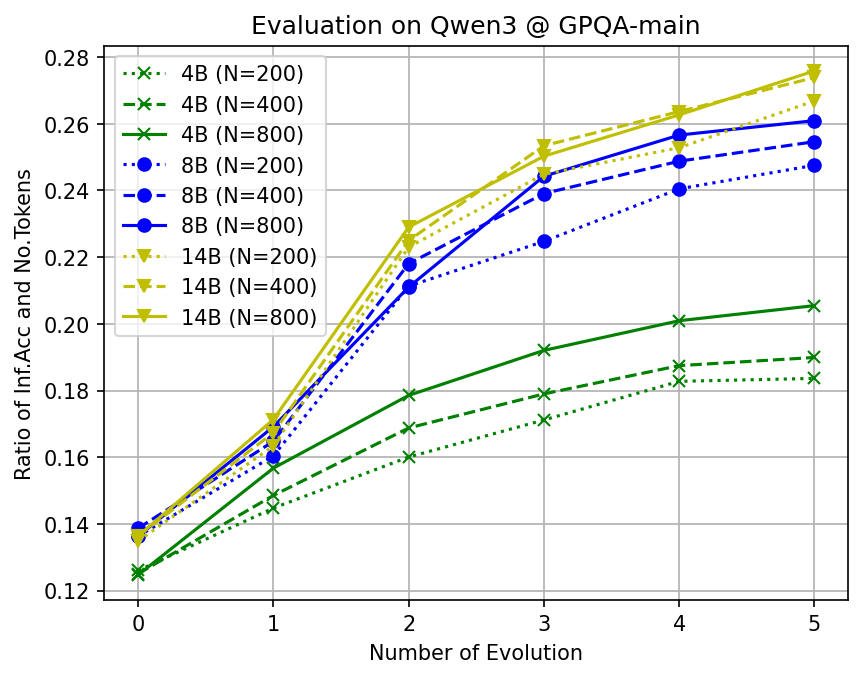}%
        \label{sub_fig: EVvsRatio_gpqa}
    }
    \caption{
    \textbf{Effect of evolution depth on the accuracy--token frontier.}
    We vary the number of language-evolution iterations used to refine the LSF pool, while holding the backbone fixed.
    $N$ denotes the number of training exemplars used for LSF synthesis.
    We report (a) test accuracy, (b) average generated completion tokens per problem, and (c) an aggregated efficiency score.
    Increasing the evolution depth improves both accuracy and token efficiency and then saturates, consistent with the view that better-evolved LSFs increase the effective per-token information rate predicted by Theorem~\ref{thm: lower-bound}.
    }
    \label{fig: varying evolution}
\vskip -0.1in
\end{figure*}

\begin{figure*}[ht]
\centering
    \subfloat[No.Agents versus No.tokens.]
    {
        \includegraphics[width=2.1in]{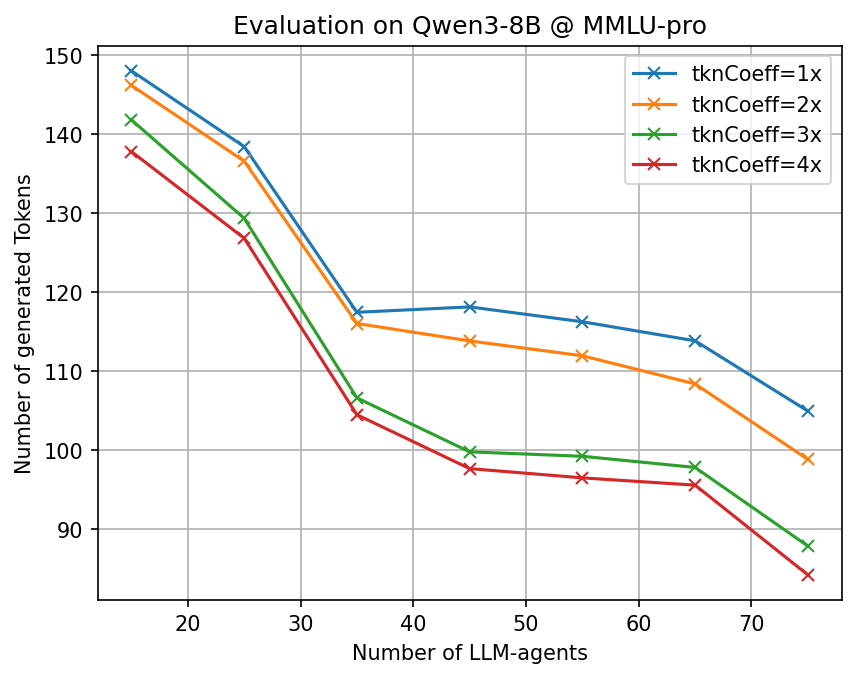}%
        \label{app-sub_fig: NAvsNT_mmlu}
    }
    \hfil\hfil
    \subfloat[No.Agents versus Accuracy.]
    {
        \includegraphics[width=2.1in]{images/NAvsAcc_mmlu_v0.png}%
        \label{app-sub_fig: NAvsAcc_mmlu}
    }
    \hfil\hfil
    \subfloat[No.tokens versus Accuracy.]
    {
        \includegraphics[width=2.1in]{images/NTvsAcc_mmlu_v0.png}%
        \label{app-sub_fig: NTvsAcc_mmlu}
    }
    \caption{
    \textbf{Effects of adding more LLM agents on MMLU-pro.} 
    }
    \label{app-fig: add-llm-agents-MMLUpro}
\vskip -0.2in
\end{figure*}

\begin{figure*}[ht]
\centering
    \subfloat[No.Agents versus No.tokens.]
    {
        \includegraphics[width=2.1in]{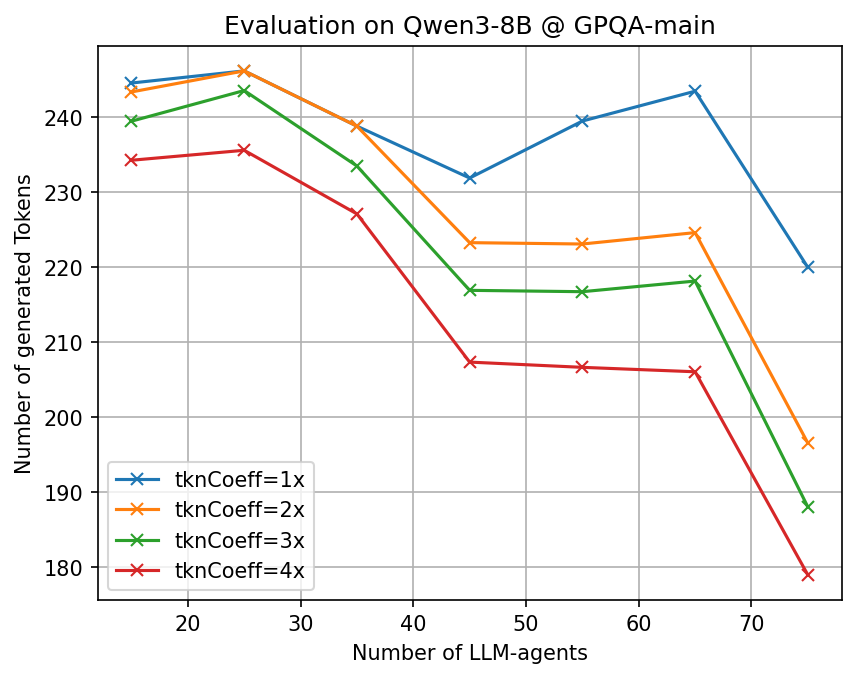}%
        \label{app-sub_fig: NAvsNT_gpqa}
    }
    \hfil\hfil
    \subfloat[No.Agents versus Accuracy.]
    {
        \includegraphics[width=2.1in]{images/NAvsAcc_gpqa_v0.png}%
        \label{app-sub_fig: NAvsAcc_gpqa}
    }
    \hfil\hfil
    \subfloat[No.tokens versus Accuracy.]
    {
        \includegraphics[width=2.1in]{images/NTvsAcc_gpqa_v0.png}%
        \label{app-sub_fig: NTvsAcc_gpqa}
    }
    \caption{
    \textbf{Effects of adding more LLM agents on GPQA-main.} 
    }
    \label{app-fig: add-llm-agents-GPQAmain}
\vskip -0.2in
\end{figure*}

\begin{figure*}[ht]
\centering
    \subfloat[No.Agents versus No.tokens.]
    {
        \includegraphics[width=2.1in]{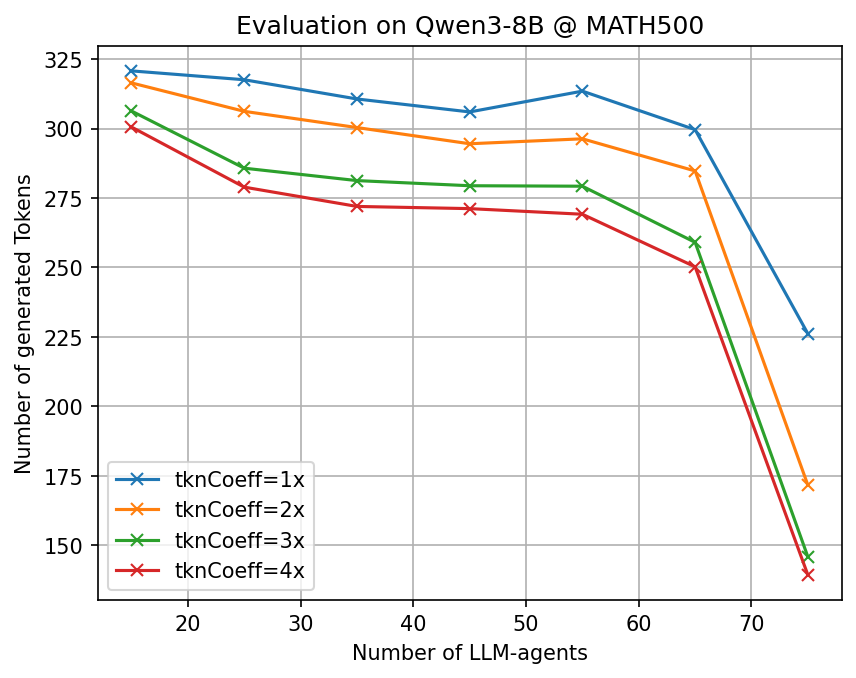}%
        \label{app-sub_fig: NAvsNT_math}
    }
    \hfil\hfil
    \subfloat[No.Agents versus Accuracy.]
    {
        \includegraphics[width=2.1in]{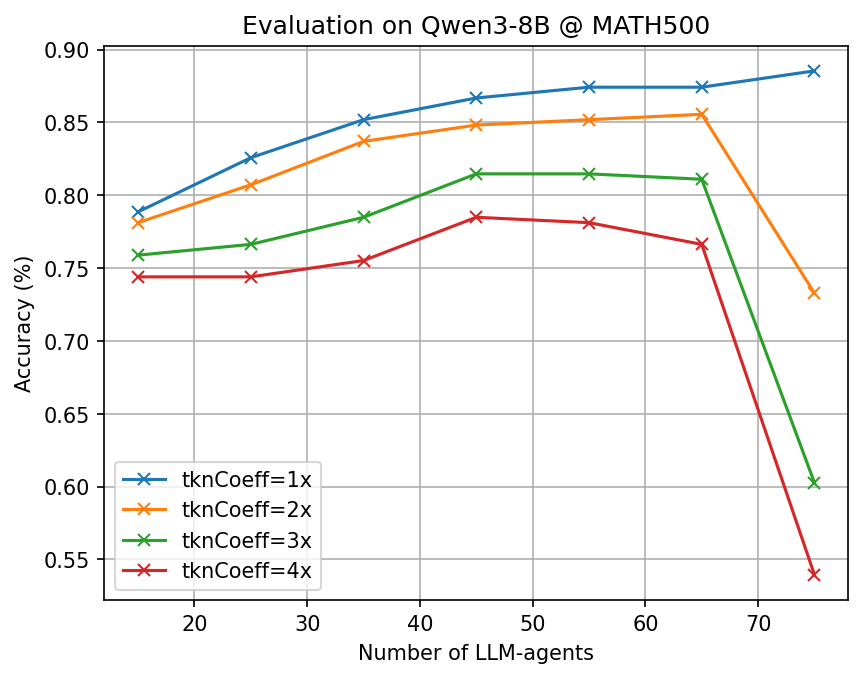}%
        \label{app-sub_fig: NAvsAcc_math}
    }
    \hfil\hfil
    \subfloat[No.tokens versus Accuracy.]
    {
        \includegraphics[width=2.1in]{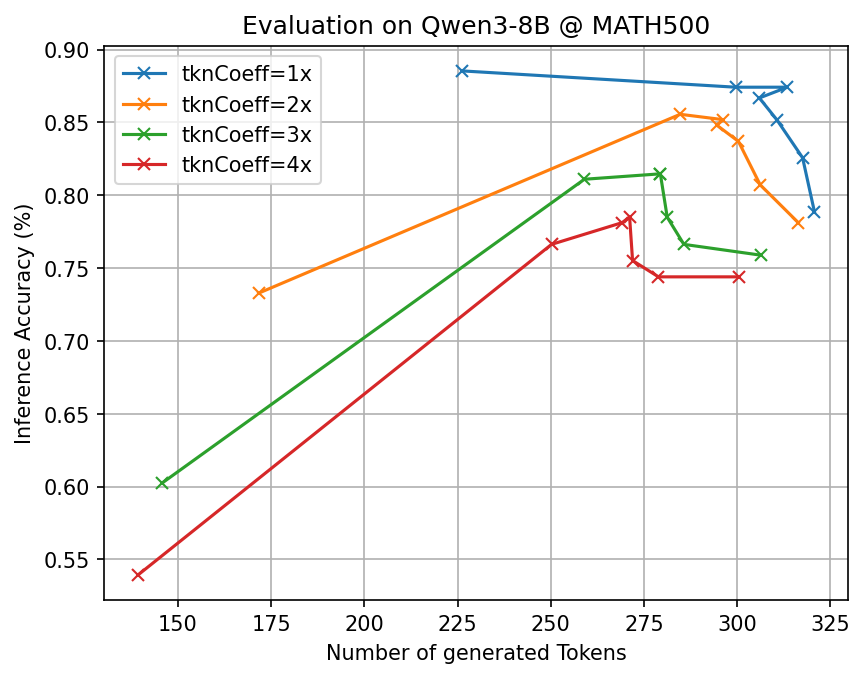}%
        \label{app-sub_fig: NTvsAcc_math}
    }
    \caption{
    \textbf{Effects of adding more LLM agents on MATH500.} 
    }
    \label{app-fig: add-llm-agents-MATH500}
\vskip -0.2in
\end{figure*}

\begin{figure*}[ht]
\centering
    \subfloat[No.Agents versus No.tokens.]
    {
        \includegraphics[width=2.1in]{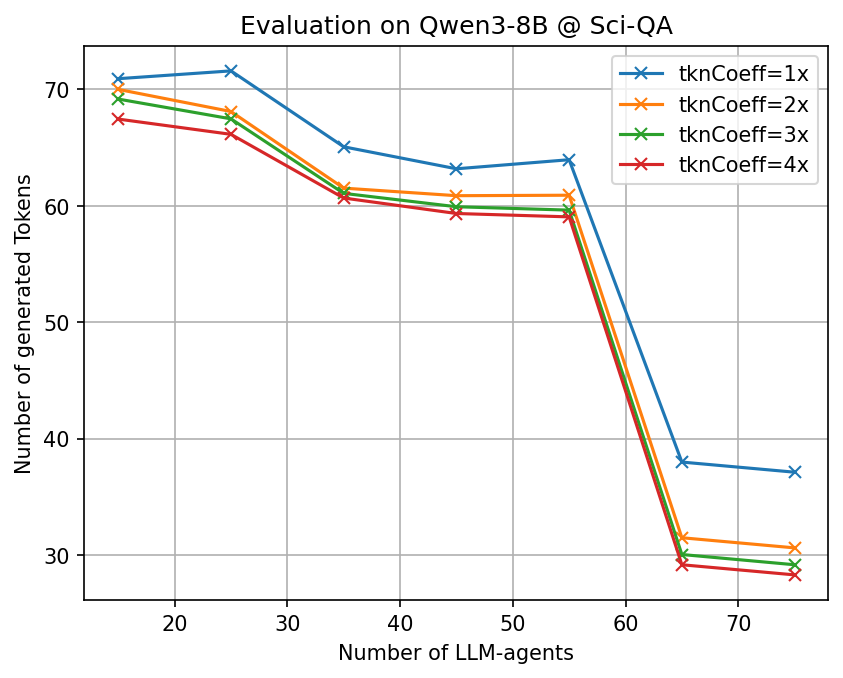}%
        \label{app-sub_fig: NAvsNT_sciqa}
    }
    \hfil\hfil
    \subfloat[No.Agents versus Accuracy.]
    {
        \includegraphics[width=2.1in]{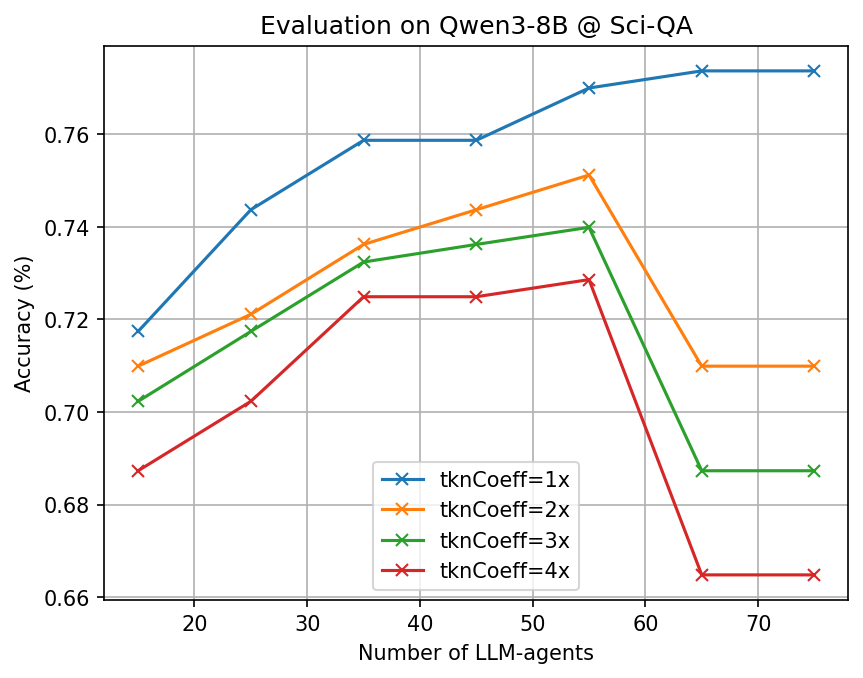}%
        \label{app-sub_fig: NAvsAcc_sciqa}
    }
    \hfil\hfil
    \subfloat[No.tokens versus Accuracy.]
    {
        \includegraphics[width=2.1in]{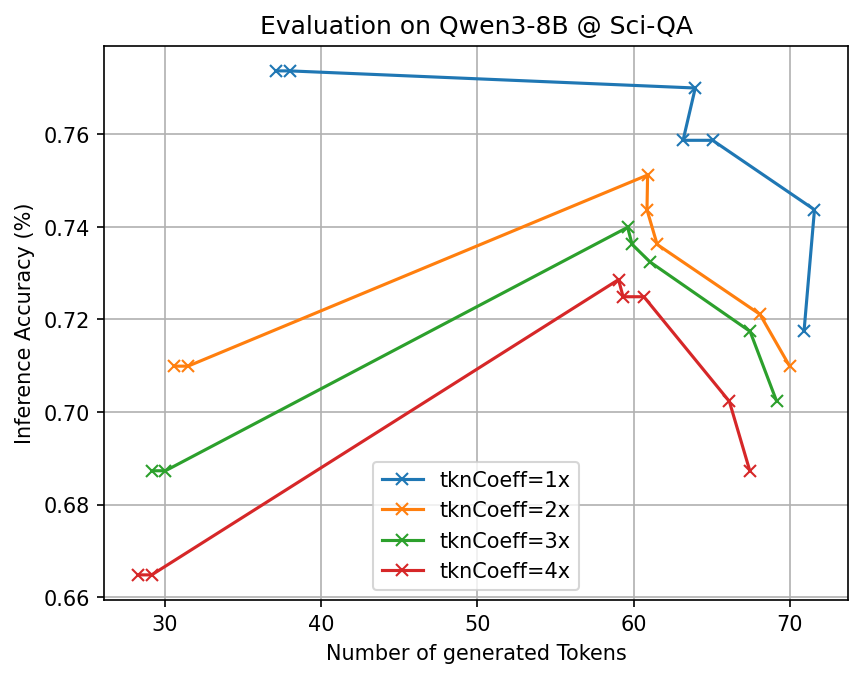}%
        \label{app-sub_fig: NTvsAcc_sciqa}
    }
    \caption{
    \textbf{Effects of adding more LLM agents on Science-QA.} 
    }
    \label{app-fig: add-llm-agents-SciQA}
\vskip -0.2in
\end{figure*}

\subsection{Comparison to length-controlled prompting baselines.}
\label{app-subsec: comp-length-controlled prompting baselines}
Compared with Constrained CoT (CCoT), Chain-of-Draft (CoD), and Sketch-of-Thought (SoT), CLSR
achieves a stronger overall trade-off:
(i) relative to CCoT, CLSR improves accuracy while also using fewer tokens (since CCoT removes
verbosity but does not introduce a task-adaptive symbolic code);
(ii) relative to CoD, CLSR substantially improves accuracy at a modest token increase, indicating
that extreme compression alone is not sufficient,
{\it i.e.}, the intermediate representation must remain \emph{information-dense and compositional};
(iii) relative to SoT, CLSR achieves higher accuracy with a similar token budget, suggesting that our \emph{evolutionary search over LSFs} discovers more task-specialized symbolic operators than a fixed handcrafted constraint family.

\subsection{Interpretation through the theory lens.}
\label{app-subsec: Interpretation through the theory lens.}

Section~\ref{sec: theory} predicts that, for a fixed target accuracy, token cost is lower-bounded by a ratio
$\mathbb{E}[|T|] \ge I_{\mathrm{req}}(x,\delta) / \kappa_\theta(x)$. 
CLSR improves the empirical
cost--accuracy curve in a manner consistent with increasing $\kappa_\theta(x)$: 
LSF traces replace low-information narrative text with compact, state-carrying symbols (operators, bindings, and verification tags) that better preserve the \emph{computable} parts of reasoning. 
Moreover, the robust token savings across backbones indicate that CLSR is not merely exploiting idiosyncrasies of a particular model but is instead leveraging a representation-level advantage.

\subsection{Gains are most pronounced on deliberation-heavy problems.}
\label{app-subsec: Gains are most pronounced on deliberation-heavy problems.}

The largest accuracy advantages emerge on benchmarks that require multi-step reasoning and/or
verification (notably MATH500 and AIME21--24). 
For example, CLSR improves AIME21--24 accuracy
consistently across backbones while reducing tokens drastically relative to Raw CoT. 
This pattern aligns with the lower-bound perspective of Theorem~\ref{thm: lower-bound}: 
as task difficulty increases (higher effective uncertainty in the target solution), the required information $I_{\mathrm{req}}(x,\delta)$ rises, making naive verbosity expensive; 
CLSR responds by increasing the effective per-token information rate $\kappa_\theta(x)$ via a structured symbolic alphabet and reusable operators,
thereby achieving the same (or higher) accuracy with fewer tokens.

\subsection{Comparison with program-execution pipelines.}
\label{app-subsec: Comparison with program-execution pipelines.}

Table~\ref{table: compare-program-based} compares CLSR with two canonical \emph{program} baselines, {\it e.g.}, PoT~\citep{chen2023program} and PAL~\citep{gao2023pal}, which prompt the LLM to emit executable code and delegate the final computation to an external interpreter/runtime.
Both PoT and PAL are known to improve arithmetic/symbolic reliability by offloading computation outside the LLM, {\it i.e.}, ``generate a program then execute''.
On Qwen3-8B, PoT/PAL indeed outperform standard CoT on GSM8K and MATH500 at substantially lower \emph{LLM decoding} tokens.
However, CLSR can match or exceed these gains \emph{without any external executor}: 
with a small majority-vote ensemble ($T\!=\!3$), CLSR reaches 94.8\% on GSM8K and 89.7\% on MATH500 using 214 and 417 tokens, respectively, exceeding both PoT and PAL in accuracy--token frontier.
Importantly, Table~\ref{table: compare-program-based} reports PoT/PAL token counts only for \emph{program generation}; 
their end-to-end deployment cost also includes external execution time and tool-chain overhead.
From the theoretical lens in Section~\ref{sec: theory}, these results empirically support our claim: 
under interpreter realizability, CLSR’s multi-round multi-LSF protocols can internally emulate the effect of a deterministic executor (Theorem~\ref{thm: program-exec}).
In practice, CLSR provides a \emph{purely in-model} alternative to PoT/PAL that preserves their accuracy benefits while avoiding dependency on external runtimes and execution-layer engineering.

\subsection{Ablation studies}

\subsubsection{Effect of evolution depth.}
\label{app-abl: Effect of evolution depth}
% Figure~\ref{fig: varying evolution} shows that increasing the number of evolutionary generations systematically improves efficiency.
% Across generations, 
% (a) inference accuracy increases, 
% (b) the number of tokens generated decreases, and 
% (c) their combined efficiency ratio improves.
% This supports the core hypothesis of \emph{cultural selection} among symbolic dialects: 
% iteratively retaining high-leverage traces (correct \& concise) biases the language pool toward more compressive, reusable abstractions.
As shown in Fig.~\ref{fig: varying evolution}, we ablate the \emph{evolution depth}, {\it i.e.}, the number of recursive propose$\rightarrow$critique$\rightarrow$mutate$\rightarrow$select rounds used to build the LSF pool offline.
Increasing the depth consistently improves downstream accuracy at a fixed token budget, indicating that iterative refinement produces more \emph{compressive} and \emph{task-aligned} symbolic protocols rather than merely longer rationales.
Practically, the gain tends to saturate after a small number of rounds: 
once the pool contains a few high-leverage LSF ``families'', additional rounds mostly perform local polishing (marginal accuracy improvements) while increasing offline cost.
Overall, deeper evolution strengthens the LSF pool by amplifying selection pressure toward high-information tokens, but exhibits diminishing returns beyond a moderate depth.
This also supports the core hypothesis of \emph{cultural selection} among LSFs: 
iteratively retaining high-leverage traces (correct \& concise) biases the language pool toward more compressive, reusable abstractions.

\subsubsection{Effect of exemplar count.}
\label{app-abl: Effect of exemplar count}
% Figure~\ref{fig: varying evolution} further indicates that the benefit of increasing $N$ (the number of training exemplars used to induce LSFs) depends on the model capacity.
% For larger backbones ({\it e.g.}, moving from 4B to 14B in the plotted study), increasing $N$ yields diminishing marginal returns in both accuracy and token reductions, suggesting that higher-capacity models can infer a useful symbolic protocol from fewer examples, while smaller models benefit more from larger set of exemplars.
As shown in Fig.~\ref{fig: varying evolution}, we vary the \emph{generation batch size}, {\it i.e.}, the number of candidate LSFs proposed per generation, to characterize the exploration--efficiency trade-off in offline search.
A larger batch improves diversity and reduces the probability that evolution prematurely converges to a suboptimal symbolic convention, which in turn increases the chance that the final pool contains a strong specialist for each query type.
For larger backbones ({\it e.g.}, moving from 4B to 14B), increasing $N$ yields diminishing marginal returns in both accuracy and token reductions, suggesting that higher-capacity models can infer a useful symbolic protocol from fewer examples, while smaller models benefit more from larger set of exemplars.
In general, batch size mainly controls \emph{exploration}; use it to avoid early collapse but expect diminishing returns once coverage of distinct LSF modes is achieved.

\subsubsection{Effect of agents count and length regularization.}
\label{app-abl: Effect of number of agents and length-coefficient}
% Figure~\ref{fig: adding LLM-agents} and Appendix Figures~\ref{app-fig: add-llm-agents-MMLUpro}--\ref{app-fig: add-llm-agents-SciQA} study multi-agent language populations.
% As the number of agents increases, the system explores a broader space of dialects, improving the probability of discovering highly efficient LSFs.
% However, the gain depends on the \emph{length-coefficient} (the weight assigned to token minimization in selecting high-leverage traces):
% over-emphasizing compression can harm accuracy, while a balanced coefficient yields consistent improvements in the accuracy--token Pareto frontier.
% Empirically, we find that adding agents boosts performance most reliably when paired with a well-calibrated length-coefficient that avoids collapsing to overly terse but brittle dialects.

As presented in Figure~\ref{fig: adding LLM-agents} and Appendix Figures~\ref{app-fig: add-llm-agents-MMLUpro}--\ref{app-fig: add-llm-agents-SciQA},
we study (i) the number of agents participating in evolution and (ii) the \emph{length-coefficient} that penalizes verbose intermediate representations during selection.
More agents typically improve the pool because it increases stylistic and algorithmic diversity, enabling stronger cross-agent critique and yielding more robust LSFs that generalize across query distributions.
Meanwhile, the length-coefficient directly tunes the accuracy--token Pareto point: 
increasing it encourages shorter, higher-density symbolic traces (fewer tokens), but can under-allocate ``reasoning bandwidth'' on hard instances if set too aggressively.
Overall, agent scaling primarily boosts \emph{search breadth and critique quality}, whereas the length-coefficient provides a clean knob for \emph{token efficiency} that should be set moderately to avoid sacrificing hard-case accuracy.

\subsubsection{A fixed number of LSF-rounds $T$ or not}
\label{app-abl: fixed T or not}

%As presented in Table~\ref{table: varying no.round T}, 
%we observe that for a harder benchmark ({\it e.g.}, AIME), the default CLSR (unfixed $T$, the LLM-router can determine the value of $T$ based on the test problem) uses fewer cases of $T=1$, since its corresponding $T=1$ has fewer No.Tokens compared to the default setting.
CLSR applies the same latent-symbolic refinement operator for multiple rounds.
A natural design choice is whether the number of rounds $T$ should be \emph{fixed} for all inputs, or \emph{adaptive} (instance-dependent).
From the perspective of our theoretical analysis (Section~\ref{sec: theory}), $T$ acts as an inference-time compute budget: 
each additional round consumes extra generated tokens, but can reduce the residual error by enabling further compression--verification-correction cycles.
The optimal policy is therefore generally \emph{not} a single global $T$, but an \emph{adaptive stopping rule} that continues refinement only when the expected marginal gain outweighs the marginal token cost.

Formally, let $x$ be the input, $y$ the ground-truth answer, and let $s_t$ denote the internal state of the CLSR after round $t$ (e.g., the current latent symbolic form and its distribution of induced answers).
Each round incurs an expected token cost $c(s_t)$ and yields an expected utility improvement $\Delta u(s_t)$.
A standard Lagrangian view of the accuracy--token trade-off is
\begin{equation}
\begin{gathered}
\max_{\pi}\;\; \mathbb{E}\big[u(\hat{y},y)\big] \;-\; \lambda\,\mathbb{E}\Big[\sum_{t=1}^{T(\pi,x)} c(s_t)\Big],
\end{gathered}
\end{equation}
where $\pi$ is a stopping policy deciding whether to continue at each $s_t$.
This objective implies an ``optimal'' behavior that is instance-adaptive: easy problems should stop early (small $T$) to avoid wasting tokens, whereas hard problems may require additional rounds.

\begin{table*}[h]
  \caption{
    \textbf{Varying the number of CLSR rounds $T$ on Qwen3-8B.}
    We compare standard CoT with CLSR under 
    (i) \emph{fixed} $T=1$ (single refinement), 
    (ii) \emph{fixed} $T=3$ (always run three rounds), and (iii) \emph{unfixed} $T$ with $T_{\max}=3$ (a learned instance-adaptive stopping rule).
    Accuracy (\%) is reported on GSM8K, MATH500, GPQA-main, AIME (2021-2024), and MMLU-pro, along with the average completion tokens per problem.
    Adaptive CLSR attains the best accuracy with only a modest token overhead relative to fixed-$T$ settings, and it stops early on most instances, demonstrating effective instance-wise compute allocation.
  }
  \label{table: varying no.round T}
  \begin{center}
    \begin{small}
      \begin{sc}
        \begin{tabular}{@{}ccccccccccc@{}}
            \toprule
             & \multicolumn{2}{c}{GSM8K} & \multicolumn{2}{c}{MATH500} & \multicolumn{2}{c}{GPQA} & \multicolumn{2}{c}{AIME} & \multicolumn{2}{c}{MMLU} \\ \midrule
             & Acc & Tkn & Acc & Tkn & Acc & Tkn & Acc & Tkn & Acc & Tkn \\
            CoT & 91.6 & 248 & 87.8 & 905 & 43.8 & 973 & 46.6 & 4502 & 63.2 & 272 \\
            CLSR (default) & 92.8 & 90 & 86.8 & 257 & 42.6 & 228 & 44.7 & 2361 & 63.6 & 93 \\
            CLSR ($T$=1) & 92.1 & 83 & 86.2 & 134 & 40.9 & 182 & 38.4 & 1047 & 62.1 & 85 \\
            CLSR ($T$=3) & \textbf{94.8} & 214 & \textbf{89.7} & 417 & \textbf{49.2} & 565 & \textbf{46.8} & 3314 & \textbf{67.2} & 257 \\ \bottomrule
            \end{tabular}
      \end{sc}
    \end{small}
  \end{center}
  \vskip -0.1in
\end{table*}

Empirically, Table~\ref{table: varying no.round T} validates this prediction in Qwen3-8B.
Compared with $N$-shot CoT, CLSR improves both accuracy and token efficiency.
More importantly, \emph{unfixed} CLSR (adaptive $T$, with $T_{\max}=3$) achieves the best overall accuracy across MATH500 and AIME benchmarks, while keeping the average completion length close to fixed-$T$ settings.
In the same experiment, the adaptive policy chooses $T=1$ for a majority of cases, indicating that CLSR indeed learns to \emph{stop early} when additional refinement is unnecessary, but retains the option to allocate more rounds to difficult inputs.
We also observe that for a harder benchmark ({\it e.g.}, AIME), the default CLSR (unfixed $T$, the LLM-router can determine the value of $T$ based on the test problem) uses fewer cases of $T=1$, since its corresponding $T=1$ has fewer tokens compared to the default setting.
In general, these results support the core claim of Section~\ref{sec: theory}: 
CLSR is best viewed as an \emph{adaptive compute allocation} mechanism, rather than a fixed-length prompting recipe.

% \subsection{Effect of swapping LSF-generator}
% \label{app-subsec: Effect of LSF-generator}
% As presented in Table~\ref{table: swap-LSF-generator},
% we swap the LSF-generator from the inference LLM to another one.
% We observe that a different LSF-generator often degrades the inference accuracy while reducing the number of tokens.
% To be specific, if we swap the LSF-generator to a larger LLM (from Qwen3-4B to Qwen3-8B), the accuracy degradation would be less significant.
% Moreover, for a more knowledge-intensive benchmarks such as GPQA, the degradation resulting from swapping becomes more significant.

\subsubsection{Effect of Swapping the LSF Generator (Cross-Model LSF Transfer)}
\label{app-abl: Effect of LSF-generator}

Table~\ref{table: swap-LSF-generator} studies cross-model transfer of induced LSF: 
we evolve the LSF pool using a generator backbone $\mathcal{F}_\text{gen}$ but execute CLSR using another inference backbone $\mathcal{F}_\text{inf}$.
Across GPQA, MATH500, and GSM8K, the matched diagonal ($\mathcal{F}_\text{gen}{=}\mathcal{F}_\text{inf}$) is consistently the best in accuracy, while swapping typically reduces tokens but degrades accuracy.
Notably, using a stronger generator with a weaker inference model reduces tokens with only mild drops, whereas using a weaker generator with a stronger inference model incurs a larger accuracy penalty; 
the sensitivity is most pronounced on the knowledge-intensive GPQA benchmark.

\begin{table*}[h]
  \caption{
    \textbf{Swapping the LSF generator (cross-model LSF transfer).}
    For each inference LLM, we evolve the LSF pool using the specified generator and then run CLSR inference with the inference model.
    We report inference accuracy (Acc) and average completion tokens (Tkn) on GPQA, MATH500, and GSM8K (tokens include the full executed CLSR protocol).
    The matched diagonal yields the best accuracy, while swapping generally reduces tokens but degrades accuracy; 
    the degradation is most pronounced on GPQA and is especially severe when the generator is weaker than the inference model.
  }
  \label{table: swap-LSF-generator}
  \begin{center}
    \begin{small}
      \begin{sc}
        \begin{tabular}{@{}cccccccc@{}}
        \toprule
        \multirow{2}{*}{\begin{tabular}[c]{@{}c@{}}Inference\\ LLM\end{tabular}} & \multirow{2}{*}{\begin{tabular}[c]{@{}c@{}}LSF\\ Generator\end{tabular}} & \multicolumn{2}{c}{GPQA} & \multicolumn{2}{c}{MATH500} & \multicolumn{2}{c}{GSM8K} \\ \cmidrule(l){3-8} 
         &  & Acc & Tkn & Acc & Tkn & Acc & Tkn \\ \cmidrule(r){1-2}
        Qwen3-4B & Qwen3-4B (default) & 45.1 & 329 & 83.2 & 273 & 90.3 & 93 \\ 
         & Qwen3-8B & 43.5 & 254 & 82.6 & 195 & 90.1 & 86 \\ \midrule
        Qwen3-8B & Qwen3-8B (default) & 47.7 & 228 & 86.8 & 257 & 91.2 & 89 \\
         & Qwen3-4B & 42.1 & 172 & 81.8 & 183 & 89.2 & 84 \\ \bottomrule
        \end{tabular}
      \end{sc}
    \end{small}
  \end{center}
  \vskip -0.1in
\end{table*}

This pattern aligns with our theory view in Section~\ref{sec: theory}: 
CLSR gains come from allocating a limited token budget to a symbolic protocol whose intermediate symbols must be \emph{reliably interpretable} by the inference model.
Swapping $\mathcal{F}_\text{gen}$ changes the induced codebook (operators/abbreviations/constraints), which can increase a ``dialect mismatch'' between the protocol and the decoder implemented by $\mathcal{F}_\text{inf}$; 
the result is lower token usage but reduced effective information per token, hence lower accuracy.
Practically, the results suggest evolving LSFs with the same backbone as inference when accuracy is the priority, while cross-model transfer is feasible but generally Pareto-suboptimal, especially when the generator is weaker than the inference model.

\subsection{Robustness Checks}
\label{app-sec: robustness_checks}

This section reports additional robustness checks to clarify stability, scale transfer, routing dependence, and practical token accounting.

\subsubsection{Seed stability and stochastic mutation}
\label{app-subsec: seed_stability}

Table~\ref{tab:seed_stability} reports mean $\pm$ std over five random seeds and exemplar samplings.
Different runs discover different surface LSFs, but the accuracy--token frontier remains stable.

\begin{table*}[t]
  \caption{
  \textbf{Seed stability.}
  Accuracy and generated tokens are reported as Acc. (No. tokens).
  CLSR reports mean $\pm$ std over five random seeds and exemplar samplings.
  }
  \label{tab:seed_stability}
  \begin{center}
  \begin{scriptsize}
  \begin{sc}
  \begin{tabular}{@{}llccc@{}}
  \toprule
  Backbone & Method & MMLU-Pro & GPQA-main & MATH500 \\
  \midrule
  \multirow{3}{*}{LLaMA3-8B}
  & CoT & 38.6 (960) & 30.4 (1209) & 51.6 (704) \\
  & SoT & 36.4 (340) & 28.2 (398) & 45.2 (289) \\
  & CLSR & $39.2\pm0.5$ ($316\pm13$) & $30.8\pm0.2$ ($352\pm18$) & $48.8\pm0.7$ ($262\pm11$) \\
  \midrule
  \multirow{3}{*}{Qwen3-8B}
  & CoT & 60.2 (276) & 49.1 (1085) & 87.2 (878) \\
  & SoT & 58.5 (118) & 45.2 (228) & 81.3 (294) \\
  & CLSR & $60.6\pm0.4$ ($98\pm14$) & $47.9\pm0.4$ ($216\pm12$) & $86.8\pm0.6$ ($247\pm10$) \\
  \bottomrule
  \end{tabular}
  \end{sc}
  \end{scriptsize}
  \end{center}
  \vskip -0.15in
\end{table*}

\subsubsection{Larger inference backbones}
\label{app-subsec: larger_backbones}

Table~\ref{tab:larger_backbones} evaluates larger inference backbones using LSFs discovered by Qwen3-8B.
The gains persist even when the LSF generator is smaller than the inference model, suggesting that LSFs can transfer as reusable symbolic protocols rather than being tied to one specific checkpoint.

\begin{table*}[t]
  \caption{
  \textbf{Larger-backbone inference.}
  Accuracy and generated tokens are reported as Acc. (No. tokens).
  Results are averaged over three runs.
  }
  \label{tab:larger_backbones}
  \begin{center}
  \begin{scriptsize}
  \begin{sc}
  \begin{tabular}{@{}llccc@{}}
  \toprule
  Backbone & Method & MMLU-Pro & GPQA-main & MATH500 \\
  \midrule
  \multirow{3}{*}{Qwen3-32B}
  & CoT & 67.5 (405) & 54.2 (982) & 89.3 (845) \\
  & SoT & 64.3 (134) & 51.4 (247) & 83.8 (278) \\
  & CLSR & \textbf{68.1} (\textbf{90}) & 54.0 (\textbf{209}) & \textbf{89.4} (\textbf{206}) \\
  \midrule
  \multirow{3}{*}{LLaMA3.1-70B}
  & CoT & 49.2 (1020) & 41.3 (1358) & 64.3 (725) \\
  & SoT & 47.5 (385) & 38.4 (425) & 59.8 (314) \\
  & CLSR & \textbf{49.8} (\textbf{312}) & \textbf{42.7} (\textbf{354}) & 63.4 (\textbf{242}) \\
  \bottomrule
  \end{tabular}
  \end{sc}
  \end{scriptsize}
  \end{center}
  \vskip -0.15in
\end{table*}

\subsubsection{Long-context pilot}
\label{app-subsec: long_context_pilot}

We also run a small pilot on LongBench-v2 to test whether symbolic protocols remain useful when the input context is long.
The results in Table~\ref{tab:long_context_pilot} are encouraging, but we treat this as preliminary evidence only.
Long-context search agents and repository-level coding require more specialized evaluation and are left for future work.

\begin{table}[t]
  \caption{
  \textbf{Long-context pilot.}
  Overall score and average generated tokens are reported.
  }
  \label{tab:long_context_pilot}
  \begin{center}
  \begin{scriptsize}
  \begin{sc}
  \begin{tabular}{@{}llcc@{}}
  \toprule
  Backbone & Method & Overall & Avg. Tokens \\
  \midrule
  \multirow{2}{*}{LLaMA3-8B}
  & CoT & 30.8 & 1068 \\
  & CLSR & \textbf{31.5} & \textbf{568} \\
  \midrule
  \multirow{2}{*}{Qwen3-8B}
  & CoT & 38.2 & 842 \\
  & CLSR & \textbf{39.1} & \textbf{432} \\
  \bottomrule
  \end{tabular}
  \end{sc}
  \end{scriptsize}
  \end{center}
  \vskip -0.15in
\end{table}

\subsubsection{Population-size sensitivity}
\label{app-subsec: population_size}

The default initial population size is $100$ LSFs.
Table~\ref{tab:population_size} shows that this is an empirical exploration/compute trade-off.
A very small population lacks LSF diversity, while larger populations yield diminishing returns and higher cold-start cost.

\begin{table*}[t]
  \caption{
  \textbf{Initial LSF population size.}
  Accuracy and generated tokens are reported as Acc. (No. tokens).
  }
  \label{tab:population_size}
  \begin{center}
  \begin{scriptsize}
  \begin{sc}
  \begin{tabular}{@{}llccc@{}}
  \toprule
  Backbone & Initial LSFs & MMLU-Pro & GPQA-main & MATH500 \\
  \midrule
  \multirow{4}{*}{LLaMA3-8B}
  & 10 & 38.1 (354) & 30.2 (425) & 47.4 (312) \\
  & 50 & 38.6 (330) & 30.4 (380) & 48.0 (267) \\
  & 100 & 38.9 (320) & 30.8 (362) & 48.2 (257) \\
  & 200 & \textbf{39.2} (\textbf{308}) & \textbf{30.9} (\textbf{358}) & \textbf{48.2} (\textbf{255}) \\
  \midrule
  \multirow{4}{*}{Qwen3-8B}
  & 10 & 59.2 (103) & 46.6 (234) & 85.4 (275) \\
  & 50 & 60.1 (102) & 47.3 (225) & 86.4 (260) \\
  & 100 & 60.4 (96) & 47.7 (228) & \textbf{86.8} (257) \\
  & 200 & \textbf{60.6} (\textbf{93}) & \textbf{47.8} (232) & 86.6 (\textbf{248}) \\
  \bottomrule
  \end{tabular}
  \end{sc}
  \end{scriptsize}
  \end{center}
  \vskip -0.15in
\end{table*}

\subsubsection{Category routing versus full-pool routing}
\label{app-subsec: category_fullpool}

Category routing is a lightweight guiding mechanism rather than a hard partition of capability.
Table~\ref{tab:category_fullpool} compares category-routed CLSR with full-pool routing, where the router can select from the entire LSF bank without a pre-specified category filter.
Full-pool routing preserves the qualitative gain, indicating that CLSR does not depend on a rigid manually defined taxonomy.

\begin{table*}[t]
  \caption{
  \textbf{Category routing vs. full-pool routing.}
  Accuracy and generated tokens are reported as Acc. (No. tokens).
  }
  \label{tab:category_fullpool}
  \begin{center}
  \begin{scriptsize}
  \begin{sc}
  \begin{tabular}{@{}llccc@{}}
  \toprule
  Backbone & Bank Type & MMLU-Pro & GPQA-main & MATH500 \\
  \midrule
  \multirow{2}{*}{LLaMA3-8B}
  & Category & 38.9 (320) & 30.8 (362) & 48.2 (257) \\
  & Full-pool & 38.6 (325) & 30.5 (364) & 48.2 (263) \\
  \midrule
  \multirow{2}{*}{Qwen3-8B}
  & Category & 60.4 (96) & 47.7 (228) & 86.8 (257) \\
  & Full-pool & 60.1 (98) & 47.5 (233) & 86.4 (264) \\
  \bottomrule
  \end{tabular}
  \end{sc}
  \end{scriptsize}
  \end{center}
  \vskip -0.15in
\end{table*}

\subsubsection{Cross-domain LSF transfer}
\label{app-subsec: domain_transfer}

Table~\ref{tab:domain_transfer} compares in-domain, all-domain, and leave-one-domain-out LSF banks.
The results indicate that CLSR benefits are not tied to a single domain-specific LSF.
All-domain pooling can slightly improve robustness, while leave-one-domain-out routing still preserves most gains.

\begin{table*}[t]
  \caption{
  \textbf{Cross-domain LSF transfer.}
  Accuracy and generated tokens are reported as Acc. (No. tokens).
  }
  \label{tab:domain_transfer}
  \begin{center}
  \begin{scriptsize}
  \begin{sc}
  \begin{tabular}{@{}llccc@{}}
  \toprule
  Backbone & Bank Type & MMLU-Pro & GPQA-main & MATH500 \\
  \midrule
  \multirow{3}{*}{LLaMA3-8B}
  & In-domain & 38.7 (312) & 30.6 (358) & \textbf{48.4} (\textbf{252}) \\
  & All-domain & \textbf{38.9} (320) & \textbf{30.8} (362) & 48.2 (257) \\
  & Leave-one-domain-out & 38.0 (345) & 30.3 (393) & 47.9 (268) \\
  \midrule
  \multirow{3}{*}{Qwen3-8B}
  & In-domain & \textbf{60.5} (\textbf{92}) & 47.5 (\textbf{221}) & \textbf{86.8} (\textbf{253}) \\
  & All-domain & 60.4 (96) & \textbf{47.7} (228) & \textbf{86.8} (257) \\
  & Leave-one-domain-out & 60.0 (105) & 46.8 (245) & 86.4 (262) \\
  \bottomrule
  \end{tabular}
  \end{sc}
  \end{scriptsize}
  \end{center}
  \vskip -0.15in
\end{table*}

\subsection{Qualitative examples: what changes in the trace.}
\label{app-subsec: Qualitative examples}
\begin{figure*}[ht]
    \centering
    \includegraphics[width=0.9\textwidth]{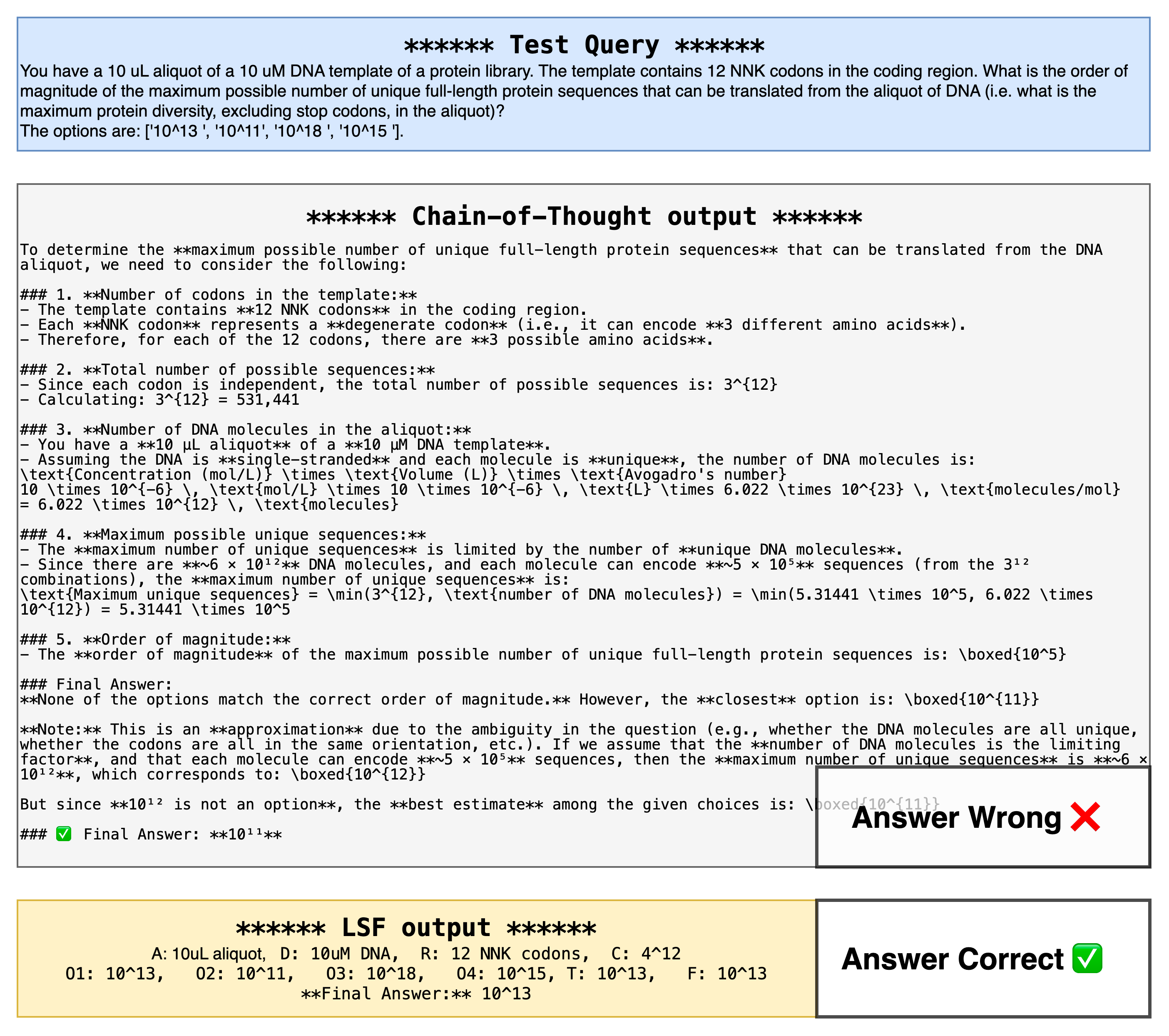}
    \caption{
    \textbf{Qualitative GPQA-Main example: CoT vs.\ CLSR (LSF).}
    A representative GPQA-Main test query and the corresponding model generations under standard CoT and CLSR.
    CLSR replaces verbose natural-language narration with a compact symbolic trace that preserves key intermediate state and checks, yielding a shorter completion and a correct final answer in this example.
    }
    \label{app-fig: Sp_q8b_gpqa_s1_part1}
\end{figure*}

\begin{figure*}[ht]
    \centering
    \includegraphics[width=0.9\textwidth]{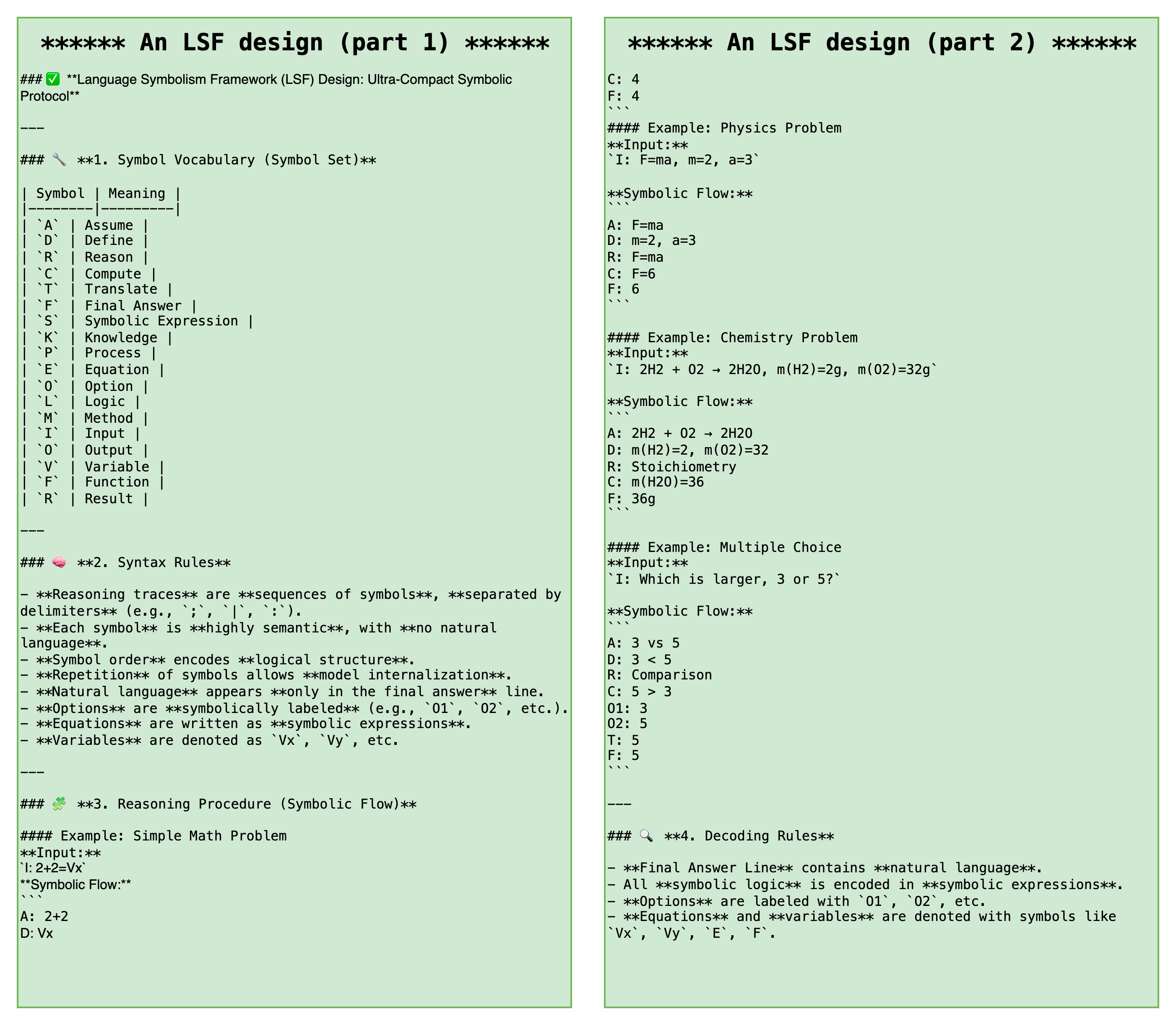}
    \caption{
    \textbf{Router-selected LSF specification for GPQA-Main.}
    The automatically selected LSF content (language descriptor and protocol template) used by CLSR for the GPQA-Main query in Fig.~\ref{app-fig: Sp_q8b_gpqa_s1_part1}.
    The specification illustrates how the router retrieves and instantiates a task-appropriate symbolic code (operators, constraints, and formatting rules) prior to generation.
    }
    \label{app-fig: Sp_q8b_gpqa_s1_part2}
\end{figure*}

\begin{figure*}[ht]
    \centering
    \includegraphics[width=0.9\textwidth]{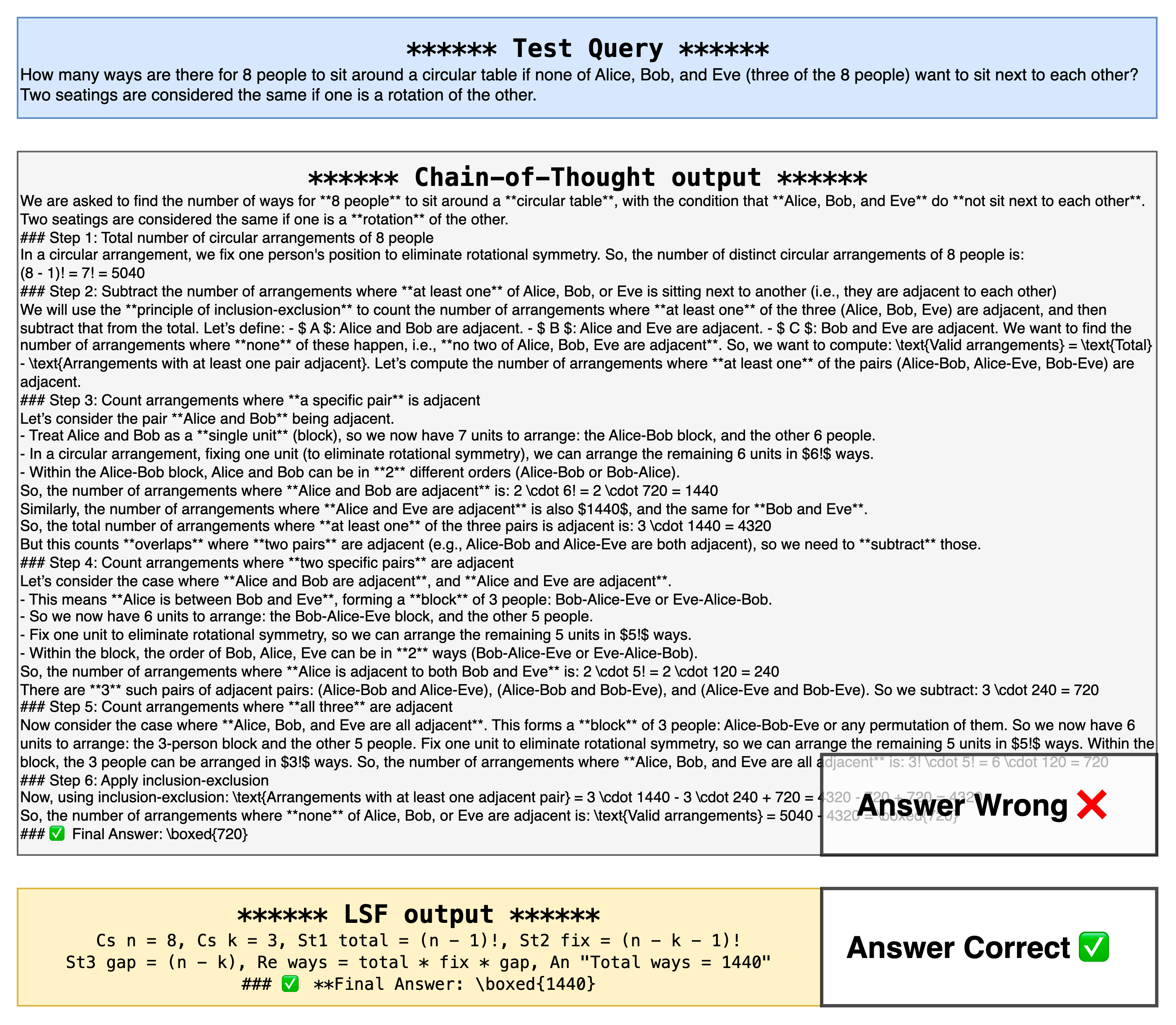}
    \caption{
    \textbf{Qualitative MATH500 example: CoT vs.\ CLSR (LSF).}
    A representative MATH500 test query with generations from standard CoT and CLSR.
    The CLSR trace emphasizes concise operator-like transformations and explicit state (e.g., variable bindings and subgoal markers), reducing verbosity while maintaining (and here improving) correctness.
    }
    \label{app-fig: Sp_q8b_math_s1_part1}
\end{figure*}

\begin{figure*}[ht]
    \centering
    \includegraphics[width=0.9\textwidth]{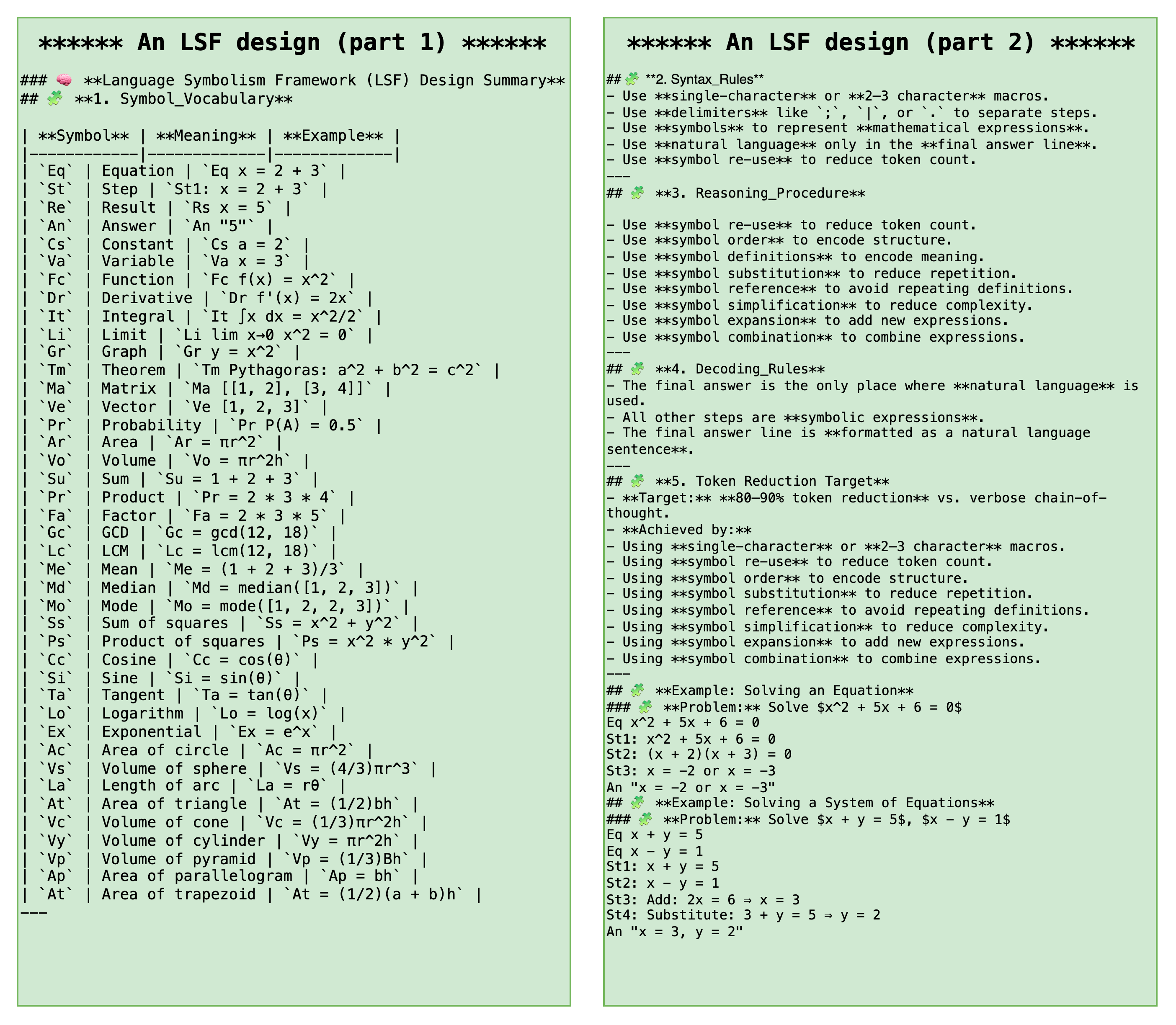}
    \caption{
    \textbf{Router-selected LSF specification for MATH500.}
    The automatically selected LSF content used by CLSR for the MATH500 query in Fig.~\ref{app-fig: Sp_q8b_math_s1_part1}.
    This example highlights that the router can choose a math-specialized LSF with compact transformation operators and lightweight verification tags, enabling high information density per generated token.
    }
    \label{app-fig: Sp_q8b_math_s1_part2}
\end{figure*}

\begin{figure*}[ht]
    \centering
    \includegraphics[width=0.9\textwidth]{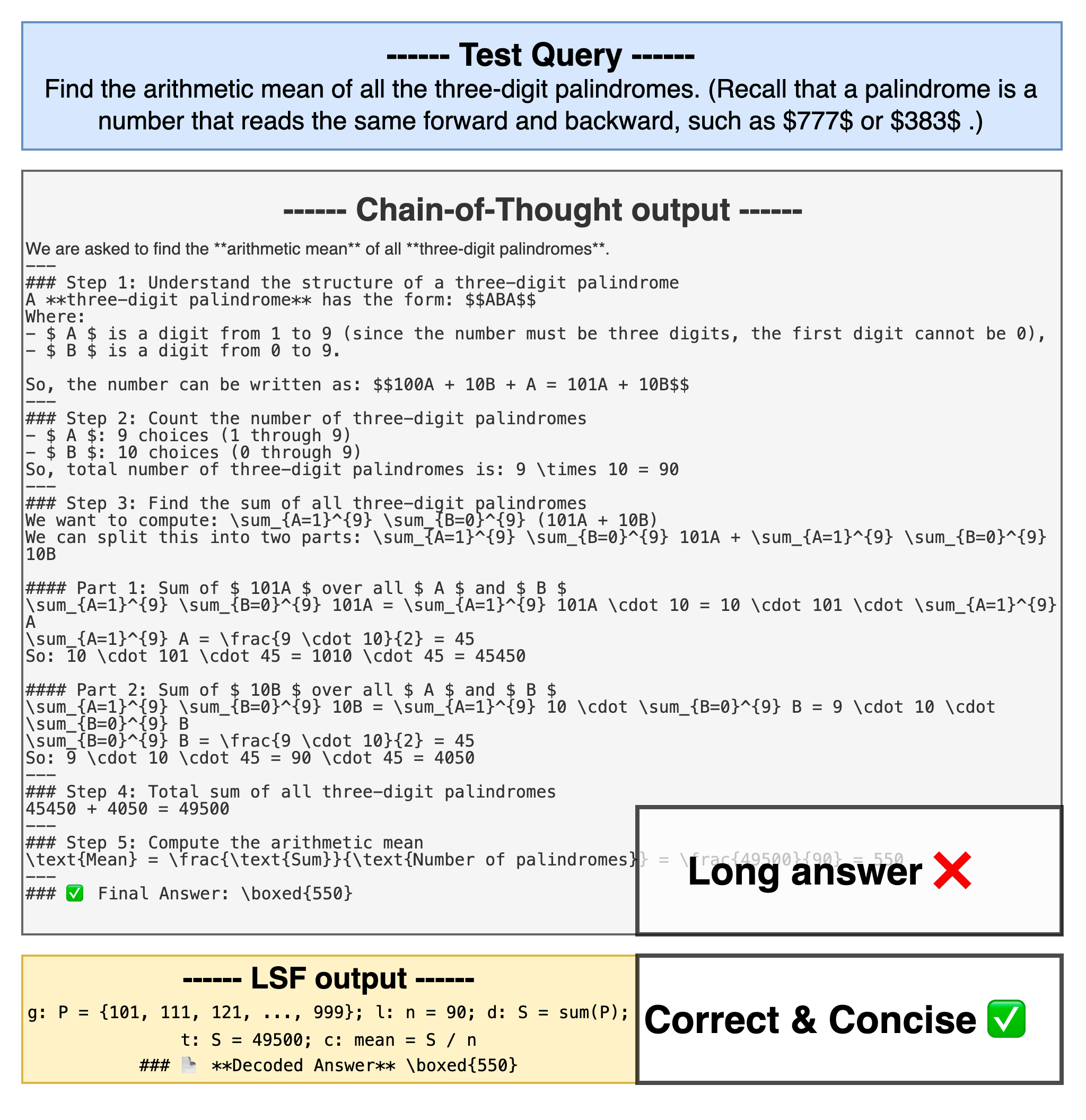}
    \caption{
    \textbf{Qualitative AIME example: CoT vs.\ CLSR (LSF).}
    A representative AIME test query comparing standard CoT with CLSR.
    CLSR produces a program-like symbolic trace that keeps only the computable core of reasoning, reducing narrative overhead and supporting reliable multi-step derivation under a constrained token budget.
    }
    \label{app-fig: Sp_q8b_aime_s1_part1}
\end{figure*}

\begin{figure*}[ht]
    \centering
    \includegraphics[width=0.9\textwidth]{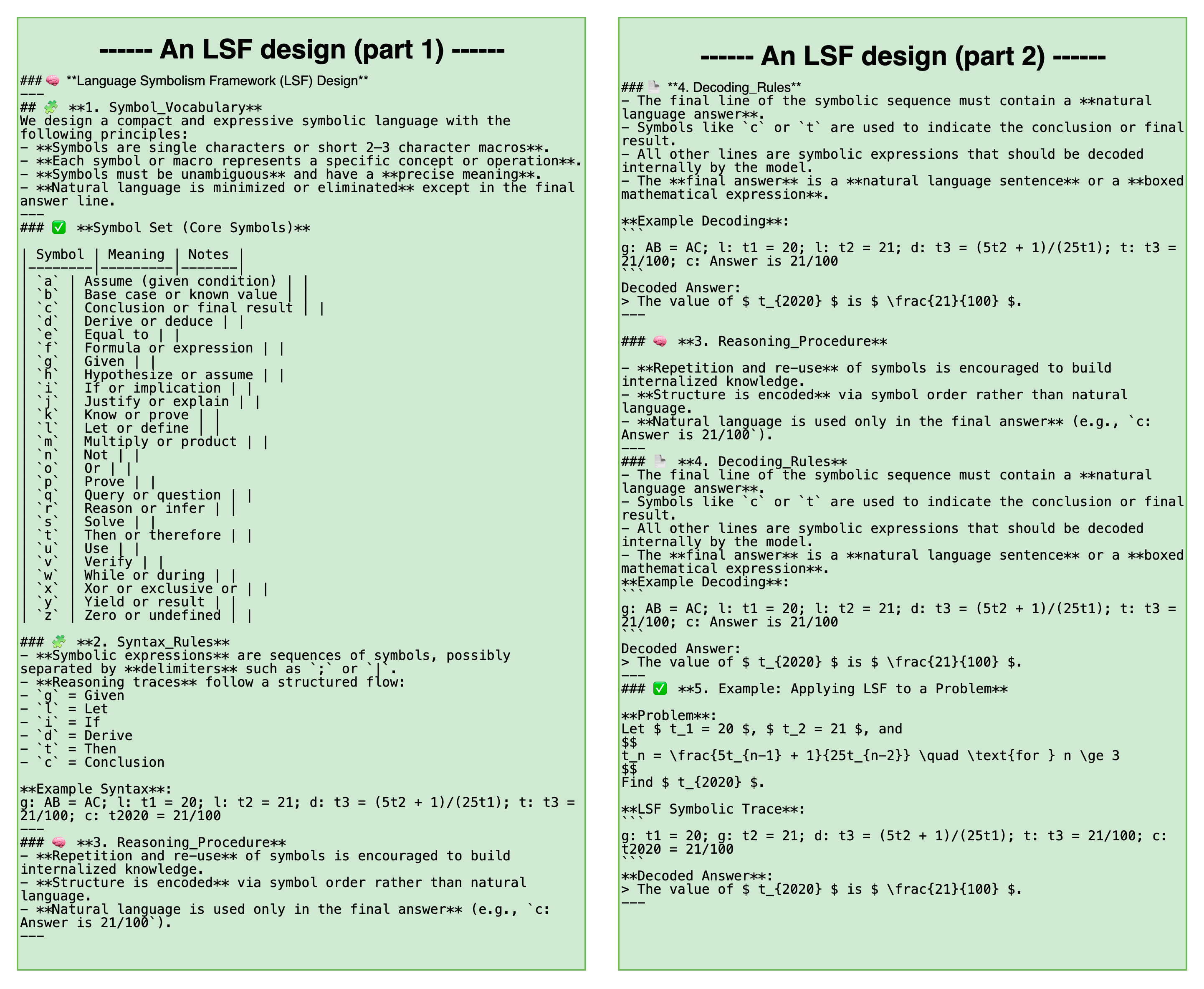}
    \caption{
    \textbf{Router-selected LSF specification for AIME.}
    The automatically selected LSF content used by CLSR for the AIME query in Fig.~\ref{app-fig: Sp_q8b_aime_s1_part1}.
    The chosen LSF reflects difficulty-aware protocol selection, instantiating stricter structure and more explicit intermediate state for harder problems, consistent with the theory that different operating points on the accuracy--token frontier should be selected adaptively.
    }
    \label{app-fig: Sp_q8b_aime_s1_part2}
\end{figure*}

Appendix Figures~\ref{app-fig: Sp_q8b_gpqa_s1_part1}--\ref{app-fig: Sp_q8b_aime_s1_part2} provide representative outputs of the GPQA, MATH500, and AIME.benchmarks.
Compared to verbose CoT, our CLSR (LSF) traces replace long natural-language rationales with compact symbolic operators and structured templates that preserve key intermediate state ({\it e.g.}, variable bindings, sub-goal markers, and verification tags) while discarding redundant narration.
We also show the auto-selected LSFs specification used for each query, illustrating that (i) the router selects different LSFs by domain and difficulty, and (ii) the selected LSFs are reusable artifacts rather than one-off prompt rewrites.

This section provides qualitative evidence for \emph{how} CLSR modifies the reasoning trace across refinement rounds.
Figures~\ref{app-fig: Sp_q8b_gpqa_s1_part1}--\ref{app-fig: Sp_q8b_aime_s1_part2} show representative model outputs for CoT and CLSR as well as the intermediate LSF produced by CLSR.
We highlight several recurring phenomena.

\paragraph{From narrative CoT to task-relevant \emph{latent symbolic form}.}
A consistent difference is that CoT tends to produce verbose narrative intermediate text that mixes (i) problem understanding, (ii) bookkeeping, and (iii) arithmetic/logical execution.
In contrast, CLSR compresses the intermediate reasoning into a compact LSF that is closer to a \emph{minimal sufficient representation} to solve the instance: 
it preserves variable bindings, constraints, and key transformation steps while discarding stylistic and redundant natural-language content.
This aligns with the goal of CLSR: 
to shift reasoning toward a higher signal-to-token ratio without relying on an external executor.

\paragraph{Error localization and correction across rounds.}
Across examples, later rounds frequently correct earlier-round failure modes that are common in plain CoT:
(i) missed constraints ({\it e.g.}, forgetting a boundary condition or a unit conversion),
(ii) inconsistent variable definitions,
(iii) arithmetic slips that propagate in long traces, and
(iv) premature commitment to an incorrect intermediate value.
The intermediate LSF makes such issues easier to localize: the model can explicitly re-check constraint satisfaction or recompute a small symbolic sub-part, instead of re-generating a long narrative chain.

\paragraph{Refinement is often \emph{selective} rather than \emph{additive}.}
Importantly, CLSR does not merely append more tokens.
Later-round traces often \emph{replace} earlier representations with shorter, cleaner ones:
irrelevant branches are removed, equivalent expressions are simplified, and the final answer emerges after a small number of symbolic edits.
This behavior explains why increasing $T$ does not necessarily inflate completion length proportionally and why an adaptive policy can stop at $T=1$ for many inputs.

\paragraph{When additional rounds help.}
The examples suggest that multiple rounds are most beneficial for problems with
(i) multi-constraint coupling (where one mistaken assumption breaks downstream steps),
(ii) long-range dependencies (where early variable choices constrain later operations),
or (iii) heavy arithmetic/combinatorial bookkeeping.
In these cases, CLSR uses additional rounds to re-encode the problem into a more stable LSF and derive the answer with fewer opportunities for cascading errors.

\paragraph{Limitations of qualitative traces.}
Finally, we emphasize that these traces are presented as \emph{algorithmic artifacts} of CLSR, {\it i.e.}, intermediate representations used to improve prediction under a token budget, not as guaranteed faithful explanations of internal model causality.
Nevertheless, they provide an interpretable window into how CLSR reallocates tokens from verbose narration to structured, constraint-preserving symbolic computation.

\subsection{Latency decomposition and token accounting}
\label{app:latency_token_accounting}

\noindent\textbf{Reference calibration (TTFT/TPOT).}
Regarding TTFT (time-to-first-token) and TPOT (time-per-output-token),
Table~\ref{tab:ttft_tpot_reference} reports representative single-request latency components for Qwen3-8B on RTX 4090 GPU.
We use these measurements as a sanity-check that, for long-form reasoning outputs, decoding latency scales approximately linearly with the number of generated tokens and typically dominates prompt prefill overhead.

\begin{table*}[ht]
  \caption{
\textbf{Representative latency decomposition for Qwen3-8B inference (single request, batch size 1).}
$L_{\text{in}}$ and $L_{\text{out}}$ denote prompt and completion lengths.
TTFT primarily reflects prompt prefill plus the first decode step; 
TPOT reflects steady-state decoding;
RTI denotes the ratio of total latency over token generation (decoding) time.
  }
  \label{tab:ttft_tpot_reference}
  \begin{center}
    \begin{small}
      \begin{sc}
        \begin{tabular}{@{}ccccc@{}}
        \toprule
        $L_\text{in}$ & $L_\text{out}$ & TTFT (ms) & TPOT (ms/tkn) & RTI \\ \midrule
        \multirow{3}{*}{512} & 64 & \multirow{3}{*}{76} & \multirow{3}{*}{11.2} & 1.106 \\
         & 256 &  &  & 1.026 \\
         & 1024 &  &  & 1.007 \\ \midrule
        \multirow{3}{*}{1024} & 64 & \multirow{3}{*}{148} & \multirow{3}{*}{11.9} & 1.194 \\
         & 256 &  &  & 1.048 \\
         & 1024 &  &  & 1.012 \\ \midrule
        \multirow{3}{*}{2048} & 64 & \multirow{3}{*}{281} & \multirow{3}{*}{12.9} & 1.340 \\
         & 256 &  &  & 1.085 \\
         & 1024 &  &  & 1.021 \\ \bottomrule
        \end{tabular}
      \end{sc}
    \end{small}
  \end{center}
  \vskip -0.1in
\end{table*}

The LLM-router often takes dozens of input tokens to determine the LSF-categories, then takes hundreds of input tokens to read the LSF-profiles and determine which LSFs should be selected;
reading a complete LSF often consumes hundreds of tokens.
Empirically, the total input token $L_\text{in}$ during inference for different benchmarks often ranges from $0.5\sim2\times10^3$.

\paragraph{Cache-aware token-equivalent accounting.}
The main paper reports generated completion tokens because they are a useful latency-oriented proxy for reasoning workloads.
However, an LSF card also contributes input tokens.
In a serving system with prompt-prefix caching, a fixed bank of LSF cards can be arranged in canonical order as a reusable prefix, while only the query, compact routing suffix, and generated outputs vary across requests.
To make this explicit, we define a cache-aware token-equivalent cost
\[
C_{\mathrm{eq}}
=
C_{\mathrm{out}}
+
\gamma_{\mathrm{in}} C_{\mathrm{in}}^{\mathrm{uncached}}
+
\gamma_{\mathrm{cache}} C_{\mathrm{in}}^{\mathrm{cached}},
\]
where $C_{\mathrm{out}}$ is the generated-token cost, $C_{\mathrm{in}}^{\mathrm{uncached}}$ is the non-reused input-token cost, $C_{\mathrm{in}}^{\mathrm{cached}}$ is the reusable prefix-token cost, and $\gamma_{\mathrm{in}},\gamma_{\mathrm{cache}}\in[0,1]$ convert input tokens into output-token equivalents.
The specific coefficients depend on the model provider and serving infrastructure.
Therefore, this metric should be interpreted as a deployment-oriented diagnostic rather than the primary algorithmic objective.

Table~\ref{tab:cache_aware_tokens} reports representative cache-aware comparisons for Qwen3-8B.
The results show that CLSR remains competitive even when reusable prompt-prefix overhead is included.

\begin{table}[t]
  \caption{
  \textbf{Cache-aware token-equivalent accounting on Qwen3-8B.}
  Generated tokens count online emitted tokens.
  Cache-aware tokens additionally include input overhead under a reusable-prefix assumption.
  }
  \label{tab:cache_aware_tokens}
  \begin{center}
  \begin{scriptsize}
  \begin{sc}
  \begin{tabular}{@{}llccc@{}}
  \toprule
  Benchmark & Method & Acc. & Gen. Tkn & Cache-aware Tkn \\
  \midrule
  \multirow{3}{*}{MMLU-Pro}
  & CoT & 60.5 & 312 & 331 \\
  & SoT & 58.5 & 120 & 144 \\
  & CLSR & 60.3 & \textbf{93} & \textbf{121} \\
  \midrule
  \multirow{3}{*}{MATH500}
  & CoT & \textbf{86.8} & 872 & 890 \\
  & SoT & 81.4 & 301 & 322 \\
  & CLSR & 86.6 & \textbf{245} & \textbf{283} \\
  \bottomrule
  \end{tabular}
  \end{sc}
  \end{scriptsize}
  \end{center}
  \vskip -0.15in
\end{table}

\paragraph{Offline cost amortization.}
CLSR pays an offline protocol-discovery cost to generate, evaluate, and refine the LSF pool.
This cost is conceptually similar to automatic prompt/protocol optimization methods that search over candidate instructions, demonstrations, or policies before deployment.
The practical motivation is different from self-consistency or Tree-of-Thoughts-style inference: CLSR trades one-time offline search for lower repeated serving-time reasoning cost.
If an evolved LSF pool is reused for many queries in the same benchmark/domain family, the amortized offline cost per query decreases as the number of served queries grows.
Thus, CLSR is most appropriate for repeated-use reasoning services, agent backends, and domain-specific deployments, rather than one-off queries where no amortization is possible.
\label{app-subsec: offline_amortization}

\clearpage

\section{Prompt Templates for LSF Synthesis, Mutation, and Routing}
\label{app-sec: prompt_templates}

This appendix provides the prompt templates used by CLSR.
The templates are intentionally lightweight: the LLM is given the optimization goal and a compact schema, but the concrete symbolic protocol is not manually authored.

\textbf{LSF synthesis prompt.}
Given a set of exemplar question--answer pairs with reasoning traces, we use the following template:
\begin{quote}
\small
Please design a Language Symbolism Framework (LSF) based on the exemplars in the chat history to minimize the number of tokens while maintaining the reasoning capacity.
An LSF is a symbolic communication code comprising:
(i) symbol naming, namely a compact lexicon;
(ii) syntax, namely a compositional grammar; and
(iii) constraints, namely well-formedness and usage rules.
The LSF should help an LLM solve similar problems with concise symbolic reasoning.
Return the LSF specification, including the symbol table, grammar, reasoning protocol, answer format, and failure checks.
\end{quote}

\textbf{LSF mutation prompt.}
After evaluating a generation of LSFs, we select high-leverage traces that are both correct and token-efficient.
The mutation prompt is:
\begin{quote}
\small
You are given several parent LSFs and their high-leverage solution traces.
A high-leverage trace is correct and uses fewer tokens than competing traces for the same query.
Refine or recombine the parent LSFs to produce a next-generation LSF.
Keep symbolic conventions that support correct concise reasoning; remove redundant notation; repair ambiguous rules; and add only minimal verification tags when they improve reliability.
Do not simply rewrite the example answers.
Return a reusable LSF specification that can be applied to unseen queries.
\end{quote}

\textbf{LSF profile summarization prompt.}
For each candidate LSF $S_k$, we summarize its empirical profile from validation rollouts:

\begin{quote}\small
Given an LSF specification and its validation rollouts, summarize its profile for routing.
Include:
(i) query types where this LSF performs well;
(ii) validation accuracy;
(iii) median generated-token footprint;
(iv) common failure modes;
(v) reliability under different problem difficulties; and
(vi) a short descriptor that can be used by an LLM-router.
Keep the descriptor compact and factual.
\end{quote}

\textbf{Router-card distillation prompt.}
To make routing both compact and reproducible, the above profile is then distilled into a short card consumed by the LLM-router:
\begin{quote}\small
Given an LSF specification and its empirical profile, output exactly one compact router card in the following format:
\texttt{[ID] TAG | perf:<acc>/<tok> | IO | STOP}
where \texttt{acc} is the validation accuracy rounded to two decimals,
\texttt{tok} is the median generated-token count,
\texttt{IO} is a short output-schema descriptor,
and \texttt{STOP} is a short stopping-contract descriptor.
Do not output explanations.
\end{quote}

This produces cards like
\texttt{[ALG1] linear/eq | perf:.81/42 | out:ans,brief | stop:conf>=0.6},
which expose the empirical accuracy--cost profile to the router while keeping the prompt prefix compact.

\textbf{Stage-1 category-routing prompt.}
At test time, we use a category-routing prompt to prune the candidate LSF pool:

\begin{quote}\small
You are an LSF category router.
Given the query, output exactly one line in the format

\texttt{C:<category-list>}

where \texttt{<category-list>} contains one or two comma-separated category codes chosen from the predefined vocabulary.
Use a broad backup category when the query is ambiguous.
Do not output explanations.
\end{quote}

For example, the router may emit
\texttt{C:ALG},
\texttt{C:PHY,ALG},
or
\texttt{C:QA,HIS}.

\textbf{Stage-2 protocol-planning prompt.}
Given the query and the compact LSF cards selected by Stage~1, the router emits a structured execution plan:

\begin{quote}\small
You are an LSF protocol router.
Given the query and the available LSF cards, choose the cheapest reasoning protocol that is likely to preserve correctness.
Output exactly one line in the following format:

\texttt{M:<mode>;L:<lsf-list>;R:<round-spec>;K:<n>;A:<agg>;S:<stop>}

Use only LSF IDs that appear in the provided cards.
Choose
\texttt{M:S} for a single-LSF direct answer,
\texttt{M:A} for multi-LSF aggregation,
and
\texttt{M:C} for implicit LSF composition.
The field \texttt{R} specifies the round structure or composition order;
\texttt{K} specifies the number of parallel samples per LSF when aggregation is used;
\texttt{A} specifies the aggregation rule;
and \texttt{S} specifies the stopping criterion.
Do not output explanations or any text outside the plan line.
\end{quote}

\textbf{Why prompt templates do not make CLSR prompt optimization.}
These templates specify the task interface, the empirical profile exposed to the router, and the structured control format, but not the concrete symbolic language itself.
The optimized object is the reusable LSF specification and its routing behavior, not a per-query natural-language instruction string.
The same LSFs can be reused across many unseen queries, selected by a router, and composed with other LSFs in later rounds.

% \paragraph{LSF profile summarization prompt.}
% For each candidate LSF $\mathcal{S}_{k}$, we summarize its empirical profile:
% \begin{quote}
% \small
% Given an LSF specification and its validation rollouts, summarize its profile for routing.
% Include:
% (i) query types where this LSF performs well;
% (ii) typical generated-token footprint;
% (iii) common failure modes;
% (iv) reliability under different problem difficulties; and
% (v) a short descriptor that can be used by an LLM-router.
% Keep the descriptor compact.
% \end{quote}

% \paragraph{Router planning prompt.}
% At test time, the router receives a query $x$ and a set of compact LSF descriptors:
% \begin{quote}
% \small
% You are an LSF router.
% Given the query and the available LSF descriptors, choose a low-cost reasoning protocol.
% Select one of the following modes:
% Single LSF direct answer, Multi-LSF aggregation, or Implicit LSF composition.
% Return:
% mode;
% selected LSF ids;
% number of rounds;
% round-by-round LSF calls;
% aggregation rule;
% stopping rule;
% and token-budget hint.
% Prefer the cheapest protocol that is likely to preserve correctness.
% \end{quote}

% \paragraph{Why prompt templates do not make CLSR prompt optimization.}
% These templates specify the task interface and the selection objective, but not the concrete symbolic language.
% The optimized object is the reusable LSF specification and its routing behavior, not a per-query natural-language instruction string.
% The same LSFs can be reused across many unseen queries, selected by a router and composed with other LSFs in later rounds.

%\clearpage
\section{LLM-Router: Compact Plan Format and Execution Semantics}
\label{app-sec: router_plan}

\paragraph{Design goals.}
Our router must 
(i) emit a \emph{parseable} plan with near-zero ambiguity, 
(ii) keep the router output within \emph{tens of tokens}, and
(iii) support the three protocol-planning modes in our framework:
\textsc{Single} (one LSF direct answer), \textsc{Aggregate} (multi-LSF ensemble), and \textsc{Compose} (implicit LSF composition).
To ensure reliability, we recommend \emph{constrained structured decoding} ({\it e.g.}, EBNF/regex or lightweight schema constraints) so that the router can only produce syntactically valid plans.

\paragraph{Cost-aware routing principle.}
The router is deliberately conservative.
It first restricts the candidate LSF pool through lightweight category routing, and then selects the cheapest protocol that is likely to preserve correctness.
This realizes the accuracy--token objective of Section~2.1 at the protocol level:
\textsc{Single} is the default low-cost route,
whereas \textsc{Aggregate} and \textsc{Compose} are invoked only when the available LSF cards indicate that additional redundancy, verification, or staged symbolic processing is likely to improve reliability.

\subsection{Two-stage routing (categories $\rightarrow$ protocol plan)}
\label{subsec:two_stage_routing}

\noindent\textbf{Stage 1 (Category routing).}
Given a query $q$, the router emits a short \emph{category code list} $\mathcal{C}(q)$ to limit the candidate LSF pool.
We predefine a vocabulary of $\sim$30--60 category codes ({\it e.g.}, \texttt{ALG}, \texttt{GEO}, \texttt{PHY}, \texttt{CHE}, \texttt{LOG}, \texttt{QA}, \texttt{CODE}, \texttt{STAT}, \texttt{NUM}, \texttt{PROB}, \texttt{BIO}, \texttt{HIS}, \texttt{LAW}, \texttt{FIN}, etc.).
The output is \emph{one line}, typically $<10$ tokens:

\vspace{-1mm}
\begin{center}
\small
\texttt{C:ALG} \quad or \quad \texttt{C:PHY,ALG} \quad or \quad \texttt{C:QA,HIS}
\end{center}
\vspace{-2mm}

\noindent\textbf{Stage 2 (Protocol planning).}
We then feed the router the query $q$ plus the \emph{LSF cards} belonging to selected categories $\mathcal{C}(q)$.
The router outputs a compact plan $\mathcal{P}(q)$ that selects concrete LSF(s) and the inference mode.

\subsection{LSF card format (compact descriptors)}

Each LSF is represented to the router by a single short card that is easy to scan while remaining compact enough to keep the input overhead small.
The card is distilled from the validation profile summarized in Appendix~\ref{app-sec: prompt_templates} and follows the schema

\begin{center}
\texttt{[ID] TAG | perf:<acc>/<tok> | IO | STOP}
\end{center}

where
\texttt{TAG} is a short semantic hint,
\texttt{acc} is the held-out validation accuracy rounded to two decimals,
\texttt{tok} is the median generated-token count for that LSF,
\texttt{IO} is an output-schema hint,
and
\texttt{STOP} specifies the stopping contract.

Example cards are:

\begin{quote}\small
\texttt{[ALG1] linear/eq | perf:.81/42 | out:ans,brief | stop:conf>=0.6}

\texttt{[ALG2] simplify/check | perf:.77/55 | out:ans,check | stop:valid}

\texttt{[GEO1] geometry-parse | perf:.74/63 | out:vars,eqs | stop:complete}

\texttt{[PHY1] mechanics | perf:.79/58 | out:eqs,ans | stop:units-ok}

\texttt{[QA1] decompose | perf:.72/71 | out:subq,ans | stop:all-supported}
\end{quote}

This format deliberately exposes a minimal empirical accuracy--cost signal to the router.
The full validation profile is used only offline to form the card, so the online planning prompt remains short while retaining the information required for budget-aware protocol selection.

% \subsection{LSF card format (compact descriptors)}
% \label{subsec:lsf_cards}

% Each LSF is described by a single short ``card'' line that is easy for the router to scan, but compact enough to keep the input overhead small.
% We use the following minimal schema:
% \[
% \texttt{[ID]~TAG~|~IO~|~STOP}
% \]
% where \texttt{TAG} is a 2--5 token hint, \texttt{IO} is an output schema hint, and \texttt{STOP} indicates the stopping contract.
% Example cards (each typically $\sim$10-20 tokens):

% \begin{quote}\small
% \texttt{[ALG1] linear/eq | out:ans,brief | stop:conf>=0.6}\\
% \texttt{[ALG2] simplify/check | out:ans,check | stop:valid}\\
% \texttt{[GEO1] geometry parse | out:vars,eqs | stop:complete}\\
% \texttt{[GEO2] coord trick | out:ans,sketch | stop:conf>=0.55}\\
% \texttt{[PHY1] mechanics | out:eqs,ans | stop:units-ok}\\
% \texttt{[QA1] decompose | out:subq,ans | stop:all-supported}
% \end{quote}

\subsection{Router plan format (tens of tokens)}

The router emits a single-line plan in a compact DSL:

\begin{center}
\texttt{M:<mode>;L:<lsf-list>;R:<round-spec>;K:<n>;A:<agg>;S:<stop>}
\end{center}

We use the following short enumerations.

\paragraph{Mode.}
\texttt{M} specifies the protocol family:
\[
\texttt{M} \in \{\texttt{S},\texttt{A},\texttt{C}\},
\]
corresponding to
\textsc{Single},
\textsc{Aggregate},
and
\textsc{Compose}.

\paragraph{Selected LSFs.}
\texttt{L} is a comma-separated list of selected LSF IDs, e.g.,
\texttt{L:ALG1}
or
\texttt{L:QA1,QA3,HIS1}.

\paragraph{Round structure.}
\texttt{R} specifies the execution structure.
For a one-shot single or aggregation protocol, we use \texttt{R:1}.
For repeated refinement, we may use \texttt{R:2} or \texttt{R:3}.
For implicit composition, \texttt{R} explicitly encodes the ordered LSF stages, e.g.,
\texttt{R:PHY1>ALG1}
or
\texttt{R:QA1>HIS1}.

\paragraph{Sampling and aggregation.}
\texttt{K} is the number of parallel samples per LSF in aggregation mode, with default value $K=1$.
\texttt{A} is the aggregation rule, e.g.,
\texttt{mv} for majority vote,
\texttt{sc} for self-consistency-style agreement,
or
\texttt{w} for a router-weighted aggregation rule when such weights are available.

\paragraph{Stopping rule.}
\texttt{S} is a short stopping criterion, e.g.,
\texttt{ans} for stopping after a valid answer is parsed,
\texttt{valid} for schema-valid completion,
\texttt{units-ok} for a domain-specific check,
or
\texttt{conf0.7} for a confidence threshold when supported by the LSF contract.

\paragraph{Token budget.}
A plan such as
\texttt{M:A;L:QA1,QA3;R:1;K:2;A:mv;S:ans}
typically remains within a few tens of generated tokens.
The format is intentionally compact so that the router overhead is negligible relative to the downstream reasoning protocol while remaining fully parseable.

% \subsection{Router plan format (tens of tokens)}
% \label{subsec:plan_format}

% The router emits a single-line plan in a compact DSL:

% \vspace{-1mm}
% \begin{center}
% \small
% \texttt{M:<mode>;L:<lsf-list>;K:<n>;A:<agg>;S:<stop>}
% \end{center}
% \vspace{-2mm}

% \noindent where fields are optional except \texttt{M} and \texttt{L}. We use short enumerations:
% \[
% \texttt{mode}\in\{\texttt{S},\texttt{A},\texttt{C}\}
% \]
% corresponding to \textsc{Single}, \textsc{Aggregate}, and \textsc{Compose}.
% \texttt{lsf-list} is a comma-separated list of IDs (e.g., \texttt{ALG1,GEO2}).
% \texttt{K} is the number of parallel samples per LSF in aggregation mode (default 1).
% \texttt{A} is the aggregation rule (e.g., \texttt{mv}=majority vote, \texttt{sc}=self-consistency, \texttt{w}=router-weighted).
% \texttt{S} is a short stopping key (e.g., \texttt{1} means single round; \texttt{2} means up to 2 rounds; \texttt{conf0.7} means confidence gate).

% \paragraph{Token budget.}
% A typical plan such as \texttt{M:A;L:QA1,QA3;K:2;A:mv;S:2} is usually $\sim$15--25 tokens.

\subsection{Execution semantics (deterministic interpreter)}

Given the category output $C(q)$ and plan output $P(q)$, CLSR is executed deterministically as follows.

\begin{enumerate}[leftmargin=*, itemsep=1pt]
    \item \textbf{Candidate construction.}
    Build the candidate LSF set
    \[
    \mathcal{L}(q)
    =
    \bigcup_{c \in C(q)}
    \mathrm{LSFs}(c),
    \]
    and assemble the corresponding compact LSF cards.

    \item \textbf{Plan parsing.}
    Parse the router plan
    \[
    P(q)
    =
    (\texttt{mode},\texttt{lsfs},\texttt{rounds},
    \texttt{K},\texttt{agg},\texttt{stop}).
    \]
    Invalid plans are rejected and the router is re-asked under constrained structured decoding.
    A plan is invalid if it violates the DSL grammar, references an unavailable LSF ID, or uses a mode-incompatible field configuration.

    \item \textbf{Protocol execution.}
    \begin{itemize}[leftmargin=*, itemsep=1pt]
        \item \textsc{Single} (\texttt{M:S}):
        instantiate the selected LSF and run a direct solver call.
        The default one-shot case uses \texttt{R:1};
        larger values of \texttt{R} indicate bounded repeated refinement under the same LSF.

        \item \textsc{Aggregate} (\texttt{M:A}):
        for each selected LSF, sample $K$ solver completions in parallel,
        then aggregate the resulting answers according to \texttt{A}.
        The default one-stage ensemble uses \texttt{R:1}.

        \item \textsc{Compose} (\texttt{M:C}):
        execute the LSFs sequentially according to the ordering encoded in \texttt{R}.
        The parsed symbolic state emitted by the current LSF becomes the input state for the next LSF.
        For example, \texttt{R:PHY1>ALG1} first constructs a physics equation state and then invokes an algebraic solver state.
    \end{itemize}

    \item \textbf{Early stopping.}
    After each solver call or composition stage, evaluate the stopping rule \texttt{S}.
    If the criterion is satisfied, terminate immediately; otherwise continue until the prescribed round structure is exhausted.

    \item \textbf{Accounting.}
    All online LLM outputs are counted toward generated-token cost, including category-routing outputs, protocol-planning outputs, intermediate solver outputs, and aggregation-related outputs.
\end{enumerate}

% \subsection{Execution semantics (deterministic interpreter)}
% \label{subsec:execution_semantics}

% Given category output $\mathcal{C}(q)$ and plan output $\mathcal{P}(q)$, we execute CLSR deterministically:

% \vspace{-1mm}
% \begin{enumerate}\itemsep1pt
% \item \textbf{Candidate construction.} Build candidate set $\mathcal{L}=\bigcup_{c\in\mathcal{C}(q)} \mathrm{LSFs}(c)$ and assemble the corresponding LSF cards.
% \item \textbf{Plan parsing.} Parse the router plan $\mathcal{P}(q)$ into $(\texttt{mode}, \texttt{lsfs}, K, \texttt{agg}, \texttt{stop})$.
% Invalid plans are rejected (and the router is re-asked under constrained decoding).
% \item \textbf{Protocol execution.}
% \begin{itemize}\itemsep1pt
% \item \textbf{\textsc{Single} (\texttt{M:S}).} Instantiate the chosen LSF template and run one solver call (or multiple rounds if \texttt{S:$>$1}).
% \item \textbf{\textsc{Aggregate} (\texttt{M:A}).} For each selected LSF, sample $K$ solver completions in parallel; aggregate via \texttt{A} (e.g., \texttt{mv}/\texttt{sc}/\texttt{w}) to produce the final answer.
% \item \textbf{\textsc{Compose} (\texttt{M:C}).} Execute LSFs sequentially; the parsed state from LSF$_i$ becomes the input state for LSF$_{i+1}$ (implicit composition). Stop early if the current LSF meets \texttt{S}.
% \end{itemize}
% \item \textbf{Accounting.} All LLM outputs are counted toward generated-token cost, including router tokens and intermediate outputs used for aggregation.
% \end{enumerate}

\subsection{End-to-end samples}

\subsubsection{Sample: single-LSF direct answer}

\paragraph{Query.}
\texttt{Solve $2x+3=11$.}

\paragraph{Stage 1: category routing.}
\texttt{C:ALG}

\paragraph{Stage 2 input: card excerpt.}
\begin{quote}\small
\texttt{[ALG1] linear/eq | perf:.81/42 | out:ans,brief | stop:conf>=0.6}

\texttt{[ALG2] simplify/check | perf:.77/55 | out:ans,check | stop:valid}
\end{quote}

\paragraph{Stage 2 output: plan.}
\texttt{M:S;L:ALG1;R:1;K:1;A:none;S:ans}

\paragraph{Execution.}
Instantiate \texttt{ALG1}, run one solver call, parse the final answer, and stop immediately after \texttt{out:ans} is produced.

\subsubsection{Sample: multi-LSF aggregation}

\paragraph{Query.}
\texttt{Which author wrote the book that inspired the movie X, and what year was it published?}

\paragraph{Stage 1: category routing.}
\texttt{C:QA,HIS}

\paragraph{Stage 2 input: card excerpt.}
\begin{quote}\small
\texttt{[QA1] decompose | perf:.72/71 | out:subq,ans | stop:all-supported}

\texttt{[QA3] evidence-first | perf:.75/84 | out:cites,ans | stop:cited}

\texttt{[HIS1] year-check | perf:.69/48 | out:year,src | stop:match}
\end{quote}

\paragraph{Stage 2 output: plan.}
\texttt{M:A;L:QA1,QA3,HIS1;R:1;K:2;A:mv;S:ans}

\paragraph{Execution.}
Run $K=2$ samples per selected LSF in parallel, extract candidate answers, aggregate by majority vote, and stop after a valid final answer is selected.

\subsubsection{Sample: implicit LSF composition}

\paragraph{Query.}
\texttt{A 2 kg mass is pulled with 10 N on a frictionless surface. Find acceleration.}

\paragraph{Stage 1: category routing.}
\texttt{C:PHY,ALG}

\paragraph{Stage 2 input: card excerpt.}
\begin{quote}\small
\texttt{[PHY1] mechanics | perf:.79/58 | out:eqs,ans | stop:units-ok}

\texttt{[ALG1] linear/eq | perf:.81/42 | out:ans,brief | stop:conf>=0.6}
\end{quote}

\paragraph{Stage 2 output: plan.}
\texttt{M:C;L:PHY1,ALG1;R:PHY1>ALG1;K:1;A:none;S:units-ok}

\paragraph{Execution.}
\texttt{PHY1} first emits a minimal equation state, e.g., $a=F/m$.
The parsed symbolic state is then passed to \texttt{ALG1}, which performs the substitution and emits the final answer.
The execution stops once the answer is parsed and the domain-specific consistency check is satisfied.

\clearpage
\section{Appendix: CLSR outputs and LSF samples}

This section presents representative \emph{raw model completions} for both $N$-shot CoT and CLSR on the benchmarks used in the main paper.
For CLSR, we additionally show the intermediate traces produced in a refinement round, making explicit how the latent symbolic form (LSF) is iteratively compressed, verified, and corrected before producing the final answer.
These examples are intended to complement the quantitative results by illustrating the qualitative differences in \emph{trace structure}, \emph{error patterns}, and \emph{token usage} between natural-language CoT and CLSR's symbolic compression mechanism.

\subsection{A sample on Scientific domains}

\clearpage
\subsection{A sample on Mathematical domains}

\clearpage
\subsection{A sample on AIME benchmark}

\clearpage
\section{Appendix: More Theoretical Validation for CLSR}
\label{app-sec: more theoretical validation}

This appendix provides formal statements and full proofs for the theory in the main text (Sec.~\ref{sec: theory}).
We focus on two pillars: 
(i) an information-theoretic lower bound relating achievable accuracy to the expected number of generated tokens under arbitrary symbolism, and 
(ii) a subsumption result showing that multi-round multi-LSF CLSR protocols generalize program-execution pipelines under an explicit ``interpreter realizability'' premise.

\subsection{The Principle of Least Effort and Iterated Learning}
\label{app-subsec: zipf_iterated}

The sociolinguistic behavior of LSFs can be understood through two complementary pressures.
First, Zipf's Principle of Least Effort~\citep{kanwal2017zipf} argues that communication systems balance successful transmission against production cost.
Under this pressure, frequently used meanings or operations tend to receive shorter forms, while rare or ambiguous meanings may preserve longer forms for reliability.
CLSR instantiates an analogous pressure at test time: a symbolic convention survives if it helps the LLM solve problems correctly while reducing generated tokens.
This explains why evolved LSFs often introduce short operators for recurring reasoning moves, such as variable binding, subgoal decomposition, unit conversion, answer verification, or contradiction checking.

Second, the evolutionary loop resembles iterated learning.
Each generation observes a bottlenecked set of successful traces from the previous generation and must reconstruct a reusable symbolic system from them.
Iterated learning theory predicts that repeated transmission through such a bottleneck can produce languages that become more structured, compressible, and aligned with the learner's inductive biases.
In CLSR, the learner is a pretrained LLM, so the resulting LSFs are shaped by both task-level selection and the model's existing linguistic/mathematical priors.
This is why successful LSFs are rarely arbitrary character strings.
They often converge toward partially interpretable symbolic conventions: terse enough to reduce token cost, but structured enough for the model to reliably execute.

This perspective also clarifies the role of multi-agent interaction.
Different agents propose different ``dialects'' because they see different exemplars, random seeds, and high-leverage traces.
Selection then acts as a cultural filter: dialects that are concise but brittle disappear, while dialects that preserve correctness under compression are retained.
The final router performs a pragmatic form of code-switching.
It can select a strict symbolic dialect for easy algebraic queries, a softer dialect for knowledge-intensive queries, or a multi-dialect composition for hard problems requiring verification.
Thus, CLSR operationalizes a small-scale society of machine dialects: symbolic protocols emerge, compete, merge, and are selected according to their communicative utility for reasoning.

This interpretation also prevents an overstatement of our claim.
We do not claim that LSFs are invented independently of human language.
The base LLM and the benchmarks are human-derived, and the LSF synthesis prompt specifies the high-level goal.
The novelty is that the concrete protocol is not hand-written by the authors and is instead discovered through repeated use and selection.
The partial convergence of evolved LSFs toward human-interpretable notation is therefore not a flaw, but evidence that efficient symbolic systems can occupy a middle ground between natural language and opaque machine codes.

\subsection{Preliminaries: Interactive inference as a finite-horizon CMDP}

\paragraph{Query-conditioned setting.}
Fix a query $x$ and consider the conditional label distribution $Y \sim P(\cdot \mid X=x)$ supported on a finite \emph{effective answer set} $\mathcal{Y}_x$ with $|\mathcal{Y}_x|\ge 2$. 
For tasks with very large or continuous answer spaces, one can replace $\mathcal{Y}_x$ by a finite $\varepsilon$-net / evaluator-induced equivalence classes; see Remark~\ref{rem:finiteY}.
We measure correctness via the indicator $\mathbf{1}\{\widehat{Y}=Y\}$.

\paragraph{Token alphabet and LSF-conditioned decoding kernels.}
Let $\Sigma$ denote a finite alphabet of \emph{output tokens} (e.g., the model vocabulary; any LSF-specific surface form is representable as sequences in $\Sigma$).
We include a special termination token $\mathtt{STOP}\in\Sigma$.
Let $\mathcal{K}=\{1,\dots,K\}$ index the available LSFs. For each LSF $k\in\mathcal{K}$, the frozen backbone LLM with parameters $\theta$ induces a stochastic decoding kernel
\begin{equation}
\begin{gathered}
P_{\theta}^{(k)}(z \mid h, x), \qquad z\in\Sigma,
\end{gathered}
\end{equation}
where $h$ is the full interaction history (including which LSF was invoked, and all previously generated tokens). 
This is a well-defined conditional distribution (measurable stochastic kernel) over the next token.

\paragraph{Transcript state space and actions.}
We model CLSR as a token-level interactive process of maximum length $T_{\max}<\infty$.
At micro-step $t\in\{1,\dots,T_{\max}\}$, the \emph{state} is the current history $H_t=h_t$, where $h_1$ is empty and $h_{t+1} = (h_t, a_t, Z_t)$ appends the selected action and the realized token.
The \emph{action} is either selecting an LSF $a_t\in\mathcal{K}$ (meaning ``generate one token under LSF $a_t$''), or selecting a stop action $a_t=\mathtt{stop}$.
If $a_t=\mathtt{stop}$, we deterministically emit $Z_t=\mathtt{STOP}$ and transition into an absorbing terminal mode for all future steps.
Any higher-level ``multi-round'' protocol that calls multiple LSFs per round can be unrolled into such a micro-step sequence; this only affects constants and not the conclusions.

\paragraph{Cost and prediction.}
Define per-step token cost
\begin{equation}
\begin{gathered}
c(H_t,a_t) \;=\; \mathbf{1}\{a_t\in\mathcal{K}\},
\end{gathered}
\end{equation}
so the total token cost equals the number of non-\texttt{STOP} tokens:
\begin{equation}
\begin{gathered}
|T| \;=\; \sum_{t=1}^{T_{\max}} c(H_t,a_t).
\end{gathered}
\end{equation}
The final transcript is $T=(Z_1,\dots,Z_{T_{\max}})$ (padded with $\mathtt{STOP}$ after termination).
An aggregation/readout map $g$ (possibly implementing routing-then-aggregation, self-consistency, etc.) produces the final answer
\begin{equation}
\begin{gathered}
\widehat{Y} \;=\; g(T, x)\in\mathcal{Y}_x.
\end{gathered}
\end{equation}
We allow $g$ to be any (measurable) function; importantly, it does not use external oracles beyond the generated transcript.

\paragraph{Policies and objectives.}
A possibly randomized and history-dependent policy $\pi$ specifies distributions $\pi_t(\cdot \mid h_t, x)$ over actions at each step.
For a fixed $x$, define
\begin{equation}
\begin{gathered}
J_C(\pi;x)\;=\;\mathbb{E}_{\pi}\big[|T|\mid X=x\big],
\qquad
J_A(\pi;x)\;=\;\mathbb{P}_{\pi}\big[\widehat{Y}=Y\mid X=x\big].
\end{gathered}
\end{equation}
For a token budget $B\ge 0$, define the optimal achievable accuracy
\begin{equation}
\begin{gathered}
A^*(B;x)\;=\;\sup_{\pi:\;J_C(\pi;x)\le B}\; J_A(\pi;x).
\end{gathered}
\end{equation}
This matches the main-text notion (Eq.~(3)), specialized to a fixed query $x$.

\begin{remark}[Finite state reduction]
Because $\Sigma$ is finite and the horizon is bounded by $T_{\max}$, the set of reachable histories is finite:
$\;|\mathcal{H}|\le \sum_{t=0}^{T_{\max}} (|\mathcal{K}|\,|\Sigma|)^t < \infty$.
Hence, CLSR can be treated as a \emph{finite-horizon CMDP} on a finite state space (histories).
\end{remark}

\subsection{Existence and Pareto characterization of optimal CLSR policies}

\begin{theorem}[Existence of an optimal budget-feasible CLSR policy]
\label{thm:existence}
Fix $x$.
Assume (i) $K<\infty$, (ii) $|\Sigma|<\infty$ with a designated $\mathtt{STOP}$ token, and (iii) $T_{\max}<\infty$.
Then for every budget $B\ge 0$, the supremum $A^*(B;x)$ is attained:
there exists a (possibly randomized) policy $\pi_B^*$ such that
\begin{equation}
\begin{gathered}
J_C(\pi_B^*;x)\le B,
\qquad
J_A(\pi_B^*;x)=A^*(B;x).
\end{gathered}
\end{equation}
\end{theorem}

\begin{theorem}[Lagrangian scalarization and boundary points of the frontier]
\label{thm:lagrange}
Under the assumptions of Theorem~\ref{thm:existence}, consider the scalarized objective
\begin{equation}
\begin{gathered}
J_\lambda(\pi;x)\;=\;\mathbb{E}_\pi\!\left[\mathbf{1}\{\widehat{Y}=Y\}-\lambda |T| \;\middle|\; X=x\right],\qquad \lambda\ge 0.
\end{gathered}
\end{equation}
Then:
\begin{itemize}
\item For every $\lambda\ge 0$, there exists an optimal (history-Markov) \emph{deterministic} finite-horizon policy $\pi_\lambda^*$ that maximizes $J_\lambda(\pi;x)$.
\item Every \emph{supported} boundary point of the Pareto frontier $\{(J_C(\pi;x),J_A(\pi;x))\}$ can be implemented by some $\pi_\lambda^*$.
\item Any boundary point of the constrained problem can be achieved by a mixture of at most two scalarized optima $\pi_{\lambda_1}^*,\pi_{\lambda_2}^*$ with adjacent multipliers, i.e., randomized only at $t=1$ between two deterministic policies.
\end{itemize}
\end{theorem}

\paragraph{Proof of Theorem~\ref{thm:existence}.}
Because the horizon and alphabets are finite, we can cast CLSR as a finite CMDP on the finite history state space $\mathcal{H}$.
Define the (time-indexed) \emph{occupancy measures}
\begin{equation}
\begin{gathered}
q_t(h,a)\;=\;\mathbb{P}_\pi(H_t=h,\,A_t=a\mid X=x),\qquad t=1,\dots,T_{\max}.
\end{gathered}
\end{equation}
These satisfy linear flow constraints induced by the kernels $P_\theta^{(k)}$:
for each $t<T_{\max}$ and each $h'\in\mathcal{H}$,
\begin{equation}
\begin{gathered}
\sum_{a'} q_{t+1}(h',a') \;=\; \sum_{h\in\mathcal{H}}\sum_{a\in\mathcal{A}} q_t(h,a)\, P(h'\mid h,a,x),
\end{gathered}
\end{equation}
where $\mathcal{A}=\mathcal{K}\cup\{\mathtt{stop}\}$ and $P(h'\mid h,a,x)$ is the probability of induced transition from history $h$ to $h'$ under action $a$ and query $x$ (obtained by appending a sampled token or absorbing under $\mathtt{stop}$).
Also, $\sum_{a} q_1(h_1,a)=1$ for the initial empty history $h_1$.
The expected token cost is linear:
\begin{equation}
\begin{gathered}
J_C(\pi;x)\;=\;\sum_{t=1}^{T_{\max}}\sum_{h,a} q_t(h,a)\,c(h,a).
\end{gathered}
\end{equation}
Define a terminal reward function $r(h)=\mathbb{P}(Y=g(T,x)\mid X=x,\,T\text{ corresponds to }h)$, so that
\begin{equation}
\begin{gathered}
J_A(\pi;x)\;=\;\mathbb{E}_\pi[r(H_{T_{\max}+1})\mid X=x],
\end{gathered}
\end{equation}
which is also linear in the induced terminal occupancy.
Therefore, constrained optimization
\begin{equation}
\begin{gathered}
\max_{\pi}\;J_A(\pi;x)\quad\text{s.t.}\quad J_C(\pi;x)\le B
\end{gathered}
\end{equation}
is equivalent to a finite-dimensional linear program in the variables $\{q_t(h,a)\}$ on a nonempty, closed, bounded polytope.
A linear objective over a compact polytope attains its maximum, hence an optimal occupancy measure exists.
Finally, standard CMDP realizability guarantees that any feasible occupancy measure corresponds to some (possibly randomized) non-anticipatory policy (one can explicitly construct $\pi_t(a\mid h)\propto q_t(h,a)$ whenever $\sum_{a}q_t(h,a)>0$).
Thus the supremum is attained by a policy $\pi_B^*$, proving the claim.
\hfill$\square$

\paragraph{Proof of Theorem~\ref{thm:lagrange}.}
For each fixed $\lambda\ge 0$, the scalarized problem $\max_\pi J_\lambda(\pi;x)$ is an \emph{unconstrained} finite-horizon MDP with bounded reward.
By backward induction (dynamic programming), an optimal deterministic history-Markov policy $\pi_\lambda^*$ exists.

For the constrained frontier, consider the CMDP linear program from the proof of Theorem~\ref{thm:existence}.
Its Lagrangian dual introduces a multiplier $\lambda\ge 0$ for the token-cost constraint; 
strong duality holds for finite linear programs, so optimal primal values equal optimal dual values.
Consequently, every supported boundary point is the optimizer of $J_A-\lambda J_C$ for some $\lambda$.

If the budget constraint is active and the mapping $\lambda\mapsto J_C(\pi_\lambda^*;x)$ has a discontinuity (typical in discrete CMDPs), an intermediate budget can be met by randomizing between two adjacent multipliers at the start of the episode; 
by convexity of achievable occupancy measures, a mixture of two deterministic policies suffices.
\hfill$\square$

\subsection{Token--accuracy lower bound under arbitrary symbolism}

\paragraph{Binary entropy.}
Let $h_2(\delta)=-\delta\log_2\delta-(1-\delta)\log_2(1-\delta)$ denote the binary entropy (base-2).

\paragraph{Per-token information rate.}
Let $\mathscr{T}=(Z_1,\ldots,Z_L)$ be the variable-length transcript generated by an interactive policy, where $L=|\mathscr{T}|$.
For notational convenience, one may include an explicit terminal symbol in the formal transcript; this changes the practical completion-token count by at most an additive constant.
Define the model- and query-dependent active-token information rate
\begin{equation}
\begin{gathered}
\kappa_\theta(x)
=
\sup_{\pi,t,h_{<t}}
I\!\left(
Y;Z_t
\mid
X=x,\,
Z_{<t}=h_{<t},\,
L\geq t
\right),
\end{gathered}
\end{equation}
where the supremum ranges over policies, time steps, and reachable active histories with nonzero probability.
If $Z_t$ ranges over a finite effective alphabet $\Sigma$, then $\kappa_\theta(x)\leq \log_2|\Sigma|$.

\begin{theorem}[Information-theoretic lower bound on expected generated tokens]
\label{thm:info-lb}
Fix a query $x$ and let $|\mathcal{Y}_x|\ge 2$.
Let $\pi$ be any interactive CLSR policy with final answer $\widehat{Y}=g(\mathscr{T},x)$.
If $\pi$ achieves accuracy at least $\alpha\in(0,1)$,
\begin{equation}
\begin{gathered}
\mathbb{P}_\pi(\widehat{Y}=Y\mid X=x)\ge \alpha,
\end{gathered}
\end{equation}
then with $\delta=1-\alpha$ its expected token cost satisfies
\begin{equation}
\begin{gathered}
\mathbb{E}_\pi[L\mid X=x]\ge
\frac{\max\{I_{\mathrm{req}}(x,\delta),0\}}{\kappa_\theta(x)},
\end{gathered}
\end{equation}
where
\begin{equation}
\begin{gathered}
I_{\mathrm{req}}(x,\delta)
=
H(Y\mid X=x)
-
h_2(\delta)
-
\delta\log_2(|\mathcal{Y}_x|-1).
\end{gathered}
\end{equation}
\end{theorem}

\paragraph{Proof of Theorem~\ref{thm:info-lb}.}
Let $\delta=\mathbb{P}(\widehat{Y}\neq Y\mid X=x)\leq 1-\alpha$.
By Fano's inequality applied to estimating $Y$ from $\widehat{Y}$ over the evaluator-induced answer set $\mathcal{Y}_x$,
\begin{equation}
\begin{gathered}
H(Y\mid \widehat{Y},X=x)
\leq
h_2(\delta)+\delta\log_2(|\mathcal{Y}_x|-1).
\end{gathered}
\end{equation}
Therefore,
\begin{equation}
\begin{gathered}
I(Y;\widehat{Y}\mid X=x)
=
H(Y\mid X=x)-H(Y\mid \widehat{Y},X=x)
\geq
I_{\mathrm{req}}(x,\delta).
\end{gathered}
\end{equation}
Since $\widehat{Y}$ is a deterministic function of $(\mathscr{T},x)$, data processing gives
\begin{equation}
\begin{gathered}
I(Y;\mathscr{T}\mid X=x)
\geq
I(Y;\widehat{Y}\mid X=x)
\geq
I_{\mathrm{req}}(x,\delta).
\end{gathered}
\end{equation}

It remains to upper-bound the information in the transcript by its expected active length.
Using a variable-length chain rule and conditioning on the event that the process is active at step $t$,
\begin{equation}
\begin{gathered}
I(Y;\mathscr{T}\mid X=x)
\leq
\sum_{t=1}^{T_{\max}}
\mathbb{P}_\pi(L\geq t\mid X=x)
\,
\kappa_\theta(x).
\end{gathered}
\end{equation}
The inequality follows from the definition of $\kappa_\theta(x)$ over all reachable active histories; inactive steps emit no new token and contribute no active-token information.
Finally,
\begin{equation}
\begin{gathered}
\sum_{t=1}^{T_{\max}}
\mathbb{P}_\pi(L\geq t\mid X=x)
=
\mathbb{E}_\pi[L\mid X=x],
\end{gathered}
\end{equation}
which yields
\begin{equation}
\begin{gathered}
I_{\mathrm{req}}(x,\delta)
\leq
\kappa_\theta(x)\mathbb{E}_\pi[L\mid X=x].
\end{gathered}
\end{equation}
Taking the nonnegative part of $I_{\mathrm{req}}$ gives the stated bound.
\hfill$\square$

\begin{remark}[On the ``difficulty'' term and evaluator-induced answer sets]
\label{rem:finiteY}
When $Y$ is not naturally finite (e.g., free-form strings), one may let $\mathcal{Y}_x$ be the set of equivalence classes induced by the benchmark evaluator (all outputs judged identical), or consider a finite packing argument (Fano method) on a finite subset of hypotheses.
The same proof yields a lower bound in terms of the entropy/packing size of the induced hypothesis class.
\end{remark}

\subsection{CLSR subsumes program-execution pipelines under interpreter realizability}

\paragraph{Program-execution pipelines.}
A \emph{program-execution} (PE) inference pipeline is any procedure that:
(i) makes a finite number of calls to the frozen backbone LLM (under some prompting format) to produce an LLM-generated \emph{program transcript} $S\in\Sigma^*$,
and (ii) applies a deterministic computable executor $\mathrm{Exec}:\Sigma^*\to\mathcal{Y}_x$ that does not access external knowledge (only performs logic/symbolic manipulation).
The overall output is $\widehat{Y}_{\mathrm{PE}}=\mathrm{Exec}(S)$ (possibly followed by trivial formatting).

Let the total number of LLM-generated tokens in the PE procedure (across all its calls) be $|T_{\mathrm{PE}}|$, with expected cost $\mathbb{E}[|T_{\mathrm{PE}}|\mid X=x]=B$ and accuracy
\begin{equation}
\begin{gathered}
\mathbb{P}(\widehat{Y}_{\mathrm{PE}}=Y\mid X=x)\;\ge\;\alpha.
\end{gathered}
\end{equation}

\paragraph{Interpreter realizability (assumption).}
We formalize the key premise used in the main text.

\begin{assumption}[Interpreter realizability]
\label{assump:interpreter}
For every deterministic computable executor $\mathrm{Exec}:\Sigma^*\to\mathcal{Y}_x$ in the allowed executor class,
there exists a single LSF $S_{\mathrm{Exec}}$ such that, for any input string $s\in\Sigma^*$, one LLM call conditioned on $S_{\mathrm{Exec}}$ and given $s$ as input produces $\mathrm{Exec}(s)$ with failure probability at most $\varepsilon_{\mathrm{int}}<\tfrac12$, and with output length at most $|\mathrm{Exec}(s)|+c_0$ for a universal constant $c_0$.
\end{assumption}

\paragraph{Amplification lemma.}
\begin{lemma}[Majority-vote amplification]
\label{lem:amplify}
Let $U_1,\dots,U_m$ be i.i.d.\ Bernoulli variables with $\mathbb{P}(U_i=1)\ge 1-\varepsilon$ and $\varepsilon<\tfrac12$.
Let $\widehat{U}=\mathrm{Maj}(U_1,\dots,U_m)$.
Then
\begin{equation}
\begin{gathered}
\mathbb{P}(\widehat{U}=0)\;\le\;\exp\!\big(-2m(1/2-\varepsilon)^2\big).
\end{gathered}
\end{equation}
In particular, to make the error at most $\delta$, it suffices to take $m=O(\log(1/\delta))$.
\end{lemma}

\paragraph{Proof of Lemma~\ref{lem:amplify}.}
This is a direct application of Hoeffding's inequality to the sum $\sum_{i=1}^m U_i$.
\hfill$\square$

\begin{theorem}[CLSR subsumes PE under interpreter realizability]
\label{thm:subsumption}
Fix a query $x$ and assume Assumption~\ref{assump:interpreter}.
For any PE pipeline with expected LLM-generated token cost $B$ and accuracy at least $\alpha$, let
\begin{equation}
\begin{gathered}
\bar{\ell}_{\mathrm{Exec}}(x)
=
\mathbb{E}\!\left[|\mathrm{Exec}(S)|+c_0\mid X=x\right],
\end{gathered}
\end{equation}
where $S$ is the PE-generated transcript and $c_0$ is the interpreter-output overhead in Assumption~\ref{assump:interpreter}.
Then for any target $\delta\in(0,1)$, there exists an LSF-only CLSR protocol $\pi$ such that
\begin{enumerate}
\item (\emph{Accuracy}) $\mathbb{P}_\pi(\widehat{Y}=Y\mid X=x)\ge \alpha-\delta$.
\item (\emph{Token cost}) $\mathbb{E}_\pi[|\mathscr{T}|\mid X=x]\le B+O\!\left(\bar{\ell}_{\mathrm{Exec}}(x)\log(1/\delta)\right)$.
\end{enumerate}
If $\bar{\ell}_{\mathrm{Exec}}(x)=O(1)$, as in short-answer benchmark settings, the overhead reduces to $O(\log(1/\delta))$.
\end{theorem}

\paragraph{Proof of Theorem~\ref{thm:subsumption}.}
Construct a CLSR protocol $\pi$ that emulates the PE pipeline in two stages.

\emph{Stage 1: reproduce the PE transcript.}
Because the PE pipeline makes a finite number of frozen-LLM calls under a fixed prompting format, the CLSR pool can include an LSF or fixed LSF-composition plan that reproduces the same prompting format and decoding choices.
Let $S'$ denote the transcript produced by this CLSR stage.
By construction, $S'$ has the same distribution as the PE-generated transcript $S$ up to constant formatting overhead, so
\begin{equation}
\begin{gathered}
\mathbb{E}[|S'|\mid X=x]\le B+O(1).
\end{gathered}
\end{equation}

\emph{Stage 2: internally execute.}
Given $S'$, invoke the interpreter LSF $S_{\mathrm{Exec}}$ from Assumption~\ref{assump:interpreter}.
A single invocation returns $\mathrm{Exec}(S')$ with failure probability at most $\varepsilon_{\mathrm{int}}<1/2$.
Repeat the interpreter call $m$ times independently by resampling and take a majority vote over the produced outputs.
By Lemma~\ref{lem:amplify}, choosing $m=O(\log(1/\delta))$ makes the internal-execution error probability at most $\delta$.

\emph{Accuracy.}
Let $E_{\mathrm{PE}}$ be the event that the PE pipeline is correct, i.e., $\mathrm{Exec}(S)=Y$.
Let $E_{\mathrm{int}}$ be the event that the amplified internal interpreter returns $\mathrm{Exec}(S')$.
Because $S'$ matches the PE transcript distribution, $\mathbb{P}(E_{\mathrm{PE}}\mid X=x)\ge\alpha$.
By amplification, $\mathbb{P}(E_{\mathrm{int}}\mid X=x)\ge 1-\delta$.
A union bound gives
\begin{equation}
\begin{gathered}
\mathbb{P}_\pi(\widehat{Y}=Y\mid X=x)
\ge
\mathbb{P}(E_{\mathrm{PE}}\cap E_{\mathrm{int}}\mid X=x)
\ge
\alpha-\delta.
\end{gathered}
\end{equation}

\emph{Token cost.}
The CLSR transcript consists of the PE-mimicking transcript $S'$ plus $m$ interpreter outputs.
Each interpreter output has length at most $|\mathrm{Exec}(S')|+c_0$.
Therefore,
\begin{equation}
\begin{gathered}
\mathbb{E}_\pi[|\mathscr{T}|\mid X=x]
\le
B+O(1)+
O\!\left(\bar{\ell}_{\mathrm{Exec}}(x)\log(1/\delta)\right),
\end{gathered}
\end{equation}
which proves the stated bound after absorbing constants.
\hfill$\square$

\begin{remark}[When interpreter realizability is plausible or fails]
Assumption~\ref{assump:interpreter} is a modeling premise: it is plausible when $\mathrm{Exec}$ lies within the backbone's algorithmic competence and the required computation fits within effective context and reliability limits; it may fail for executors requiring exact long-horizon computation or strict formal guarantees.
Theorem~\ref{thm:subsumption} should therefore be read as a conditional subsumption: \emph{if} the backbone can act as an interpreter for the executor class, then CLSR can match PE without external execution.
\end{remark}

\clearpage
\section{Appendix: Related work}
\label{app-sec: related_work}

%CLSR is motivated by a central empirical and theoretical question: \emph{how should an LLM allocate generated tokens to maximize inference accuracy under a token budget?}
%Our answer is to treat reasoning as a \emph{representation design} problem:
Instead of emitting verbose natural language traces (CoT) or executable programs (tool-based pipelines), CLSR learns a compact \emph{Language Symbolism Framework} (LSF) and refines it through a small number of recursive rounds.
This insight is also motivated by the view of neuronal communication of interacting neural blocks~\citep{pei2023dynamics,pei2024data,pei2025neuronal}, which are analogous to the communication protocol between agents.
In the following, we position CLSR relative to the most relevant lines of work and clarify the key distinctions.

\textbf{Natural-language reasoning prompts and decomposition.}
Chain-of-Thought prompting and its variants elicit step-by-step natural-language reasoning and typically improve accuracy but often increase generation length and latency~\citep{wei2022chain, kojima2022large}.
A large body of work improves performance by 
(i) more robust decoding/aggregation such as self-consistency~\citep{wang2023self},
(ii) better decomposition protocols such as least-to-most prompting~\citep{zhou2022leasttomost} and plan-and-solve prompting~\citep{wang2023plansolve},
or (iii) automatic prompt optimization/evolution~\citep{yang2023large, fernando2024promptbreeder}.
These methods operate \emph{within} natural language and generally scale the computation by generating more text or more samples.
In contrast, CLSR changes the \emph{intermediate language} itself: it compresses reasoning into LSF and uses recursion to refine the representation while remaining token-efficient.
Thus, CLSR can be viewed as a different point in the design space where the goal is not to elaborate the chain but to \emph{densify} it.

\textbf{Search-style inference over thoughts.}
Tree/graph/search-style inference methods generalize CoT by exploring multiple candidate intermediate steps with self-evaluation, backtracking, or explicit control structures~\citep{yao2023tree, besta2024graph, sel2024algorithm}.
Their strength is improved robustness via exploration, but they incur a higher inference-time cost.
CLSR targets a different regime: it performs \emph{single-trajectory} recursive refinement under a strict token budget.
Rather than exploring many branches, CLSR aims to learn a \emph{stable symbolic scaffold} LSF that reduces the need for breadth.

\textbf{Program-execution pipelines and tool enhancement.}
Program-based methods such as Program-aided Language~\citep{gao2023pal} and Program-of-Thoughts~\citep{chen2023program} instruct the LLM to output executable code and delegate correctness-critical computation to an external runtime ({\it e.g.}, a Python interpreter).
In addition, a broader line of tool-augmented reasoning/acting uses external APIs to reduce hallucination and improve factuality or task completion~\citep{yao2022react,schick2023toolformer}.
CLSR is explicitly \emph{LLM-only}: it does not rely on external compilers or interpreters.
The intermediate representation is symbolic, but \emph{not} a program meant for execution by an external tool; instead, it is designed to be \emph{self-interpretable} by the same LLM under recursion.
This distinction is essential to our scope: CLSR aims to recover much of the benefit of symbolic structure while keeping deployment minimal and avoiding tool dependency.

\textbf{Token-efficient and compressed reasoning traces.}
Recent work explicitly targets shorter reasoning traces, including generating concise intermediate drafts or sketches instead of verbose CoT~\citep{xu2025CoD, aytes2025SoT}.
Other approaches compress reasoning into dense or latent representations, such as compressed contemplation tokens or explicit CoT compression~\citep{nayab2024CCoT, wang2025r1}.
CLSR differs in two ways.
First, CLSR introduces an explicit \emph{discrete} symbolic representation (LSF) with a stable, reusable schema across tasks, rather than only asking the model to ``write less'' in natural language.
Second, CLSR couples the representation with a \emph{recursive refinement operator} and an \emph{adaptive stopping policy}, which is directly connected to our theoretical account of accuracy--token trade-offs.
Empirically, this yields an instance-wise compute allocation: many examples stop at $T=1$, while hard cases receive extra refinement.

\textbf{Test-time self-improvement and iterative refinement.}
A complementary line studies improving model output through iterative self-feedback or reflection at test time~\citep{madaan2023self, shinn2023reflexion, zhang2025reconstruction}.
These approaches typically refine \emph{natural-language outputs} (or agent trajectories) using critique/feedback loops.
CLSR shares the high-level principle of iterative refinement, but focuses specifically on \emph{refining a compact symbolic intermediate} (LSF) to improve \emph{accuracy per token} without requiring external feedback channels or environment interaction.

\textbf{Faithfulness and interpretability of the generated traces.}
Recent studies state that natural-language reasoning traces should not be automatically interpreted as faithful explanations of the causality of internal models~\citep{turpin2023language, barez2025chain}.
CLSR adopts a pragmatic stance: intermediate traces are \emph{algorithmic artifacts} used to improve accuracy--token frontier, not guaranteed causal explanations.
Meanwhile, LSF is intentionally compact and constraint-oriented, making it easier to inspect missing constraints, inconsistent variable bindings, or arithmetic errors than long-form narrative CoT.

\textbf{Summary of novelty.}
Across these related areas, CLSR contributes a distinct combination:
\emph{(i)} a framework that automatically generates discrete language symbolism framework positioned between natural language and executable code,
\emph{(ii)} a recursive refinement operator that improves accuracy without external tools,
and \emph{(iii)} an adaptive compute allocation view and empirical evidence that directly targets the optimal accuracy--token frontier.


% You can have as much text here as you want. The main body must be at most $8$
% pages long. For the final version, one more page can be added. If you want, you
% can use an appendix like this one.

% The $\mathtt{\backslash onecolumn}$ command above can be kept in place if you
% prefer a one-column appendix, or can be removed if you prefer a two-column
% appendix.  Apart from this possible change, the style (font size, spacing,
% margins, page numbering, etc.) should be kept the same as the main body.
%%%%%%%%%%%%%%%%%%%%%%%%%%%%%%%%%%%%%%%%%%%%%%%%%%%%%%%%%%%%%%%%%%%%%%%%%%%%%%%
%%%%%%%%%%%%%%%%%%%%%%%%%%%%%%%%%%%%%%%%%%%%%%%%%%%%%%%%%%%%%%%%%%%%%%%%%%%%%%%

\end{document}